%% file: neurips_2026.tex
\documentclass{article}

\usepackage[main, final]{neurips_2026}

\usepackage[utf8]{inputenc} 
\usepackage[T1]{fontenc}    
\usepackage{hyperref}       
\usepackage{url}            
\usepackage{booktabs}       
\usepackage{amsfonts}       
\usepackage{nicefrac}       
\usepackage{microtype}      
\usepackage{xcolor}         
\usepackage{graphicx} 
\usepackage{amsmath} 
\usepackage{wrapfig}  
\usepackage{makecell}
\usepackage{tikz}
\usepackage{dsfont}
\usetikzlibrary{positioning, arrows.meta, shapes.geometric}
\title{Leveraging Dale’s Principle as an Inductive Bias in Recurrent Neural Networks}

\author{
  Jiyi Wang$^{1}$ \quad Chenxiao Yang$^{2}$ \quad Yujia Zhao$^{1}$ \quad \textbf{Jingzhao Zhang}$^{2}$ \quad \textbf{Lu Mi$^{2}$}\thanks{Corresponding author: \texttt{milu@mail.tsinghua.edu.cn}.} \\
  $^{1}$Shanghai Qi Zhi Institute, Shanghai, China \quad $^{2}$Tsinghua University, Beijing, China
}

\begin{document}

\maketitle

\input{paper}



\end{document}

%% file: paper.tex
\begin{abstract}\label{sec:abstract}

\input{section/abstract}
\end{abstract}

\section{Introduction}\label{sec:intro}

\input{section/introduction}

\section{Related work}\label{sec:related_work}
\input{section/related_work}

\section{Methods}\label{sec:methods}

\input{section/method}

\section{Experiments on cognitive tasks}\label{sec:exp}

\input{section/experiment_cog}

\section{Extension to challenging learning regimes}\label{sec:result_generality}
\input{section/extension}

\section{Ablation study}\label{sec:result_ablation}

\input{section/ablation}

\section{Conclusion \& discussion}\label{sec:discussion}

\input{section/discussion}

\bibliography{reference}
\bibliographystyle{unsrt}

\input{section/appendix}\label{sec:appendix}
\clearpage


%% file: section/abstract.tex
One of the fundamental differences between biological neural networks and artificial neural networks (ANNs) is that biological neurons obey Dale’s principle: each neuron projects either exclusively excitatory or exclusively inhibitory outputs. While this constraint is central to biological circuits, previous attempts to impose Dale’s principle in ANNs have often led to training instability or degraded performance. In this work, we introduce EISep RNN, a recurrent neural network architecture that implements Dale’s principle through a learnable excitatory–inhibitory separation. Across a series of learning settings, including multi-task learning, continual learning, and reinforcement learning, EISep achieves more stable training dynamics, competitive or improved performance, and sparser, more modular connectivity than vanilla RNNs and previous Dale-constrained models. EISep also exhibits stable and informative latent dynamics. These results establish Dale’s principle as a biologically grounded and computationally effective inductive bias for recurrent neural networks. 

%% file: section/introduction.tex
\begin{wrapfigure}{r}{0.4\textwidth}
    \centering
    \vspace{-0.5em}
    \includegraphics[width=\linewidth]{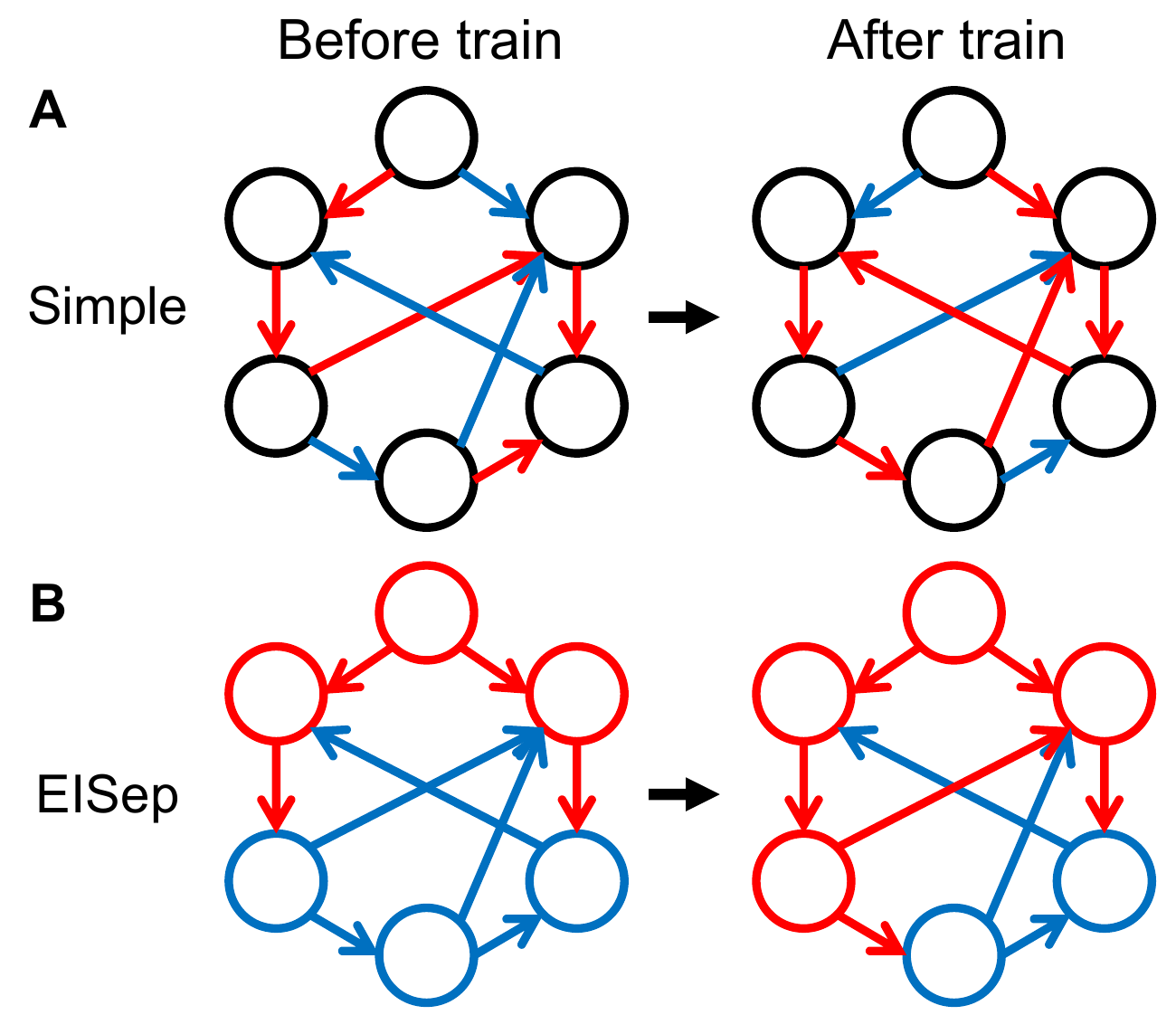}
    \caption{\textbf{A.} In Simple RNN, each neuron can project positive (red) \textit{and} negative (blue) weights. \textbf{B.} In EISep RNN, each neuron can only project positive weights (excitatory), \textit{or} negative weights (inhibitory). This sign is learnable (left to right) during training.}
    \label{fig:teaser}
    \vspace{-1em}
\end{wrapfigure}
Neuroscience and artificial intelligence share a long history of cross-disciplinary research, where breakthroughs in one field routinely drive progress in the other~\citep{fukushima1980neocognitron, schultz1997neural, whittington2021relating, liu2023seeing, kim2023transformer, li2023emergence, zador2023catalyzing, jiang2025neural, darlow2025continuous}. A prominent direction in NeuroAI is to translate biological principles into inductive biases through architectural constraints or learning rules~\citep{mathis2025leveraging, ramadan2025computational, wang2025hierarchical}. For example, explicitly encouraging modularity can improve generalization in reinforcement learning~\citep{zhang2024inductive}, imposing synaptic diversity can improve stability and robustness in vision models~\citep{hofmann2025concept}, and biologically inspired memory-regulation mechanisms can alleviate catastrophic forgetting~\citep{wang2023incorporating}. Apart from improving artificial intelligence (AI), these inductive biases also produce new tools for neuroscience: networks that not only perform well, but do so in a manner more consistent with how real neural circuits operate. One potentially underexplored inductive bias is Dale's principle, which we investigate in this paper.


Dale's principle states that, each biological neuron is either excitatory or inhibitory, and only projects either positive \textit{or} negative weights to all downstream neurons (Fig.~\ref{fig:teaser}B left). This reflects the fact that a neuron releases the same type of neurotransmitter at all its synapses~\citep{dale1935pharmacology, eccles1976electrical}. In contrast, artificial neural networks (ANNs) typically allow each unit to project both positive \textit{and} negative weights (Fig.~\ref{fig:teaser}A left). The lack of excitatory-inhibitory separation simplifies optimization but deviates from biological organization. Previous work has suggested that enforcing Dale’s principle may reduce network capacity~\citep{balwani2025constructing}, narrow the loss basin~\citep{hofmann2025concept} and impair learning performance~\citep{cornford2020learning, li2023learning}. As a result, most ANNs have continued to use unconstrained units, prioritizing trainability over biological realism, and leaving Dale-constrained networks largely underexplored in practice.

In this work, we introduce EISep RNNs – a new recurrent neural network architecture that implements Dale’s principle as an inductive bias. EISep RNNs naturally develop modular structure, sparse connectivity, and stable dynamics. Crucially, these structural and dynamical benefits translate into improved performance in multi-task learning, continual learning and reinforcement learning, showing that biological realism and effective learning are not necessarily contradictory.

%% file: section/related_work.tex
In computational neuroscience literature, Dale's principle has become a widely adopted assumption to study the neural dynamics~\citep{van1996chaos, van1998chaotic, liang2025excitation, deneve2016efficient, tian2020excitation, lim2014balanced, ingrosso2019training}.  Beyond its use in models of biological circuits, several studies have incorporated E-I separation into trainable neural networks. For example, ~\citep{song2016training} trained E–I RNNs on cognitive tasks and reproduced features of population activities, while ~\citep{minni2019understanding} investigated emergent E-I structural differences. Other work developed trainable Dale-constrained ANNs on general supervised learning tasks~\citep{cornford2020learning}, used spectral analysis to improve their learning performance~\citep{li2023learning}, and combined Dale-constrained learning with pruning~\citep{balwani2025constructing}. Together, these studies establish that Dale-constrained networks can support effective learning and provide useful models of neural circuits.

From a machine learning perspective, requiring each neuron to have outgoing weights of a single sign restricts the available connectivity patterns at a fixed hidden size. Several early attempts to impose E-I separation revealed substantial difficulties. Simply constraining networks’ neurons to fixed excitatory or
inhibitory identities often narrows the solution space and makes optimization harder, leading to
degraded performance and slower or unstable training~\citep{hofmann2025concept, chelu2025excitatory}.These observations motivated a series of recent works that revisit Dale’s principle and propose architectures to recover performance while
respecting sign constraints~\citep{song2016training, cornford2020learning, li2023learning, balwani2025constructing}. While effective in certain scenarios, such designs are often sensitive to task complexity and optimization details. As a result, Dale’s principle is frequently viewed as a constraint rather than an advantage. Our work challenges this prevailing view and reveals that Dale's principle can actually act as a beneficial inductive bias for RNNs.

%% file: section/method.tex
\subsection{Background and Notation}

\input{section/methods/background}
\label{sec:background}

\subsection{Definition of EISep}

\input{section/methods/definition}\label{sec:definition_eisep}

\subsection{Baselines}

\input{section/methods/baseline}\label{sec:baseline}


%% file: section/methods/background.tex
We first introduce the following notations for RNN operations. Consider an RNN with hidden size $N$, activation function $\phi$, recurrent weight $W$, input weight $U$, output weight $V$, and bias terms $b_h, b_o$. At time step $t$, it receives input $x_t$, maintains hidden state $h_t$, and produces output $o_t$:
\vspace{-0.5em}
\begin{align}
    h_{t} = \phi(Wh_{t-1}+Ux_t+b_h),\quad o_t = Vh_t +b_o.
\end{align}
Enforcing Dale's principle during learning is nontrivial. Simply initializing neurons as excitatory or inhibitory is insufficient, since unconstrained optimization will flip the weight signs over time. A common approach~\citep{song2016training, li2023learning, balwani2025constructing} is to parameterize the recurrent weight matrix as $W=QD$, where $Q$ is a learnable non-negative matrix, and $D$ is a fixed diagonal sign matrix with $D_{ii}=\pm1$. The non-negativity of $Q$ is typically enforced via gradient masking (zeroing out violating updates) or by applying non-negative transformations to the parameters. We briefly review prior Dale-constrained ANNs in Table~\ref{tab:dale_init_train}, with additional details provided in Appendix~\ref{sec:appendix_baseline_init}.

Despite their efforts, current Dale-constrained networks still struggle to match the performance of vanilla RNN~\citep{chelu2025excitatory}. We attribute this to two key limitations. \textbf{(1) Fixed neuron identity.} The sign matrix \(D\) is fixed after initialization, preventing the network from adapting excitatory and inhibitory types during training, which constrains the optimization trajectory. \textbf{(2) Initialization.} To enforce non-negativity, they sample weights from gamma~\citep{song2016training} or exponential~\citep{cornford2020learning, li2023learning} distributions, which can induce a large initial spectral radius and may harm training stability.

\begin{wraptable}{r}{0.45\linewidth}
\centering
\footnotesize
\vspace{-1em}
\setlength{\tabcolsep}{8pt}
\renewcommand{\arraystretch}{1}
\caption{
Comparison of initialization methods and E-I sign learnability
across Dale-constrained ANNs. Rect.\ Gauss.\ denotes a rectified Gaussian distribution.
}
\vspace{0.3em}  
\begin{tabular}{lccc}
\toprule
Model &
Initialization &
E-I sign\\
\midrule
Simple
& Gaussian
& --\\
ColEI~\citep{li2023learning}
& Exponential
& fixed\\
DBPN~\citep{balwani2025constructing}
& Uniform
& fixed\\
DANN~\citep{cornford2020learning}
& Exponential
& fixed\\
EISep (ours)
& Rect.\ Gauss.
& learnable\\
\bottomrule
\end{tabular}
\vspace{-3em}
\label{tab:dale_init_train}
\end{wraptable}

To address these concerns, we make two adjustments in EISep. \textbf{(1)} We make the sign matrix \(D\) learnable, allowing neuron types to adapt during training (Fig.~\ref{fig:teaser}B), which enables the network to explore different E/I organizations and follow more flexible optimization trajectories. \textbf{(2)} We sample weights from a rectified Gaussian distribution which better controls the initial spectral radius. We verify the roles of these two designs in the ablation study (Sec.~\ref{sec:result_ablation}). Next, we give the formal definition of EISep.

%% file: section/methods/definition.tex
We introduce a learnable sign assignment using the Gumbel-Softmax (GSoftmax) reparameterization~\citep{jang2016categorical}. Let $M\in\mathbb{R}^{N\times2}$ be a learnable parameter matrix. For each neuron $i$, we sample a soft sign mask $m_i\in\mathbb{R}^2$ as $m_i=\mathrm{GSoftmax}(M_i), \quad i=1,...,N$, and define a soft sign value $e_i=m_i[0]-m_i[1].$ We then construct a diagonal soft sign matrix $D$ with
\vspace{-0.6em}
\begin{align}
    D_{ij} = 
    \left\{ \renewcommand{\arraystretch}{0.8}
    \begin{array}{ll}
    e_i, & i = j,\\
    0,   & i \ne j.
    \end{array}
    \right.
\end{align}
The soft sign value $e_i$ is concentrated around $\pm1$ during training and exactly equals $\pm1$ during test (Appendix~\ref{sec:appendix_gumbel_softmax_sampling}). This parametrization enables end-to-end gradient-based optimization, while allowing neuron identities to adapt during training. At test time, the sign mask is discretized to remove sampling noise by setting
$e_i = 1$ if $M_i[0] > M_i[1]$, and $-1$ otherwise.

Let $W_0\in\mathbb{R}^{N\times N}$ be a learnable parameter matrix with entries initialized as $W_{0,ij}\sim\mathcal{N}(0, 1/N)$. The recurrent weight $W$ is defined as:
\vspace{-0.6em}
\begin{align}{\label{eq:w=w0d}}
    W = \mathrm{ReLU}(W_0)D.
\end{align}
Under this parameterization, the value matrix $Q=\mathrm{ReLU}(W_0)$ is initialized using rectified Gaussian distribution, which leads to a smaller initial spectral radius (Fig.~\ref{fig:appendix_spectral_radius_initialization}), and is associated with improved training stability.

To encourage stable neuron identities during later training stages, we introduce an entropy 
regularization term $L_e=\frac{1}{N}\sum_i H(\mathrm{softmax}(M_i))$, where 
$H(p)=-\sum_j p_j\log p_j$, and optimize $L=L_t+\lambda_e L_e$. This completes the definition of the EISep RNN. 

%% file: section/methods/baseline.tex
We compare EISep against a vanilla RNN baseline (denoted \emph{Simple}) as well as several Dale-constrained architectures, including DANN and ColEI. As our experiments use the Adam optimizer, while the original implementations of DANN and ColEI only employ stochastic gradient descent (SGD), we additionally implement Adam-based variants of these models to ensure a fair comparison. We denote these baselines as DANN-SGD, DANN-Adam, ColEI-SGD, and ColEI-Adam. 

%% file: section/experiment_cog.tex
\subsection{Performance}\label{sec:result_performance}
\input{section/experiment_cog/result1_performance}

\subsection{Training stability}\label{sec:result_stability}
\input{section/experiment_cog/result2_stability}

\subsection{Emergent properties}

The above results suggest that EISep promotes stable training dynamics. We next investigate whether its architectural constraints also give rise to structured connectivity. Specifically, we examine two emergent structural properties: \textit{modularity} and \textit{sparsity}.

\subsubsection{Modularity}\label{sec:result_modularity}

\input{section/experiment_cog/modularity}

\subsubsection{Sparsity}\label{sec:result_sparsity}
\input{section/experiment_cog/sparsity}

%% file: section/experiment_cog/result1_performance.tex
\begin{figure}[t]
    \centering
\includegraphics[width=\linewidth]{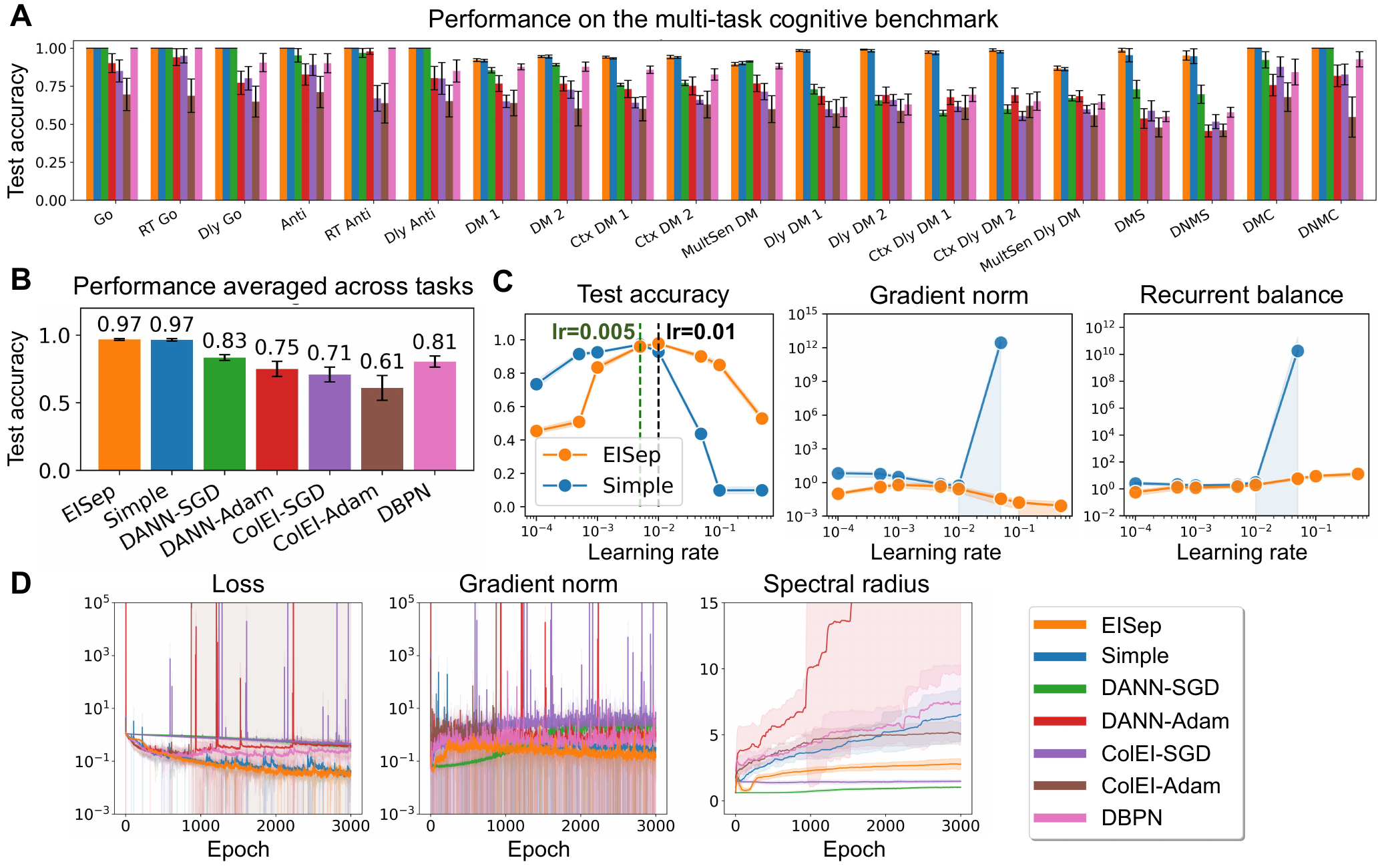}
    \vspace{-0.2in}
  \caption{Multi-task cognitive benchmark consisting of 20 cognitive tasks. The same legend at the right of \textbf{D} is used throughout the figure. \textbf{A.} EISep consistently achieves higher accuracy than prior Dale-constrained baselines and matches Simple across tasks. \textbf{B.} EISep achieves the highest average performance. \textbf{C.} At high learning rates, EISep maintains higher test accuracy (left), lower gradient norms (middle), and tighter recurrent balance (right) than Simple. \textbf{D.} EISep exhibits lower decoding loss (left), lower gradient norms (middle), and more stable spectral radius trajectories (right) throughout training, indicating more stable dynamics. All models use the same learning rate $10^{-2}$ except for \textbf{C}.}
    \vspace{-0.2in}
  \label{fig:cog_performance_stability}
\end{figure}

We first evaluate EISep on a multi-task cognitive benchmark widely used to study multi-task capacity and representations of RNNs \citep{yang2019task, driscoll2024flexible}. The benchmark consists of 20 interrelated tasks that require a single RNN to flexibly switch between task rules across trials. In each trial, the network receives a rule input indicating the active task, followed by noisy sensory stimuli that provide evidence toward choices. The correct choice and stimulus strength are randomized on each trial, yielding a mixture of easy and difficult decisions. This benchmark tests working memory, perceptual decision-making and rule-based context switching in a single RNN. Additional details are provided in Appendix~\ref{sec:appendix_hyperparameter} and \ref{sec:appendix_env}.

Across the full task suite, EISep consistently achieves higher accuracy than prior Dale-constrained architectures and matches Simple's performance (Fig.~\ref{fig:cog_performance_stability}A). This advantage is also evident when performance is averaged across tasks (Fig.~\ref{fig:cog_performance_stability}B). These results demonstrate that our implementation achieves better performance than previous works, and that enforcing Dale’s principle does not necessarily compromise task performance.

%% file: section/experiment_cog/result2_stability.tex
To evaluate training stability, we track the decoding loss, gradient norm, and spectral radius of the recurrent weight matrix throughout training (Fig.~\ref{fig:cog_performance_stability}D). Most baseline models exhibit occasional spikes in loss and gradient norms, whereas EISep exhibits lower and more stable curves. Moreover, EISep achieves tighter control of spectral radius than most baselines. These results suggest EISep stabilizes training dynamics. When tested across different learning rates, EISep performs better at higher learning rates (Fig.~\ref{fig:cog_performance_stability}C, left), which correlates with its lower gradient norms (Fig.~\ref{fig:cog_performance_stability}C, middle). 

In biological networks, the balance between excitatory and inhibitory inputs is known to stabilize dynamics and improve computational capacity~\citep{deneve2016efficient}. To assess whether a similar mechanism emerges, we quantify recurrent balance as the average input to each neuron. We find that EISep achieves a tighter balance than Simple across learning rates (Fig.~\ref{fig:cog_performance_stability}C, right). Overall, these results indicate that by enforcing a biologically realistic E-I balance, EISep keeps the spectral radius bounded, mitigates gradient explosion and promotes stable optimization. More metrics are provided in Appendix~\ref{sec:appendix_multilr}.


To ensure the fairness of the comparison, we further test all baseline models across different learning rates (Appendix~\ref{sec:appendix_multilr}). We find that EISep performs well across a relatively larger scope of learning rates. In other words, it can achieve a wider stable zone than most other baselines. Moreover, its stable zone resides at higher learning-rate areas, consistent with the stabilization effects induced by balanced recurrent dynamics.




%% file: section/experiment_cog/modularity.tex
\begin{figure}{}
    \centering
\includegraphics[width=\linewidth]{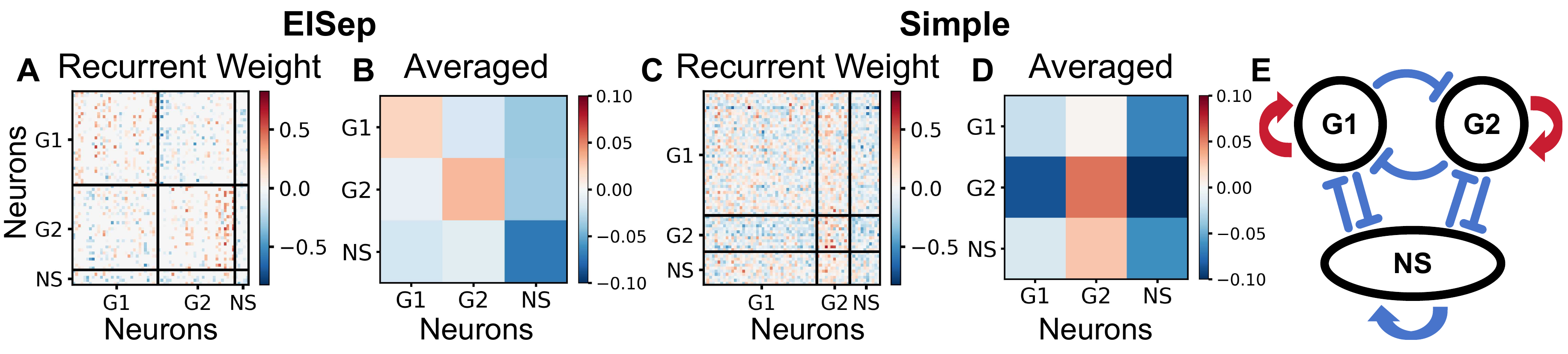}
    \vspace{-0.2in}
  \caption{Case study on the 2AFC task. Neurons are grouped by selectivity using the signal-detection index \(d'\). \textbf{A.} EISep exhibits a clear block structure in the recurrent weight $W$, with within-group excitation and cross-group inhibition. \textbf{B.} Averaged connectivity within neuron groups further reveals strong self-excitation and mutual inhibition. \textbf{C-D.} Simple lacks clear separation between intra-group and inter-group connectivity. \textbf{E.} The emergent EISep connectivity resembles the conceptual winner-take-all circuit commonly used in perceptual decision-making models~\citep{wong2006recurrent}.}
      \vspace{-0.1in}
  \label{fig:cog_case}
\end{figure}

The neural circuits have functionally specialized modules to support efficient computation~\citep{grossberg2000complementary, bertolero2015modular}. To investigate whether Dale's principle promotes modular and interpretable network structure, we test EISep on a noisy two-alternative forced-choice (2AFC) perceptual decision-making task where the network must choose the stronger signal out of two noisy inputs. We use signal-detection index $d'$ (Appendix~\ref{sec:appendix_case_modularity}) to quantify the neuron selectivity and partition the neurons into Group 1 (G1), Group 2 (G2) and Non-selective (NS). EISep develops a clear block structure in recurrent weight $W$ (Fig.~\ref{fig:cog_case}A, B). Weights within groups are primarily excitatory, while weights between groups are primarily inhibitory. This produces two populations with strong self-excitation and cross-inhibition. This organization resembles a winner-take-all circuit model commonly adopted in perceptual decision-making studies (Fig.~\ref{fig:cog_case}E, \cite{wong2006recurrent}). In contrast, Simple RNN has no clear distinction between intra-group and inter-group connections (Fig. \ref{fig:cog_case}C, D).

\begin{figure}{}
    \centering
\includegraphics[width=0.9\linewidth]{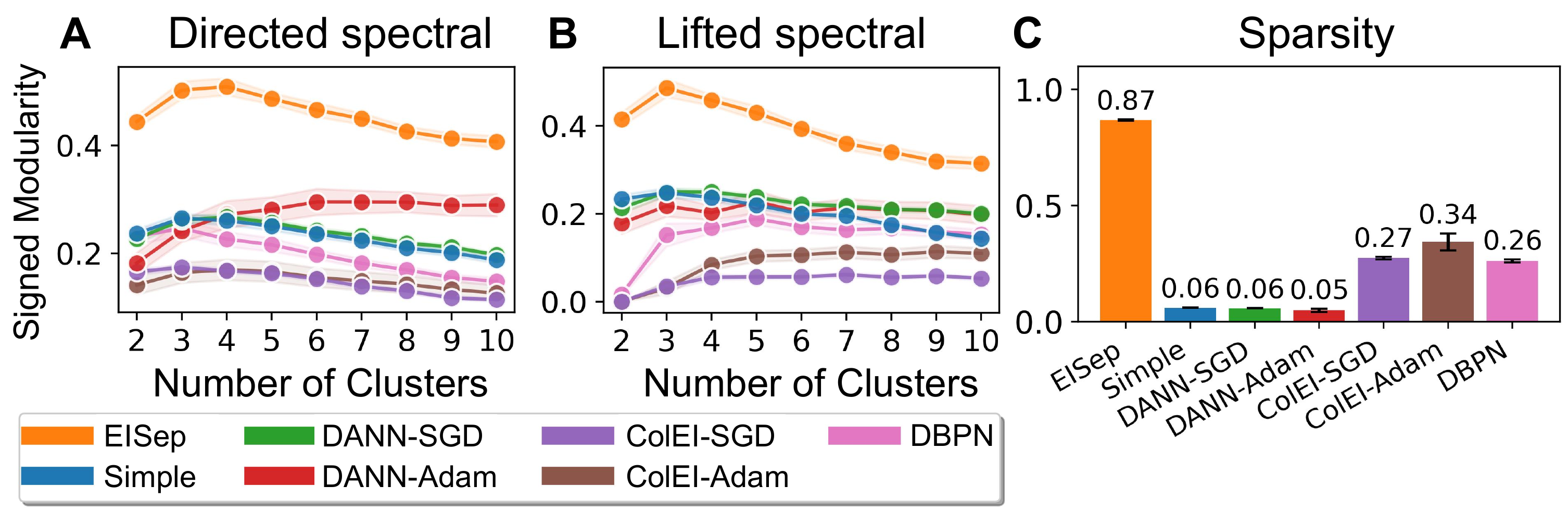}
    \vspace{-0.1in}
  \caption{Population analysis on the multi-task cognitive benchmark. Neurons are partitioned into groups using two clustering methods. EISep achieves the highest modularity under (\textbf{A.}) directed spectral clustering and (\textbf{B.}) lifted spectral clustering, indicating more modular circuit organization. \textbf{C.} EISep also achieves substantially higher recurrent weight sparsity than baseline models.}
    \vspace{-0.1in}
  \label{fig:modularity}
\end{figure}

Next, we carry out population analysis to quantitatively examine the network modularity. We use spectral clustering methods to partition the neurons into groups, and use signed modularity (Appendix~\ref{sec:appendix_population_modularity}) to measure if there are larger intra-group connections and smaller inter-group connections. EISep exhibits the highest modularity across both directed spectral clustering (Fig.~\ref{fig:modularity}A) and lifted spectral clustering (Fig.~\ref{fig:modularity}B). This comparison suggests that enforcing E-I separation biases the network towards modular and interpretable circuit motifs, whereas unconstrained RNNs rely on more distributed and less structured connectivity.

    \vspace{-0.1in}

%% file: section/experiment_cog/sparsity.tex
Sparsity is a hallmark of biological neural circuits, and is widely believed to support energy-efficient and structured computation in both neuroscience~\citep{litwin2017optimal, fruengel2025sparse} and AI~\citep{bartoldson2020generalization, hoefler2021sparsity}. Here, we define sparsity as the fraction of small weights ($|w|<0.1\sigma$), and provide the detailed definition and validity check in Appendix~\ref{sec:appendix_sparsity_validity}. EISep achieves substantially higher sparsity than baseline models in multi-task cognitive benchmark (Fig.~\ref{fig:modularity}C) and RL tasks (Appendix~\ref{sec:appendix_rl_sparsity}). The increased sparsity induced by EISep suggests that E-I separation encourages more efficient connectivity, without compromising task performance.


%% file: section/extension.tex
Given the computational and structural advantages, we test whether they translate into improved performance in challenging scenarios. Specifically, building on the modularity advantages (Sec.~\ref{sec:result_modularity}), we investigate whether Dale's principle helps resist gradient interference and mitigate catastrophic forgetting in continual learning. Building on the stability advantages (Sec.~\ref{sec:result_stability}), we investigate whether it reduces gradient instability and improves performance in reinforcement learning. 
\subsection{Continual learning}\label{sec:continual_learning}
\input{section/general_task/continual_learning}

\subsection{Reinforcement learning}\label{sec:reinforcement_learning} 
\subsubsection{Performance}\label{sec:rl_performance}
\input{section/general_task/reinforcement_learning}

\subsubsection{Internal dynamics}\label{sec:rl_dynamics}

\input{section/general_task/rl_dynamics}

In summary, EISep produces more stable optimization and better performance in continual learning and reinforcement learning problems. We also include traditional RNN tasks such as Sequential MNIST in Appendix~\ref{sec:appendix_supervised_learning}, where EISep can still outperform most Dale-constrained baselines. We discuss the importance of selecting appropriate benchmarks in evaluating inductive biases in Sec.~\ref{sec:discussion}.

%% file: section/general_task/continual_learning.tex
\begin{figure}[t]
    \centering
\includegraphics[width=\linewidth]{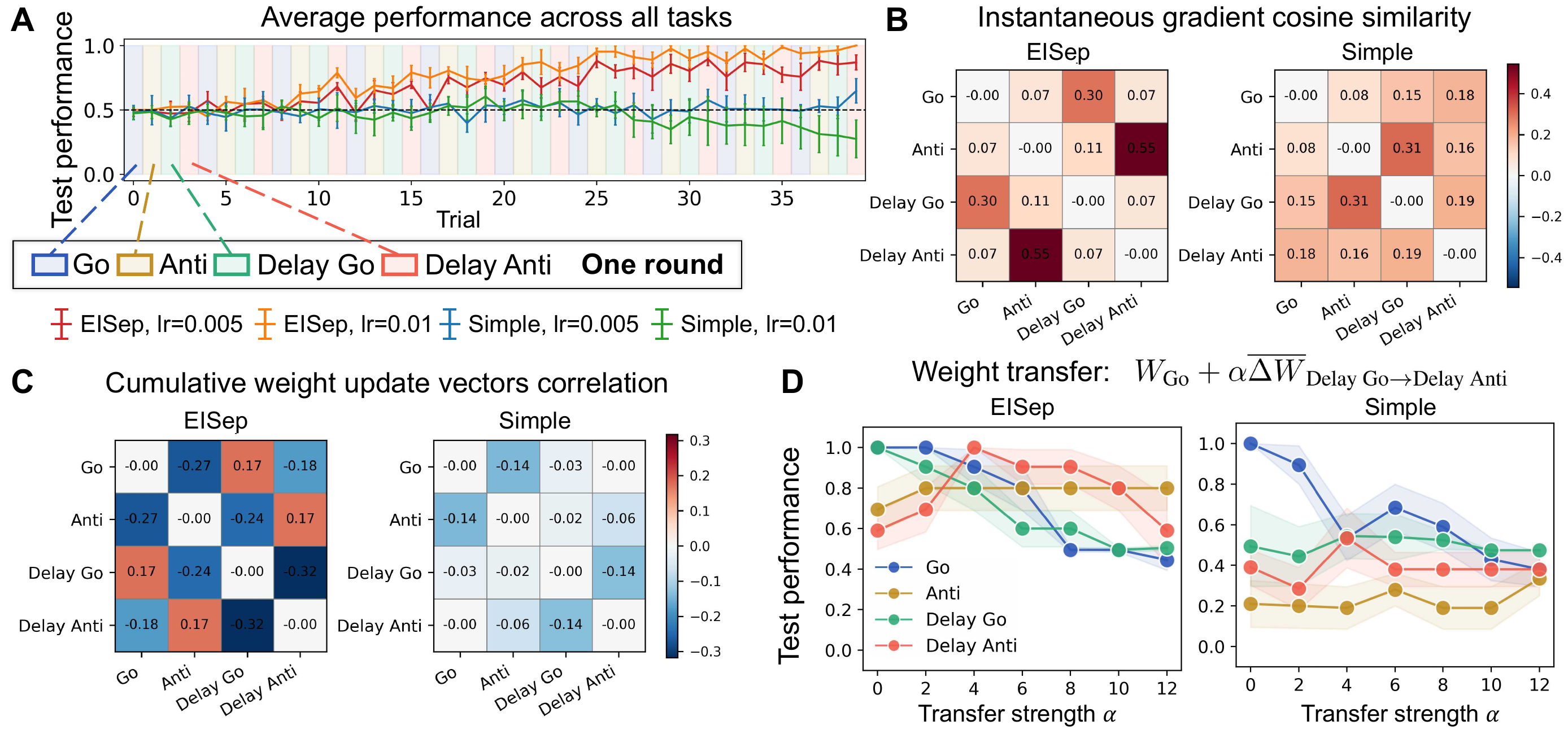}
    \vspace{-0.2in}
  \caption{Continual learning on four sequentially presented cognitive tasks repeated over 10 rounds. \textbf{A.} EISep achieves higher performance and stronger resistance to catastrophic forgetting than Simple across learning rates. \textbf{B.} EISep exhibits high gradient similarity for similar tasks (\textit{Go} vs. \textit{Delay Go}) and near-zero similarity for opposing tasks (\textit{Go} vs. \textit{Anti}), indicating reduced gradient interference. \textbf{C.} EISep shows structured correlations between task-induced weight updates, with positive correlations for similar tasks and negative correlations for opposing tasks. \textbf{D.} EISep's performance shifts from \textit{Go} to \textit{Anti} as transfer strength $\alpha$ increases.}
    \vspace{-0.2in}
  \label{fig:biology_continual}
\end{figure}

Continual learning is a non-stationary problem where different tasks are presented sequentially to the network. The gradients from different tasks often interfere with each other, leading to unstable optimization and catastrophic forgetting~\citep{mccloskey1989catastrophic, wang2024comprehensive}. On the other hand, modular networks can mitigate this by allocating tasks to different subspaces, effectively reducing gradient interference~\citep{duncker2020organizing}. 

We build a continual learning dataset using four tasks from the multi-task cognitive benchmark (Sec.~\ref{sec:result_performance}): \textit{Go}, \textit{Anti}, \textit{Delay Go}, and \textit{Delay Anti}. In each trial, the agent receives a noisy two-channel stimulus and must select one channel. In the two Go tasks, the correct response is the stronger channel, whereas in the two Anti tasks, it is the weaker one. In the non-delay tasks, responses are required immediately right after the stimulus period, while the two delay tasks insert a silent memory period between stimulus and response, requiring the network to maintain internal representations in the absence of external input (Appendix~\ref{sec:appendix_env}). Together, these tasks constitute a $2\times2$ design that disentangles sensory-response mapping (Go vs. Anti) and memory requirements (Delay vs. non-Delay). These tasks are presented sequentially for several times to the network. A similar setup has been used as a benchmark for continual learning algorithms \citep{duncker2020organizing}.

We train each model using the optimal learning rates from the multi-task learning (Simple: $5\times10^{-3}$; EISep: $1\times10^{-2}$, Fig.~\ref{fig:cog_performance_stability}C left). Performance on all four tasks is evaluated at every trial (Fig.~\ref{fig:biology_continual}A). At both learning rates, Simple rapidly forgets earlier tasks as training proceeds, while EISep performs better than Simple and other baselines (Appendix.~\ref{sec:appendix_continual_learning}). It also exhibits smoother loss curves (Appendix~\ref{sec:appendix_continual_loss}).

Next, we investigate whether EISep mitigates gradient interference. In continual learning, updates that improve the current task can impair the performance of previously learned tasks, causing catastrophic forgetting~\citep{lopez2017gradient, riemer2018learning}. We use the cosine similarity between task-specific gradients as a local diagnostic of the interference~\citep{riemer2018learning, yu2020gradient}. Positive similarity indicates that an update can benefit both tasks; negative similarity indicates conflict; and near-zero similarity indicates orthogonal local update directions. We record the average gradients when training each task and compute the cosine similarity between task-specific gradients (Fig.~\ref{fig:biology_continual}B). EISep exhibits high similarity for tasks with identical stimulus–response mappings (e.g., \textit{Go} vs. \textit{Delay Go}) and near-zero similarity for opposing tasks (e.g., \textit{Go} vs. \textit{Anti}). In contrast, Simple shows comparable similarity for both types. These results suggest that EISep exhibits gradient alignment that reflects task relationships, with more aligned directions for tasks sharing stimulus–response mappings and approximately orthogonal directions for tasks with opposing mappings.

Finally, we examine whether EISep organizes task representations in a structured manner. For each task, we compute the cumulative weight update $\Delta W=W_{\text{after}}-W_{\text{before}}$, and measure correlations between these task-induced updates (Fig.~\ref{fig:biology_continual}C). Since this cumulative update integrates many steps of parameter updates, it probes the global organization of the weight transitions between tasks. In EISep, updates are positively correlated for same-type tasks and negatively correlated for opposing tasks, while Simple lacks such patterns. Although negative update correlations may appear inconsistent with near-zero gradient similarity, cumulative updates are also shaped by optimizer dynamics and training history~\citep{kingma2014adam}, and are not necessarily aligned with instantaneous gradients. 

To further probe the update geometry, we perform a weight-transfer experiment, analogous to task-vector arithmetic in~\citep{ilharco2022editing}. If weight updates encode reusable task transformations, an update learned between one pair of tasks should induce the corresponding behavioral change when applied to another task with the same underlying relationship. Specifically, we normalize the update vector from \textit{Delay Go} to \textit{Delay Anti}: $\overline{\Delta W}_{\text{DG}\rightarrow\text{DA}}={\Delta W}_{\text{DG}\rightarrow\text{DA}}/||{\Delta W}_{\text{DG}\rightarrow\text{DA}}||_F$, average it across trials, and add it to the weights after training on \textit{Go}, $W_{\text{Go}}+\alpha\overline{\Delta W}_{\text{DG}\rightarrow\text{DA}}$. As $\alpha$ increases, performance shifts from \textit{Go} toward \textit{Anti} in EISep, but not in Simple (Fig.~\ref{fig:biology_continual}D). These results show that the weights of EISep lie in a structured subspace, which enables efficient parameter reuse and reduced interference, while Simple has more entangled weights.


%% file: section/general_task/reinforcement_learning.tex
\begin{figure}[t]
    \centering
\includegraphics[width=\linewidth]{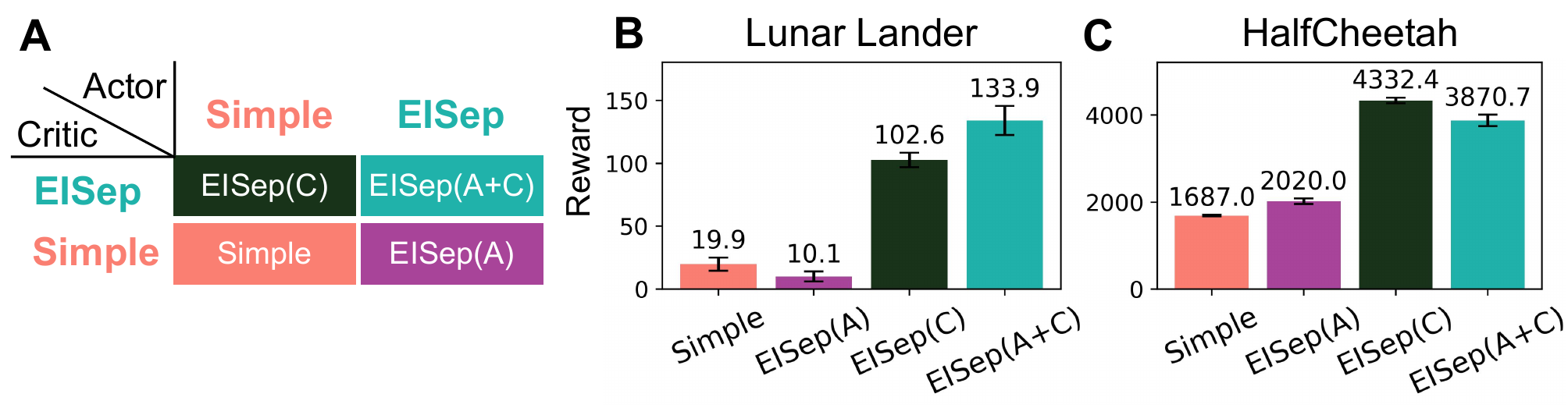}
    \vspace{-0.2in}
  \caption{Reinforcement learning benchmarks using recurrent A2C agents. \textbf{A.} Four actor--critic variants combining Simple and EISep recurrent modules. \textbf{B.} On the discrete Lunar Lander task, EISep achieves higher rewards than Simple. \textbf{C.} On the more challenging HalfCheetah task, EISep again outperforms Simple, demonstrating improved optimization stability and control performance across discrete and continuous RL settings.}
    \vspace{-0.2in}
  \label{fig:rl_best_model_performance_eisep}
\end{figure}

Reinforcement learning (RL) tests adaptive control in complex environments, and is well known to suffer from unstable optimization and high-variance gradient updates~\citep{mnih2015human}. EISep's stable dynamics observed before may help alleviate this problem. 

Building on the Advantage Actor–Critic (A2C~\citep{mnih2016asynchronous}) framework, we evaluate EISep and Simple on two RL benchmarks: \textit{Lunar Lander}~\citep{towers2024gymnasium}, a discrete control task, and \textit{HalfCheetah}~\citep{wawrzynski2009cat}, a more challenging continuous control task. To isolate architectural effects on actor and critic, we use separate recurrent modules for them. This yields four variants (Fig.~\ref{fig:rl_best_model_performance_eisep}A): Simple (Simple for both actor and critic), EISep(A) (EISep for the actor and Simple for the critic), EISep(C) (EISep only for the critic), and EISep(A+C) (EISep for both). To accelerate training in HalfCheetah, we upgrade the backbone from vanilla RNN to LSTM (Appendix~\ref{sec:appendix_eiseplstm}). We report the best performance across actor and critic learning rates for each variant. Full results are provided in Appendix~\ref{sec:appendix_lunarlander_eisep_lrall}. 


On both tasks, EISep achieves higher rewards than Simple (Fig.~\ref{fig:rl_best_model_performance_eisep}B, C), and other Dale-constrained baselines (Appendix~\ref{sec:appendix_lunarlander_baselines}), showing its stability advantage generalizes across different task settings and backbones. Notably, EISep also improves the performance of feedforward networks (Appendix~\ref{sec:appendix_eisep_ffn}), indicating that EISep can alleviate gradient instability not only from RNN iteration, but also from the RL task itself. We discuss the architectural effects on actor and critic in Appendix~\ref{sec:appendix_critic_actor_asymmetry}.


%% file: section/general_task/rl_dynamics.tex
\begin{figure}[t]
    \centering
\includegraphics[width=\linewidth]{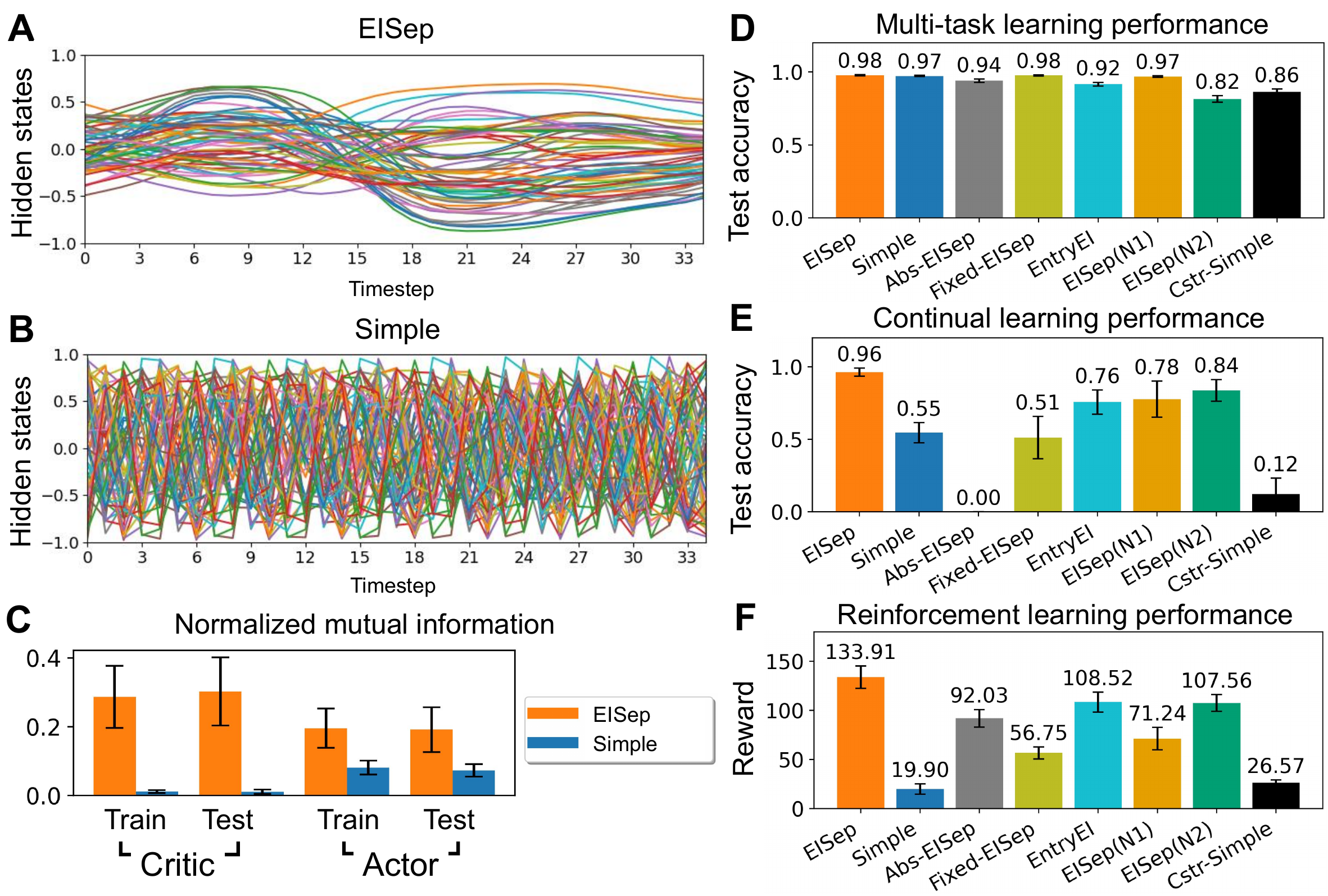}
    \vspace{-0.2in}
  \caption{Hidden dynamics and ablation analysis. \textbf{A.} EISep exhibits smooth and diverse hidden dynamics during Lunar Lander control. \textbf{B.} Simple shows high-frequency repetitive jittering, suggesting low-dimensional limit-cycle collapse. \textbf{C.} EISep achieves substantially higher normalized mutual information between HMM latent states and manually annotated environment phases, indicating that its hidden dynamics encode richer environment information. \textbf{D-F.} Across multi-task learning (\textbf{D}), continual learning (\textbf{E}), and reinforcement learning (\textbf{F}), EISep consistently matches or outperforms ablation models, highlighting the importance of initialization method, learnable signs, biologically grounded E-I separation, and the $\mathrm{ReLU}$ parameterization.}
    \vspace{-0.2in}
  \label{fig:hmm_dynamics_ablation}
\end{figure}

To examine how E-I separation shapes internal dynamics, we visualize one clip of the actor's hidden states from a test episode in Lunar Lander (Fig.~\ref{fig:hmm_dynamics_ablation}A-C). This clip is taken from EISep(A+C) and Simple trained with learning rate lr=0.005 for critics and lr=0.001 for actors, where both agents can achieve higher performance than chance (Table~\ref{tab:lunarlander_lr_sweep}). EISep exhibits smoother and more diverse hidden dynamics (Fig.~\ref{fig:hmm_dynamics_ablation}A), while Simple's dynamics show high-frequency and repetitive jittering (Fig.~\ref{fig:hmm_dynamics_ablation}B), suggesting a collapse to low-dimensional limit cycles. Longer clips are provided in Appendix~\ref{sec:appendix_hmm_details}.

To quantify how much internal dynamics encode environment information, we perform the following calculation: (1) fit the hidden trajectories using Hidden Markov Models (HMMs), and assign latent labels; (2) manually annotate phases using rule-based criteria (Appendix~\ref{sec:appendix_phase_detection}); (3) calculate the normalized mutual information (NMI) between unsupervised HMM labels and rule-based phases labels. EISep achieves substantially higher NMI than Simple for both actor and critic (Fig.~\ref{fig:hmm_dynamics_ablation}C), indicating that its latent dynamics are not only temporally smoother, but also more informative about the environment. In this way, EISep can support complex and robust control in challenging scenarios (Fig.~\ref{fig:rl_best_model_performance_eisep}), whereas Simple tends toward latent collapse and inferior performance.

%% file: section/ablation.tex
Finally, we present an ablation study to disentangle the contributions of individual design choices (Fig.~\ref{fig:hmm_dynamics_ablation}D-F). We briefly review each variant here and provide the details in Appendix~\ref{sec:appendix_ablation}. For multi-task learning, we report the best performance across learning rates. For continual learning, we report average performance in the final round when all four tasks are trained for the last time. For reinforcement learning, we report the best performance among variants on Lunar Lander. In Appendix~\ref{sec:appendix_sensitivity}, we test the EISep's sensitivity to hyperparameters including hidden size, temperature, entropy loss coefficient, activation function, and training order. In Appendix~\ref{sec:appendix_new_ablation}, we examine how ongoing sign plasticity affects performance and whether the performance can be explained by conventional sparsity regularization alone.

To assess the role of the $\mathrm{ReLU}$ operator (Eq.~\ref{eq:w=w0d}), we replace it with an absolute value operator: $W=|W_0|D$ and denote the new variant as \textbf{Abs-EISep}. To assess the benefit of learnable signs (Sec.~\ref{sec:background}), we fix the sign matrix $D$ after initialization, resulting in \textbf{Fixed-EISep}. To study the effect of initialization (Sec.~\ref{sec:background}), we apply ColEI-style (\textbf{EISep(N1)}) and DBPN-style (\textbf{EISep(N2)}) initialization to EISep (Appendix~\ref{sec:appendix_baseline_init}). We exclude DANN due to its fundamentally different parameterization with three coupled matrices (Appendix~\ref{sec:appendix_baseline_init}). To test if EISep's performance completely comes from its initialization method alone, we apply EISep's initialization method to Simple and denote the constrained-initialized version as \textbf{Cstr-Simple}. We summarize these models in Fig.~\ref{fig:ablation_schematic}.

Compared to EISep, these ablation models exhibit performance degradation to varying degrees, from minimal drops in simple multi-task learning (Fig.~\ref{fig:hmm_dynamics_ablation}D), to severe declines in the challenging RL task (Fig.~\ref{fig:hmm_dynamics_ablation}F). These results highlight the importance of the design choices in Sec.~\ref{sec:definition_eisep}, including learnable signs, initialization method and $\mathrm{ReLU}$ operator. It also suggests that EISep's success arises from their combined effect rather than any single component alone.


Another question is whether EISep's advantages arise from \emph{neuron-level} E-I separation induced by Dale’s principle, or more generally from value-sign separation. Therefore, we introduce \textbf{EntryEI}, which assigns an independent sign constraint to each weight: $W = \mathrm{ReLU}(W_0) \odot K, K_{ij} \in \{-1, 1\},$ where $K$ is a dense matrix sampled using Gumbel-Softmax reparameterization as in Sec.~\ref{sec:definition_eisep}, and $\odot$ denotes element-wise multiplication. This modification increases parameter size by allowing each weight to independently choose its sign. However, EntryEI underperforms EISep on all tasks (Fig.~\ref{fig:hmm_dynamics_ablation}D-F). This result highlights that a biologically realistic inductive bias can outperform a more flexible but less realistic alternative, even when the latter has a larger hypothesis space.

%% file: section/discussion.tex
We introduced EISep RNNs, a recurrent neural network architecture that enforces Dale’s principle through a learnable separation between excitatory and inhibitory neurons. Across a range of settings, including multi-task cognitive benchmark, continual learning, and reinforcement learning, EISep matches or exceeds the performance of Simple and other Dale-constrained variants, while exhibiting more stable optimization. Our analyses further reveal EISep induces brain-like sparse and modular structure, and promotes diverse, informative dynamics. These findings highlight a broader lesson for NeuroAI: biologically inspired inductive biases can simultaneously enhance performance, stability, and interpretability in artificial neural networks.

EISep also has several limitations that may motivate future work. First, EISep exhibits reduced performance at lower learning rates (Fig.~\ref{fig:cog_performance_stability}C left), which reveals a trade-off between stability and convergence speed. Future research can explore adaptive methods to expand its stable regime (Fig.~\ref{fig:appendix_multilr_baselines}). Although we have shown in Appendix~\ref{sec:appendix_multihidden} that EISep benefits more from scaling, it is still unclear whether EISep's advantages persist in even larger models (like LLM) and more challenging problems. Finally, there are many biological details that we do not attempt to reproduce in this work. For example, the ratio of excitatory to inhibitory neurons is typically around 4:1~\citep{wang2024we}, and long-range projections are predominantly excitatory~\citep{stepanyants2009fractions}. However, our goal is not to faithfully replicate the brain, but to investigate why biological systems are effective, by systematically testing biologically inspired inductive biases in ANNs. We expect future work along this direction to further bridge neuroscience and AI. Appendix~\ref{sec:appendix_neuroscience} discusses the implications for computational neuroscience and biological plausibility.

We would like to emphasize that our contribution is not limited to proposing a new RNN architecture, but also a systematic evaluation of biological inductive biases in biologically relevant learning scenarios, which may be challenging and underexplored in traditional RNN research. As shown in Appendix~\ref{sec:appendix_supervised_learning}, when tested on vision or algorithmic tasks where gradient instability or interference is not a significant issue, the advantage of EISep over Simple is much weaker, consistent with previous research. This result reveals an optimization–expressivity trade-off. Compared with vanilla RNN, EISep has many restrictions. (1) It constrains all outgoing connections from a neuron to share a sign. (2) It keeps the value matrix nonnegative by a ReLU filter. (3) It also has high sparsity that reduces the effective connections even further. These restrictions can act as dynamical regularization. In stationary, low-variance supervised problems, optimization is already relatively stable, so this additional regularization may provide little benefit while still limiting the capacity. However, the regularization becomes more beneficial in non-stationary or high-variance modules where gradient instability and interference are substantial. This also suggests that the value of biological inductive biases is inherently task-dependent, and systematic evaluation is essential to identify where these inductive biases provide the greatest benefits in machine learning.


%% file: section/appendix.tex
\newpage

\appendix

\section{Impact statement}\label{sec:appendix_impact_statement}
\input{section/appendix/impact_statement}

\section{Reproducibility statement}

\input{section/appendix/reproducibility}

\section{Use of large language models}\label{sec:appendix_llm_usage}
\input{section/appendix/llm_usage}

\section{Connection to neuroscience}\label{sec:appendix_neuroscience}

\input{section/appendix/neuroscience}

\section{Baseline initialization}\label{sec:appendix_baseline_init}

\input{section/appendix/baseline_init}

\section{Hyperparameter and training details}\label{sec:appendix_hyperparameter}
\input{section/appendix/hyperparameter}

\section{Hyperparameter sensitivity}\label{sec:appendix_sensitivity}

\input{section/appendix/sensitivity}

\section{Other ablation models}\label{sec:appendix_new_ablation}

\input{section/appendix/new_ablation}


\section{Environment introduction in the multi-task cognitive benchmark}\label{sec:appendix_env}

\input{section/appendix/environment_introduction}

\section{Training stability in the multi-learning-rate experiment}\label{sec:appendix_multilr}

\input{section/appendix/multilr}

\section{Modularity metric}\label{sec:appendix_modularity}

\input{section/appendix/modularity_q}

\section{Continual learning}\label{sec:appendix_continual_learning}

\input{section/appendix/continual_learning}
\section{Supervised learning}\label{sec:appendix_supervised_learning}

\input{section/appendix/supervised_learning}

\newpage
\section{Early convergence of neuron identities}\label{sec:appendix_logit_convergence}

\input{section/appendix/eilogit_convergence}

\section{Validity of the sparsity measure}\label{sec:appendix_sparsity_validity}

\input{section/appendix/sparsity_validity}


\section{Details of the RL performance}\label{sec:appendix_rl_performance_details}

\input{section/appendix/rl_performance_details}

\section{Details of the RL environment}\label{sec:appendix_phase_detection}

\input{section/appendix/phase_detection}

\section{Design of EISep-LSTM}\label{sec:appendix_eiseplstm}

\input{section/appendix/eisep_lstm}

\section{Sparsity of modules in the RL tasks}\label{sec:appendix_rl_sparsity}

\input{section/appendix/rl_sparsity}

\section{Details for HMM fitting}\label{sec:appendix_hmm_details}

\input{section/appendix/hmm_details}

\section{Details for ablation study}\label{sec:appendix_ablation}
\input{section/appendix/ablation}

\section{Critic actor asymmetry}\label{sec:appendix_critic_actor_asymmetry}

\input{section/appendix/critic_actor_asymmetry}

%% file: section/appendix/impact_statement.tex
\textbf{Applications.} This work contributes to foundational NeuroAI research by investigating how a core biological principle—neuron-level excitatory–inhibitory (E-I) separation—acts as an inductive bias in recurrent neural networks. These properties may benefit machine learning applications that rely on sequential reasoning and control, such as robotics, adaptive control systems, scientific modeling, and cognitive modeling, where stable training and interpretable dynamics are important. These properties may also support computational neuroscience for studying excitation–inhibition balance, by providing mechanistic frameworks that are both biologically realistic and computationally effective.

\textbf{Implications.} On the AI side, improved optimization stability and interpretability may increase the reliability of recurrent models for sequential decision-making, supporting more predictable and controllable behavior in downstream systems. On the neuroscience side, E–I–separated models offer mechanistic testbeds for studying neural computation and complementing experimental approaches. This work does not introduce new data, sensing, or surveillance capabilities and therefore does not raise immediate concerns related to privacy or fairness. However, as increased robustness can amplify both beneficial and harmful uses, the ultimate impact of such methods will depend on their deployment context.

\textbf{Initiatives.} On the AI side, future work could investigate how biologically inspired constraints interact with fairness, safety, and interpretability in applied settings, as well as how such constraints scale to larger or more complex models. On the neuroscience side, promising directions include integrating E–I–separated architectures with biophysical or spiking neuron models, comparing learned dynamics with neural recordings, and developing benchmarks for evaluating biological plausibility in machine learning systems. More broadly, this work supports an interdisciplinary research agenda in which insights from neuroscience inform machine learning design, promoting the interaction between artificial intelligence and brain science that benefits both fields and society.

%% file: section/appendix/reproducibility.tex
We release the full training and evaluation code in the supplementary material to support independent replication of our experiments. The appendix contains a complete specification of all hyperparameters, providing sufficient information to reproduce the reported results.

%% file: section/appendix/llm_usage.tex
In preparing this manuscript, we used large language models (LLMs) as a writing aid only, primarily to assist with text editing, organization, and clarity. Specifically, LLMs were used to help improve readability and flow, refine grammar and style, and suggest high-level organizational structures for sections of the paper.

All scientific contributions—including the formulation of the research questions, model design, experimental setup, implementation, analysis, and interpretation of results—were conceived, conducted, and validated entirely by the authors. LLMs were not used to generate original technical ideas, algorithms, code, experimental results, proofs, or numerical values. Any suggested references or rephrasings were independently verified by the authors before inclusion.

The authors take full responsibility for the content of the paper and include this statement solely for transparency regarding the limited use of generative AI tools.

%% file: section/appendix/neuroscience.tex
\subsection{Implications for computational neuroscience}

Here we talk about the neuroscience implications. We do not intend to argue that EISep should universally replace vanilla RNNs. Rather, the appropriate model depends on the scientific question.

In other RNN studies~\citep{yang2019task, driscoll2024flexible}, a unit in their networks can have both positive and negative outgoing weights. It cannot be directly interpreted as a biological neuron, but rather as a population of neurons. Therefore, vanilla RNNs are effective for identifying task representations and dynamical motifs in a neuron population without differentiation, but may need caution when asked what the roles of different cell types are in implementing a dynamical motif. Therefore, EISep enables the following analyses. (1) Learned circuits can be compared more directly with experimentally observed cell-type-specific connectivity and neural activity. (2) E/I-specific lesions and perturbations can be performed explicitly.

Fig.~\ref{fig:cog_case} provides a very simple example of this framework. EISep develops two stimulus-selective groups with within-group excitation and cross-group inhibition, revealing a winner-take-all circuit motif. We wish this network could help computational neuroscientists (1) understand the normative role of E/I separation better, and (2) develop cell-type-specific models to fit the neuronal population data for more powerful neuron signal modeling (like~\citep{ladenbauer2019inferring}).

The additional parameter cost is modest: EISep adds an $N\times2$ logit matrix, which scales as $O(N)$, whereas the recurrent matrix scales as $O(N^2)$. And from Fig.~\ref{fig:appendix_multihidden}, a medium EISep network ($n=64$) is often better than larger vanilla RNNs. Therefore, EISep helps us to save parameters rather than wasting them. Nevertheless, as we stress in the discussion and Appendix~\ref{sec:appendix_supervised_learning}, EISep’s constraints can be disadvantageous when E/I organization is unrelated to the scientific question. We therefore recommend using EISep when E/I-specific mechanisms are central to the hypothesis, while retaining vanilla RNNs as an important unconstrained baseline.

\subsection{Biological plausibility}

There remains a question regarding the biological plausibility of EISep. In related NeuroAI literature~\citep{yang2019task}, neural system development and machine learning training are roughly compared on the concept level. The neural network training may compress multiple biological processes, including evolutionary selection, developmental differentiation, and experience-dependent learning. Therefore, training EISep can be roughly interpreted as during an early "developmental" stage, the learnable signs allow the model to search over possible E/I organizations, while the entropy regularization encourages a stable and discrete organization. The "differentiated" neurons can then learn the task at a later "experience-driven learning" stage.

However, this analogy is only conceptual, and does not indicate that sign flips happen universally in biological neural systems, although there exist some similar phenomena. In fact, the effective sign of a neuron depends on both its transmitters and its postsynaptic environment. At an early period, the neurotransmitters could change (glutamate 
GABA~\citep{borodinsky2004activity}), and the postsynaptic environment could also be different (to make GABA excitatory~\citep{cherubini1991gaba}). Coexpression of excitatory and inhibitory transmitters also exists~\citep{yao2023high}. Our work is just a simplified conceptual abstraction. The real process may be more complicated, but our work provides a model for asking how task demands and circuit organization could jointly shape the eventual allocation of excitatory and inhibitory populations.

%% file: section/appendix/baseline_init.tex
In this part, we introduce some initialization details of different baselines models mentioned in Sec.~\ref{sec:background}.

\textbf{ColEI}~\citep{li2023learning} adopts an explicit value-sign separation $W=QD$. The initialization details are as follows.

Given an E/I ratio $r$, the numbers of excitatory neurons $n_E$ and inhibitory neurons $n_I$ are set to 
\begin{align}
    n_E=\frac{r}{r+1}N,\quad n_I=\frac{1}{r+1}N.
\end{align}
The rate parameters $\lambda_E, \lambda_I$ are chosen as 
\begin{align}
    \lambda_E&=1/\sqrt{\frac{2\pi-1}{2\pi}Nr},\\
    \lambda_I&=r\lambda_E
\end{align}
For the sampling process, we first sample weights $w$ for excitatory neurons $w^E$ and inhibitory neurons $w^I$
\begin{align}
    w^E_{i,j}&\sim\mathrm{Exp}(\lambda_E),\\
    w^I_{i,j}&\sim\mathrm{Exp}(\lambda_I).
\end{align}
Then the diagonal entries are set to zero: $w_{ii}=0$. The next step is to renormalize the spectral radius to 1.5: 
\begin{align}
    w=w*1.5/\max(|k|),
\end{align}
where $k$ is the eigenvalue of $w$. Finally an absolute-value operation is applied to obtain the value matrix $Q=|w|$. 

The sign matrix satisfies $D_{ii}^E=1$ for excitatory neurons, and $D_{jj}^I=-1$ for inhibitory neurons. The final recurrent weight is constructed as 
\begin{align}
    W=QD.
\end{align}
Non-negativity of $Q$ is enforced by clipping $Q\leftarrow\mathrm{ReLU}(Q)$ at every update.

\textbf{DBPN}~\citep{balwani2025constructing} enforces Dale’s principle through projection-based updates rather than an explicit sign matrix. During training, outgoing weights of excitatory neurons are projected to zero whenever they become negative (and analogously for inhibitory neurons). Although no explicit $D$ is used, their approach is effectively equivalent to a value–sign decomposition in which the value matrix is clipped to non-negative $Q\leftarrow\mathrm{ReLU}(Q)$ at every update, with an implicit fixed sign matrix $D$. 

The initial weights are sampled from a uniform distribution:
\begin{align}
    Q_{i,j}\sim\mathcal{U}(0,1/\sqrt{N})
\end{align}

\textbf{DANN}~\citep{cornford2020learning} enforces Dale's principle by explicitly separating excitatory and inhibitory pathways rather than factorizing a single recurrent matrix. The recurrent weight is parameterized as $W=W^{EE}-W^{EI}W^{IE}$. Given the hidden size $N$, the initialization adopts the following rules: 
\begin{align}
    W^{EE}_{i,j}&\sim \mathrm{Exp}(\frac{1}{\sqrt{3(N-1)}}),\\
    W^{IE}_{i,j}&=\mathrm{mean}(W^{EE}_{i,j}),\\
    W^{EI}_{i,j}&\sim \mathrm{Exp}(\sqrt{\frac{10}{3(N-1)}}).
\end{align}

All three matrices are constrained to be non-negative via clipping after each update. 

%% file: section/appendix/hyperparameter.tex
All recurrent networks use $N=64$ hidden units and are trained with backpropagation through time (BPTT).

\subsection{Gumbel-Softmax sampling}\label{sec:appendix_gumbel_softmax_sampling}

For the soft Gumbel-Softmax used in Sec.~\ref{sec:definition_eisep}, let $u_{i,k}\sim\mathcal{U}(0,1)$ independently and $g_{i,k}=-\log(-\log u_{i,k})$. Then
\begin{align}
 m_{i,k}&=\frac{\exp((M_{i,k}+g_{i,k})/\tau)}{\sum_{\ell=0}^{1}\exp((M_{i,\ell}+g_{i,\ell})/\tau)},\\
 e_i&=m_{i,0}-m_{i,1}=\tanh\!\left(\frac{(M_{i,0}+g_{i,0})-(M_{i,1}+g_{i,1})}{2\tau}\right).
\end{align}
With \texttt{hard=False}, $e_i$ is in $(-1,1)$ during training. Temperature $\tau=0.1$ is fixed without annealing, and the entropy penalty encourages concentrated identities. Intermediate values occur when the two perturbed logits are close. At test time, the larger unperturbed logit determines the discrete sign, with ties assigned $-1$ as in Sec.~\ref{sec:definition_eisep}.

Even if a soft value $e_i$ is not exactly $\pm1$ and could change during training, it does not violate Dale’s principle. Because $W=\text{ReLU}(W_0)D$, all outgoing weights of neuron $i$ are the results of non-negative values multiplied by the same sign scalar $e_i$. They therefore remain either all non-negative or all non-positive in every forward pass. 

\subsection{Multi-task Cognitive Benchmark}
\label{sec:appendix_hyperparameter_cog}

\paragraph{Task setup.}
For each architecture, a single recurrent network is trained to perform all 20 cognitive tasks using supervised learning, with the ground-truth choice as the training target. We run 10 random seeds and report mean and standard error ($\textrm{SEM}=\sigma/\sqrt{N}$) across seeds. Models are tested for 100 epochs after training.

\paragraph{Network initialization.}
\begin{itemize}
    \item Activation function: Softplus.
    \item Recurrent bias: $b_{W,i} \sim \mathcal{N}(0, 1/N)$.
    \item Input and output weights ($U, V$): orthogonal initialization; biases ($b_U, b_V$) $\sim \mathcal{N}(0, 1/N)$.
    \item E–I logit matrix: $M_{i,j} \sim \mathcal{N}(0, 1/N)$.
    \item Gumbel-Softmax temperature: $0.1$ (no annealing).
\end{itemize}

\paragraph{Optimization.}
\begin{itemize}
    \item Sequence length: 100.
    \item Batch size: 32.
    \item Training epochs: 3000.
    \item Optimizer: Adam.
    \item Loss: cross-entropy.
    \item Regularization: none, except E–I logit entropy loss (Sec.~\ref{sec:definition_eisep}) with coefficient $\lambda_e=1$.
    \item Output weighting: fixation output weight $0.05$, decision outputs weight $1$ (following the original benchmark).
\end{itemize}

\paragraph{Multi–learning-rate sweeps.}
For learning-rate stability experiments (e.g., Fig.~\ref{fig:cog_performance_stability}C), we use 5 seeds and sweep learning rates:
\[
\{10^{-4}, 5\times10^{-4}, 10^{-3}, 5\times10^{-3}, 10^{-2}, 5\times10^{-2}, 10^{-1}, 0.5\}.
\]

\subsection{Continual Learning}
\label{sec:appendix_hyperparameter_continual_learning}

\paragraph{Task protocol.}
We use 5 random seeds. In one complete round, each of the four tasks is trained for 200 epochs in the order \textit{Go - Anti - Delay Go - Delay Anti}. The full task round is repeated 10 times, resulting in 40 task trials. The task rule is provided at every trial to avoid ambiguity. After training on each task, models are tested for 200 epochs on all four tasks. We report the standard error ($\textrm{SEM}=\sigma/\sqrt{N}$) across seeds on the plot as the error bar.

\paragraph{Optimization details.}
All network architecture and optimization settings follow Sec.~\ref{sec:appendix_hyperparameter_cog}.

\subsection{Reinforcement Learning}

\paragraph{Environments.}
We use \texttt{LunarLander-v3} and \texttt{HalfCheetah-v5} from Gymnasium v1.1.1. Maximum episode length is 500 steps for LunarLander and unbounded for HalfCheetah.

\paragraph{Data collection.}
\begin{itemize}
    \item Parallel environments: $n_p=10$ (LunarLander), $n_p=100$ (HalfCheetah).
    \item Steps per update: $n_s=128$ per environment.
    \item Number of updates: $n_u=1000.$
    \item Batch size: $n_s \times n_p$.
\end{itemize}

\paragraph{Optimization and losses.}
\begin{itemize}
    \item Advantage estimation: Generalized Advantage Estimation (GAE).
    \item Critic loss: Smooth L1 loss $L_1^s(V_t, [A_t + V_t])$, where $A_t$ is the advantage term and $[\cdot]$ indicates the gradient of $A_t+V_t$ is detached from the computational graph.
    \item Actor loss: advantage-normalized policy gradient, which means the advantage $A_t$ is normalized in each batch.
    \item Separate optimizers and learning rates for actor and critic.
    \item Regularization: none except E–I logit entropy loss with coefficient $\lambda_e=1$.
\end{itemize}

\paragraph{RL hyperparameters.}
\begin{itemize}
    \item Discount factor: $\gamma = 0.99$.
    \item GAE parameter: $\lambda = 0.95$.
    \item Entropy coefficient: $0.01$.
    \item The ranges of learning rates of actor:
    $[3\times10^{-4}, 1\times10^{-3}, 3\times10^{-3}]$
    \item The ranges of learning rates of critic:
    $[1.5\times10^{-3}, 5\times10^{-3}, 1.5\times10^{-2}]$

\end{itemize}

\paragraph{Evaluation.}
We use 5 random seeds and evaluate each model on 20 episodes per seed. We report the standard error across episodes and seeds ($\textrm{SEM}=\sigma/\sqrt{N}$) on the plot as the error bar. 

\subsection{Computation resource and execution time}\label{sec:appendix_compute_resource}

\paragraph{Hardware.}
All experiments were conducted on a workstation equipped with an AMD Ryzen 7 7840H CPU, 16 GB RAM, and an NVIDIA GeForce RTX 4060 Laptop GPU (8 GB VRAM). The GPU driver version is 566.14 with CUDA 12.7 support.

\paragraph{Software.}
Experiments were implemented in PyTorch with standard CUDA acceleration. Full environment details are provided in the Appendix.

\paragraph{Per-experiment cost.}
Multi-task learning: \~1 hour for 5 seeds (\~12 minutes per run)
Continual learning: \~1.5 hours for 5 seeds (\~18 minutes per run)
Reinforcement learning: \~10 hours for 5 seeds, 9 learning rates, and 3 variants, including EI(A), EI(C), EI(A+C) (\~6 minutes per run)

\paragraph{Total compute.}
Across all experiments, including hyperparameter tuning and ablations, the total compute is estimated to be on the order of tens of GPU-hours.

\paragraph{Additional details.}
All models were trained from scratch without pretraining. Experiments can be reproduced on a single consumer-grade GPU with similar memory capacity.

\subsection{Licenses for External Assets}\label{sec:appendix_license}

We use standard benchmark environments and datasets, including:

\begin{itemize}

\item \textbf{Neurogym cognitive task suite}~\cite{yang2019task}  
License: Apache License 2.0.  
Copyright 2020--2026, Manuel Molano, Guangyu Robert Yang, Jorge F. Mejias, and contributors; 
additional contributions from Giulia Crocioni et al.  
URL: \url{https://github.com/neurogym/neurogym}

\item \textbf{Gymnasium environments (e.g., LunarLander, HalfCheetah)}~\cite{towers2024gymnasium}  
License: MIT License.  
Copyright 2016 OpenAI; Copyright 2022 Farama Foundation.  
URL: \url{https://github.com/Farama-Foundation/Gymnasium}

\item \textbf{MuJoCo (for HalfCheetah)}  ~\cite{wawrzynski2009cat}  
License: Source code is licensed under the Apache License 2.0.  
Copyright 2021 DeepMind Technologies Limited; 
additional components Copyright 2016 Svetoslav Kolev.  
Documentation is licensed under CC BY 4.0.  
URL: \url{https://github.com/google-deepmind/mujoco}

\end{itemize}

All assets are used in accordance with their respective licenses. 
No datasets containing personal or sensitive information are used.

%% file: section/appendix/sensitivity.tex
\subsection{Hidden size}\label{sec:appendix_multihidden}

\begin{figure}{}
    \centering
\includegraphics[width=0.6\linewidth]{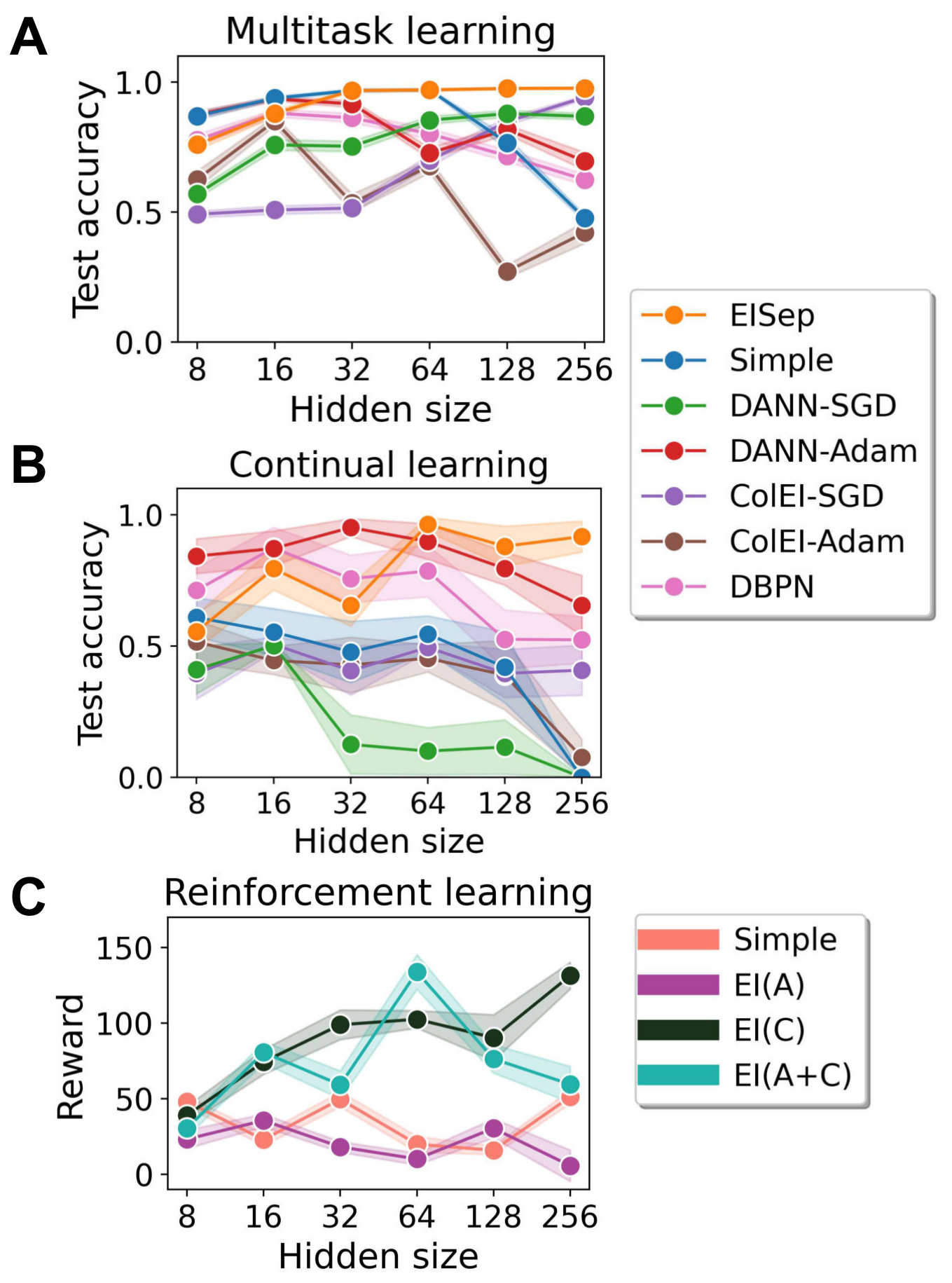}
  \caption{Under different hidden sizes, the performance in multi-task learning (\textbf{A}), continual learning (\textbf{B}) and reinforcement learning (\textbf{C}). }.
    \vspace{-0.1in}
  \label{fig:appendix_multihidden}
\end{figure}

To test the sensitivity of different models to the hidden size, we evaluated model performance across hidden sizes (Fig.~\ref{fig:appendix_multihidden}). Compared to the default setting ($64$), EISep remains robust as the hidden size increases, while other models tend to overfit and degrade at larger sizes. Moreover, the performance gap between EISep and Simple widens with increasing model size, indicating that EISep benefits more from scaling.

\subsection{Sampling temperature}\label{sec:appendix_multitemp}

To test the sensitivity of EISep to the temperature parameter in Gumbel-Softmax sampling, we evaluated model performance across a range of temperatures. Compared to the baseline temperature ($\tau=0.1$) used in the main text, decreasing the temperature leads to a minor performance drop, likely because lower temperatures push the Gumbel-Softmax output toward a step function, causing gradients to vanish and the E/I mask to become effectively frozen — approximating the behavior of \textbf{Fixed-EISep} (Section~\ref{sec:result_ablation}). Conversely, increasing the temperature substantially impairs performance in multi-task learning and continual learning, as the soft E/I assignments become dominated by Gumbel noise, causing frequent sign flips across training steps and preventing stable weight updates from accumulating. Furthermore, high temperatures introduce a train-test mismatch: the sign assignments at test time may differ from those seen during training, leading to inconsistent network behavior.

\begin{figure}{}
    \centering
\includegraphics[width=0.6\linewidth]{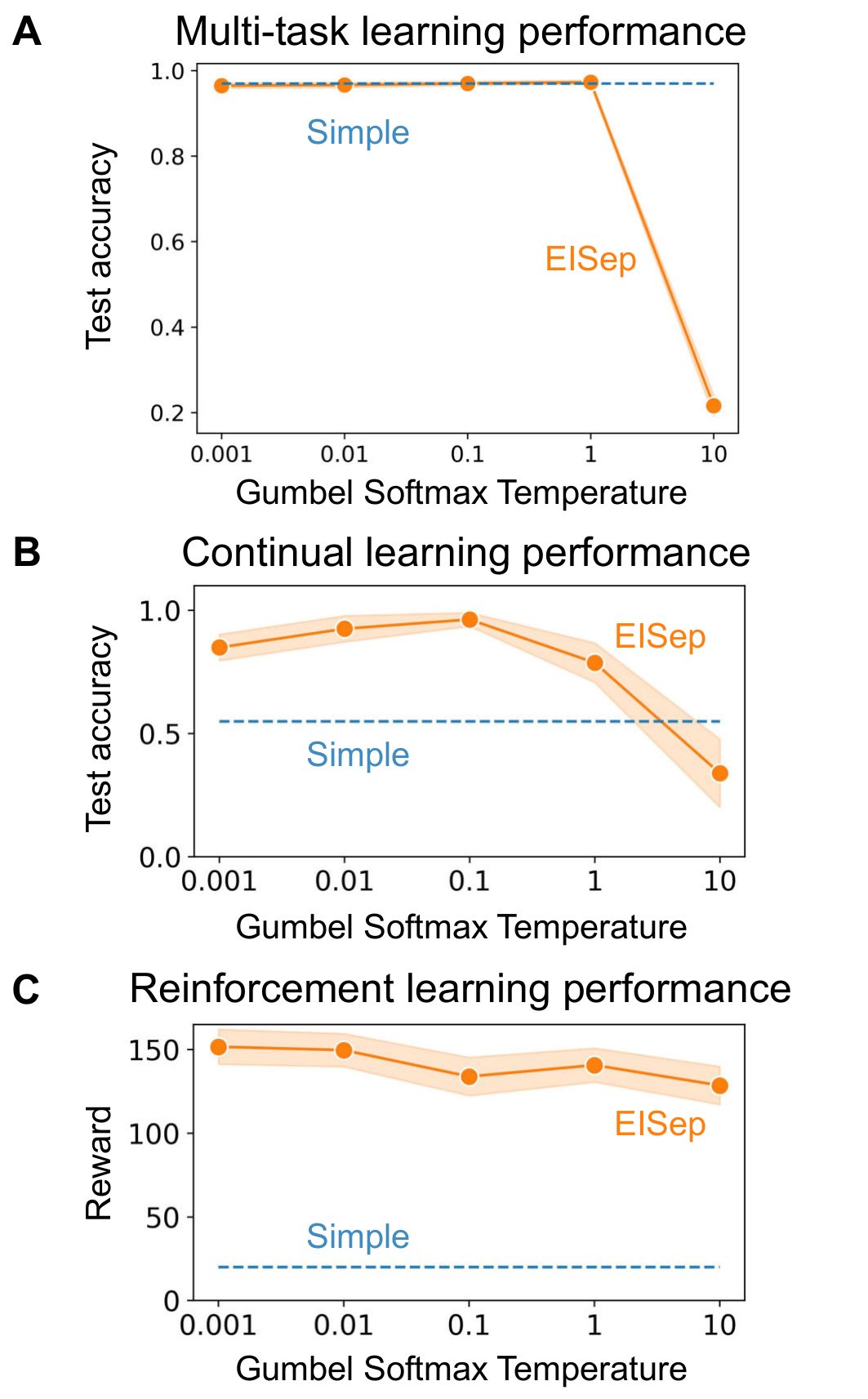}
  \caption{Under different temperature for Gumbel-Softmax sampling, the performance in multi-task learning (\textbf{A}), continual learning (\textbf{B}), and reinforcement learning (\textbf{C}).}
    \vspace{-0.1in}
  \label{fig:appendix_multitemp}
\end{figure}

\subsection{Entropy loss coefficient}\label{sec:appendix_multieie}

\begin{figure}{}
    \centering
\includegraphics[width=0.5\linewidth]{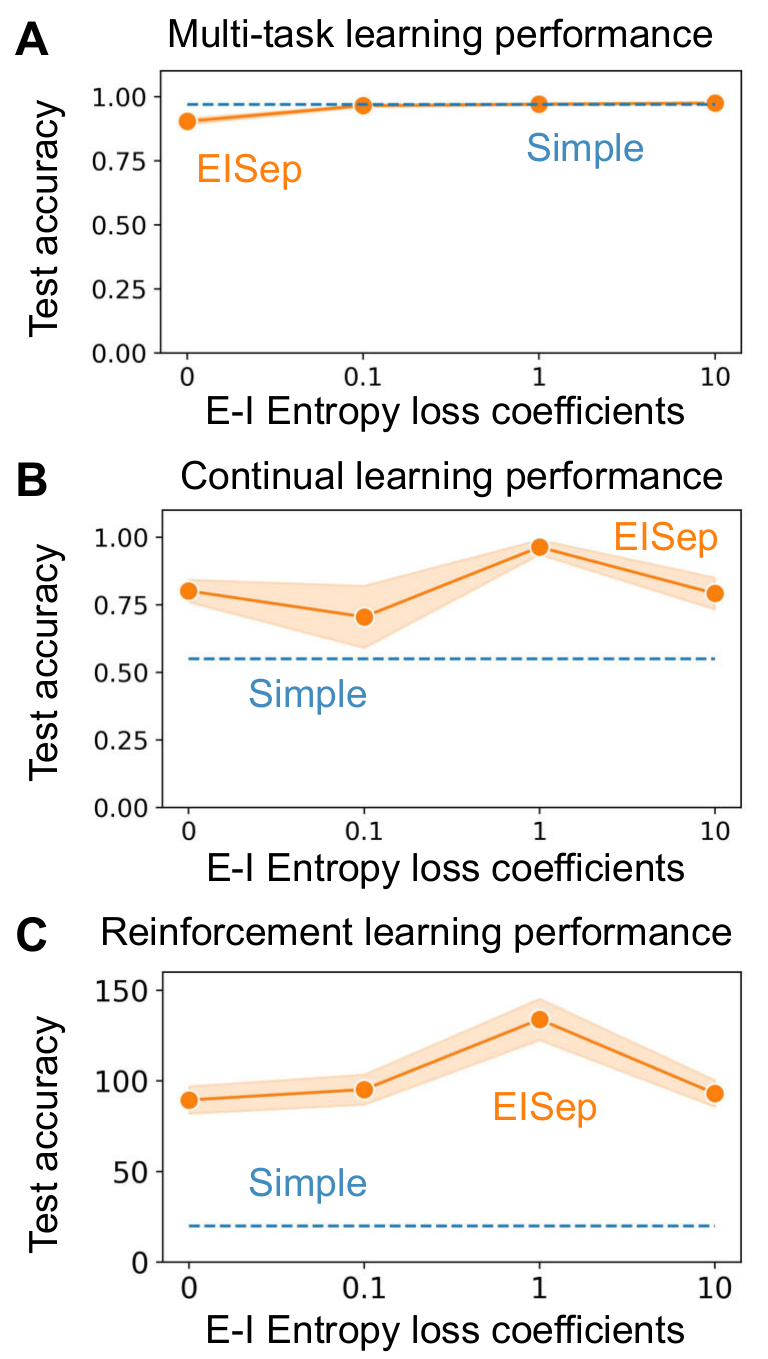}
  \caption{Under different E-I logit entropy loss coefficients, the performance in multi-task learning (\textbf{A}), continual learning (\textbf{B}) and reinforcement learning (\textbf{C})).}
    \vspace{-0.1in}
  \label{fig:appendix_multieie}
\end{figure}

To test the sensitivity of EISep to the entropy loss coefficient $\lambda_e$ (Sec.~\ref{sec:definition_eisep}), we evaluated model performance across a range of coefficients. Compared to the default setting ($\lambda_e=1$) used in the main text, removing the entropy loss ($\lambda_e=0$) significantly degrades performance (Fig.~\ref{fig:appendix_multieie}), as the non-convergent signs will have occasional flips that cause train-test mismatch. On the other hand, moderate changes to $\lambda_e$ have little effect on relatively simple multi-task learning, but noticeably impair performance in continual learning and reinforcement learning.

\subsection{Activation function}\label{sec:appendix_activation}

\begin{figure}{}
    \centering
\includegraphics[width=0.6\linewidth]{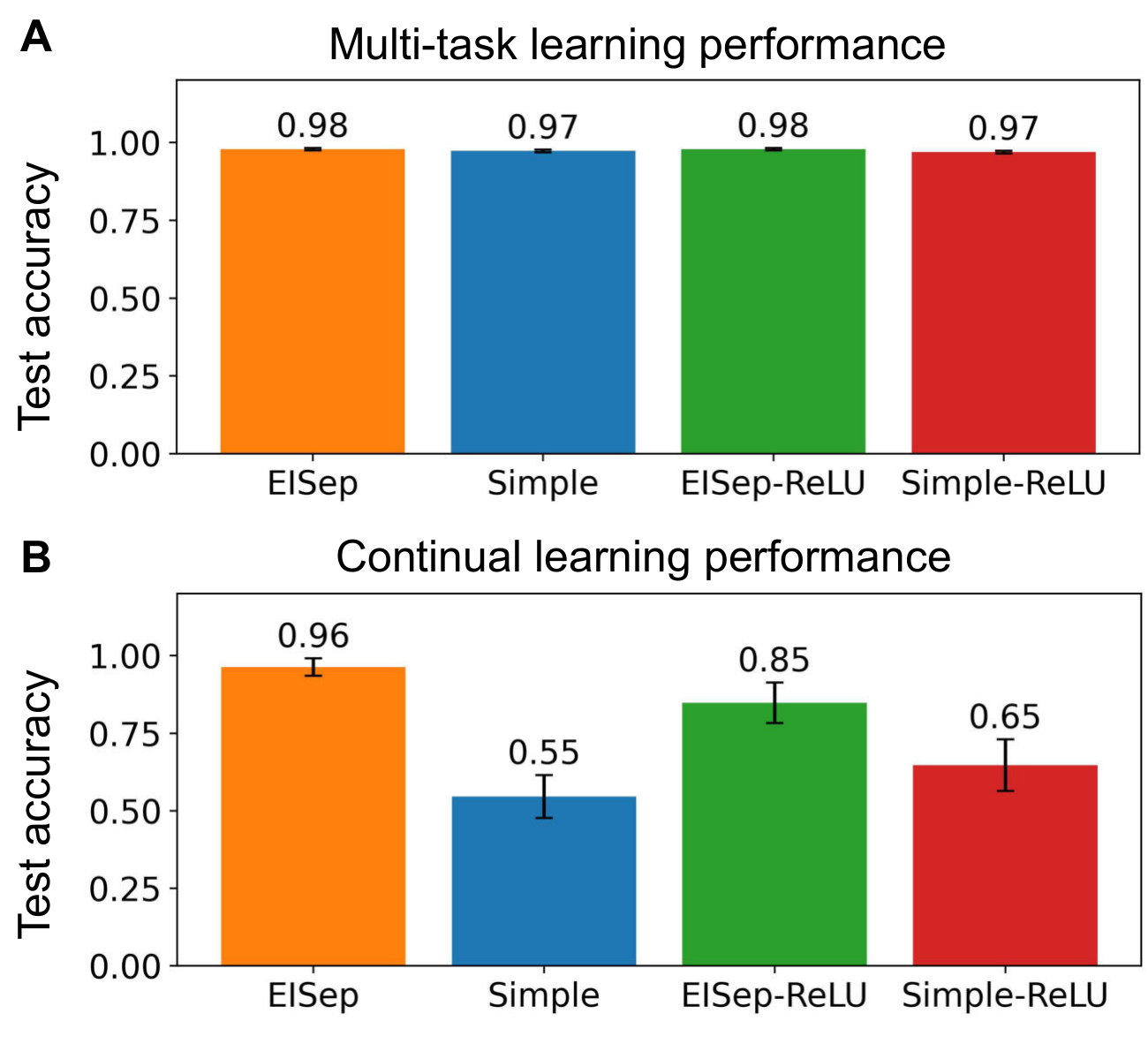}
  \caption{Under activation functions, the performance in multi-task learning (\textbf{A}) and continual learning (\textbf{B}).}
  \label{fig:appendix_activation_function}
\end{figure}

To test the sensitivity of EISep to the activation function, we evaluate model performance using $\mathrm{ReLU}$ ((Fig.~\ref{fig:appendix_activation_function})). Compared to the default setting ($\mathrm{Softplus}$), the new activation function causes a slight drop in continual learning performance but does not affect multi-task learning, and the advantage of EISep to Simple still exists. In reinforcement learning, however, training becomes unstable due to persistent gradient explosions, and the training is not able to converge.

\subsection{Initialization method}\label{sec:appendix_initialization_method}

To test the fairness of the initialization method, we adapt EISep's initialization method and apply it to all other baselines. All baselines show worse performance than EISep (Table~\ref{tab:initialization_method_family}). This result also shows EISep’s performance cannot be naively explained by its initialization method.

\begin{table}[htbp]
\centering
\small
\setlength{\tabcolsep}{8pt}
\caption{Performance under default and EISep-like initialization.}
\label{tab:initialization_method_family}
\begin{tabular}{lcc}
\toprule
Model & Default initialization & EISep-like initialization \\
\midrule
EISep      & \textbf{$0.9699 \pm 0.0025$} & \textbf{$0.9699 \pm 0.0025$} \\
Simple     & \textbf{$0.9654 \pm 0.0089$} & \textbf{$0.9705 \pm 0.0024$} \\
DANN-SGD   & $0.8344 \pm 0.0215$ & $0.6276 \pm 0.0217$ \\
DANN-Adam  & $0.7503 \pm 0.0558$ & $0.6528 \pm 0.0523$ \\
ColEI-SGD  & $0.7095 \pm 0.0545$ & $0.7065 \pm 0.0397$ \\
ColEI-Adam & $0.6108 \pm 0.0918$ & $0.6230 \pm 0.0980$ \\
DBPN       & $0.8055 \pm 0.0405$ & $0.8581 \pm 0.0115$ \\
\bottomrule
\end{tabular}
\end{table}

\subsection{Spectral radius}\label{sec:appendix_spectral_radius_control}
To test the fairness of the spectral radius, we rescaled the recurrent weight matrix of each model so that all models had the same initial spectral radius as EISep, while leaving their respective parameterizations unchanged. Under this matched initialization, EISep still achieves the highest average performance. Notably, DANN-SGD and ColEI-SGD have smaller final spectral radii than EISep but substantially lower performance, indicating that simply minimizing the spectral radius may not necessarily lead to good task performance. These results show that EISep’s advantage cannot be attributed solely to its initial spectral radius. Its parameterization also shapes how the recurrent dynamics evolve during training.

\begin{table}[htbp]
\centering
\small
\setlength{\tabcolsep}{8pt}
\caption{
Control with matched initial spectral radius.
All models have the same initial spectral radius,
$0.6715 \pm 0.0093$, measured across all 5 seeds.
Final spectral radius and performance are reported as mean $\pm$ SEM
over valid terminal runs: 5 seeds for every model except
ColEI-Adam (3 valid seeds).
}
\label{tab:matched_spectral_radius}
\begin{tabular}{lcc}
\toprule
Model & Final spectral radius & Accuracy \\
\midrule
EISep     & $2.7762 \pm 0.1374$  & \textbf{$0.9745 \pm 0.0024$} \\
Simple    & $6.9175 \pm 0.5926$  & $0.9705 \pm 0.0024$ \\
DANN-SGD  & $1.0200 \pm 0.0210$  & $0.8539 \pm 0.0118$ \\
DANN-Adam & $14.4388 \pm 4.4922$ & $0.8203 \pm 0.0355$ \\
ColEI-SGD & $0.9769 \pm 0.0281$  & $0.6734 \pm 0.0618$ \\
ColEI-Adam & $5.8978 \pm 0.7504$ & $0.7823 \pm 0.1355$ \\
DBPN      & $5.1758 \pm 0.6728$  & $0.8556 \pm 0.0230$ \\
\bottomrule
\end{tabular}
\end{table}

\subsection{Training order}\label{sec:appendix_training_order}

\begin{figure}{}
    \centering
\includegraphics[width=0.8\linewidth]{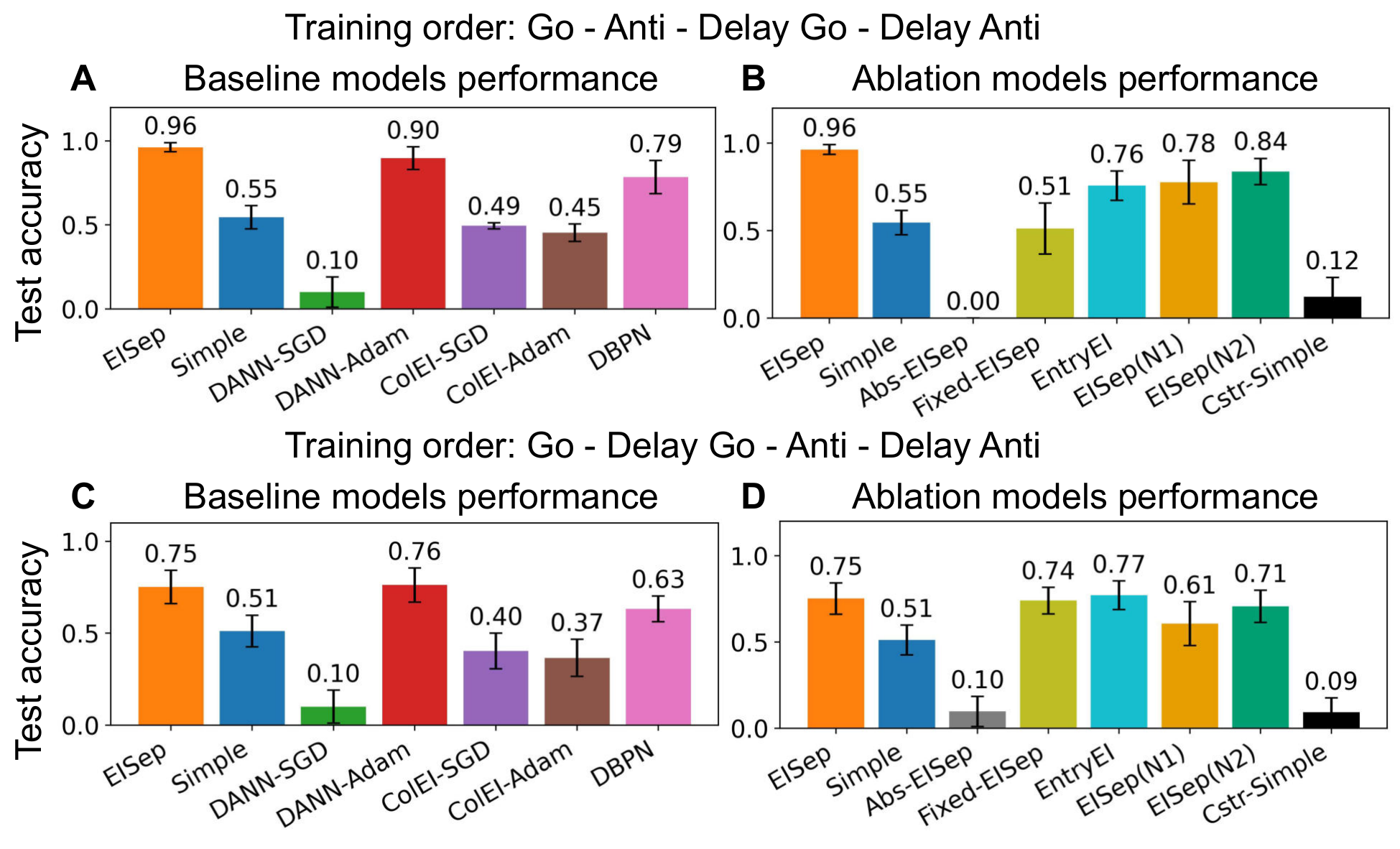}
  \caption{Under different training orders, the performance in continual learning for all baseline models (\textbf{A, C}) and ablation models (\textbf{B, D}).}
  \label{fig:appendix_training_order}
\end{figure}

Training order is known to affect continual learning dynamics~\citep{duncker2020organizing}. We vary the task sequence and find that the “grouped” order ($\textit{Go} \rightarrow \textit{Delay Go} \rightarrow \textit{Anti} \rightarrow \textit{Delay Anti}$) performs worse than the “alternating” order ($\textit{Go} \rightarrow \textit{Anti} \rightarrow \textit{Delay Go} \rightarrow \textit{Delay Anti}$) used in the main experiments (Fig.~\ref{fig:appendix_training_order}), although EISep can still maintain its own advantage over other models. We attribute this to increased gradient interference: training two similar tasks consecutively drives the network toward a narrow subspace, making subsequent adaptation to opposing stimulus–response mappings more difficult. In contrast, the alternating order interleaves conflicting tasks, encouraging a more balanced representation and enabling the network to converge to a solution that better accommodates all tasks.

%% file: section/appendix/new_ablation.tex
Here, we introduce two new ablation models that are not presented in the main text.

\subsection{The ongoing learnability of signs}

In Sec.~\ref{sec:result_ablation}, we show that Fixed-EISep underperforms EISep, highlighting the importance of learnable signs. However, it remains unclear whether this benefit arises primarily from discovering a good sign configuration, or from maintaining sign adaptability throughout training. To disentangle these effects, we introduce \textbf{EISep-ReInit}, which initializes the sign matrix using a pretrained EISep model, while reinitializing and optimizing all other parameters as in standard training.

\begin{figure}{}
    \centering
\includegraphics[width=0.9\linewidth]{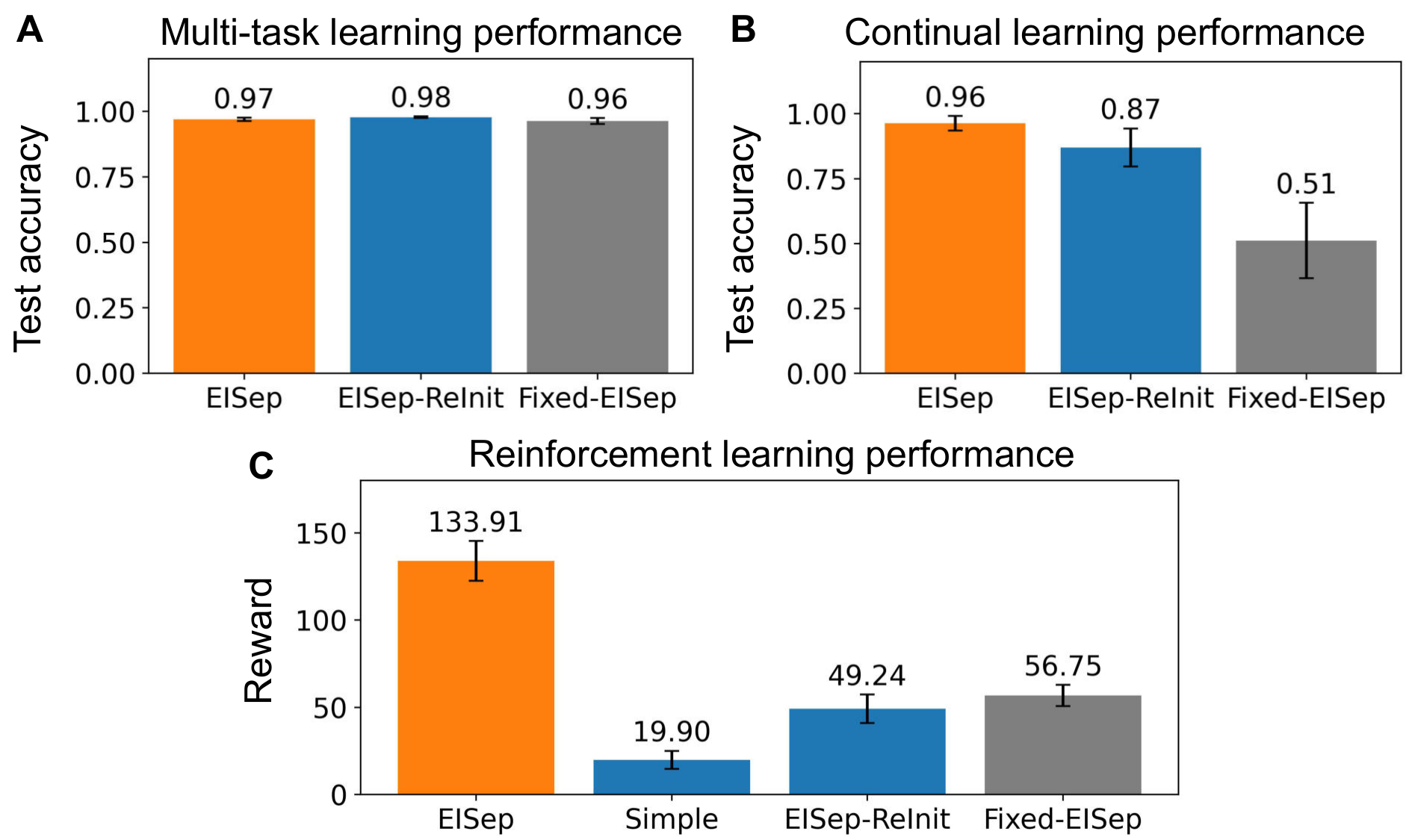}
  \caption{The performance of EISep-ReInit in multi-task learning (\textbf{A}), continual learning (\textbf{B})} and reinforcement learning (\textbf{C}).
    \vspace{-0.1in}
  \label{fig:appendix_reinit}
\end{figure}

The result is in Fig.~\ref{fig:appendix_reinit}. On multi-task learning, EISep and EISep-ReInit perform similarly, suggesting that identifying a suitable sign configuration is largely sufficient in simpler settings. In contrast, on continual learning, EISep-ReInit underperforms EISep but still surpasses Fixed-EISep, indicating that both a good initialization and ongoing sign adaptability contribute to mitigating interference. In reinforcement learning, EISep-ReInit performs comparably to Fixed-EISep and significantly worse than EISep, suggesting that continual adaptation of signs is critical for handling non-stationarity and stabilizing high-variance gradient updates. Together, these results indicate that learnable signs not only provide a good initial inductive bias, but also act as a dynamic mechanism for reorganizing computation during training.

\begin{figure}{}
    \centering
\includegraphics[width=0.5\linewidth]{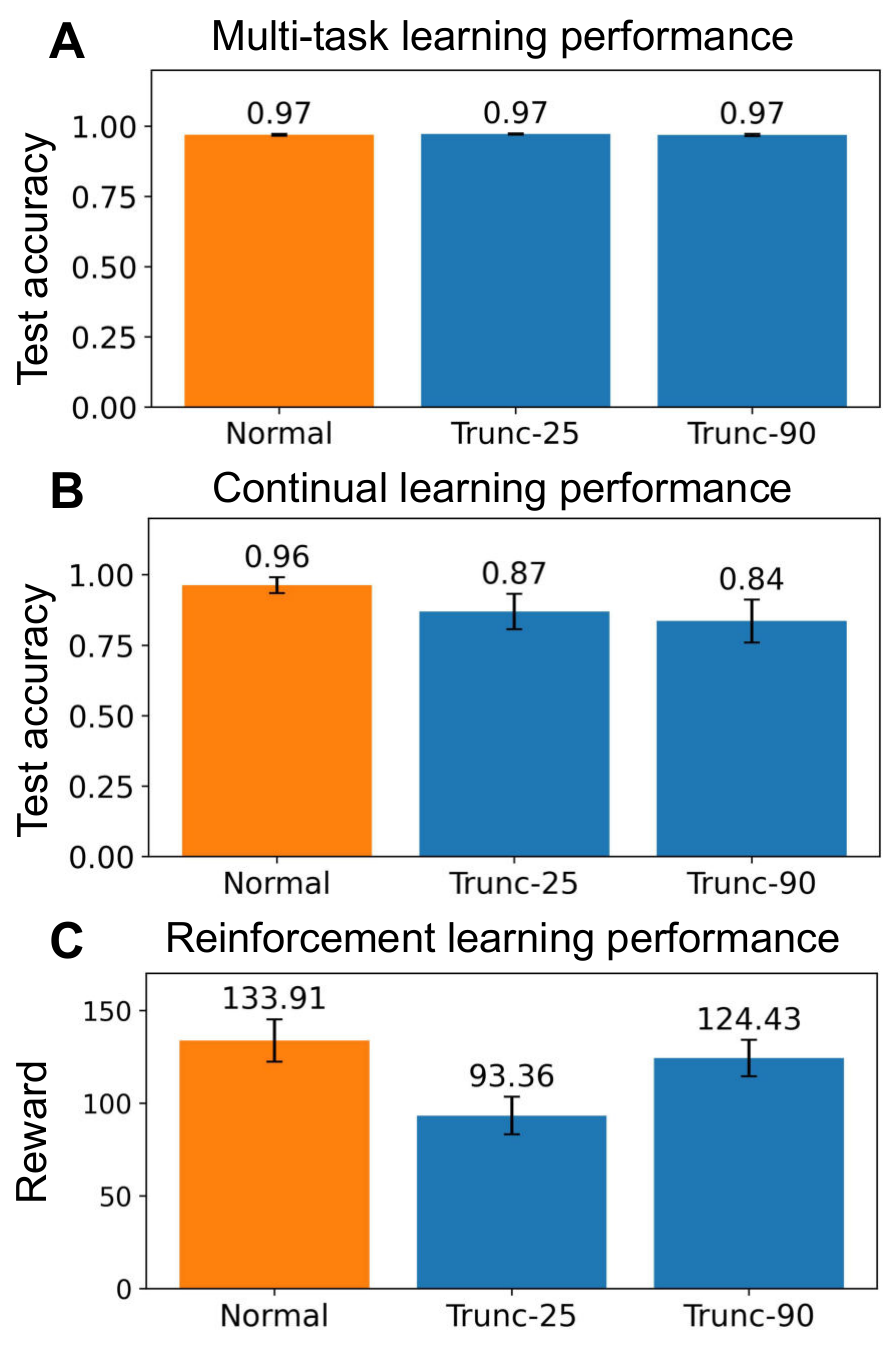}
  \caption{Under different truncation proportions, the performance in multi-task learning (\textbf{A}), continual learning (\textbf{B}), and reinforcement learning (\textbf{C})}.
    \vspace{-0.1in}
  \label{fig:appendix_truncation}
\end{figure}

We further investigate whether this adaptability is required throughout training or primarily during early stages. To this end, we introduce \textbf{Trunc-X}, where after X\% of training, the sign matrix is fixed and the entropy regularization is disabled (Fig.~\ref{fig:appendix_truncation}). While multi-task learning remains largely unaffected, both continual learning and reinforcement learning exhibit clear performance degradation compared to EISep. This suggests that ongoing sign plasticity helps the network adapt to shifting task demands and maintain stable dynamics under non-stationary optimization.

\subsection{Simple with L1 regularization}

\begin{figure}{}
    \centering
\includegraphics[width=0.9\linewidth]{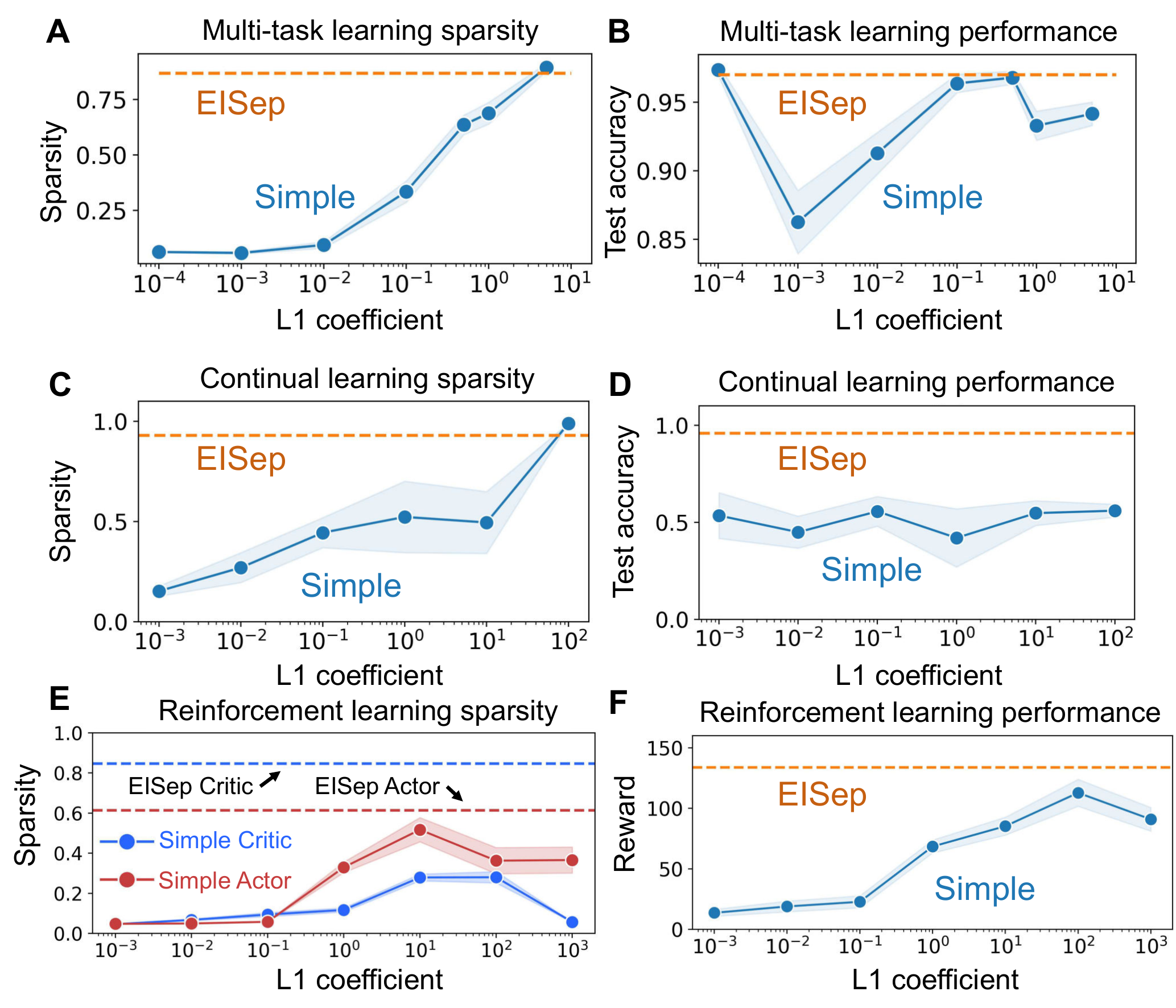}
  \caption{Under different L1 coefficients, the sparsity and performance in multi-task learning (\textbf{A, B}), continual learning (\textbf{C, D}), and reinforcement learning (\textbf{E, F}). In RL, EISep(A+C) is compared with Simple, whose actor and critic are both L1-regularized with the same coefficients.}
    \vspace{-0.1in}
  \label{fig:appendix_multil1}
\end{figure}

As mentioned in Sec.~\ref{sec:discussion}, EISep faces the trade-off between stability and convergence speed, while achieving much higher sparsity than Simple (Sec.~\ref{sec:result_sparsity}). This raises the question of whether EISep is simply a regularized version of Simple. To test this, we apply L1 regularization to Simple and vary the regularization coefficients. Although large L1 coefficients can induce sparsity comparable to EISep (Fig.~\ref{fig:appendix_multil1}A, C), performance remains inferior in continual learning, and becomes unstable in multi-task learning (Fig.~\ref{fig:appendix_multil1}B, D). In reinforcement learning, even extremely large L1 coefficients fail to achieve comparable sparsity and yield inferior performance, indicating a mismatch between L1-induced sparsity and effective optimization. 

These results suggest that EISep provides more than a regularization effect, instead acting as a structured, biologically inspired inductive bias that promotes both efficient connectivity and stable learning.

%% file: section/appendix/environment_introduction.tex
We adopt the cognitive task environments introduced in \cite{yang2019task}, comprising a total of 20 tasks with a shared underlying structure. Here we adapt and simplify the task introductions to help readers understand the environment quickly and clearly. Unless otherwise specified, all tasks follow the same trial format, input--output conventions, and noise standard deviation ($\sigma_n=1.0$). Task-specific features are described in the corresponding subsections.

\subsection{Common Task Structure}

Each trial is associated with a one-hot {rule input} corresponding to the current task, which remains active throughout the entire trial and specifies the task context. The network also receives a \emph{fixation input} that is active from trial onset. While the fixation input is on, the network is required to maintain fixation by activating the fixation output unit. The offset of the fixation input typically signals the onset of the response period, during which the network must report a response direction via a population of directional output units arranged on a ring. In this paper, only two directions (left vs. right) are used. 

The network receives one or two stimuli per trial. Each stimulus conveys directional information and may appear in modality~1, modality~2, or both. A trial is considered correct only if the network maintains fixation when required and produces the correct directional response during the go period.

Each task can be segmented into distinct periods: a fixation period (\textbf{fix}) preceding stimulus presentation, followed by stimulus period~1 (\textbf{stim1}). If two stimuli are presented sequentially, they are separated by a delay period, and the second stimulus is shown during stimulus period~2 (\textbf{stim2}). 

For clarity, the 20 tasks are grouped into five task families: Go, Anti, Decision Making (DM), Delayed Decision Making (Dly DM), and Matching.

\subsection{Go Task Family}

The Go family includes the \textbf{Go}, \textbf{Reaction-Time Go (RT Go)}, and \textbf{Delayed Go (Dly Go)} tasks. In all three tasks, a single stimulus is presented in either modality~1 or modality~2, and the correct response is to report the stimulus direction.

These tasks differ in their stimulus onset and offset time. In the Go task, the stimulus appears before the fixation cue goes off. In the RT Go task, the fixation input remains active throughout the trial, and the network is expected to respond immediately upon stimulus onset. In the Dly Go task, the stimulus is presented briefly and followed by a delay period until the fixation cue goes off. 

\subsection{Anti Task Family}

The Anti family consists of the \textbf{Anti}, \textbf{RT Anti}, and \textbf{Dly Anti} tasks. These tasks are structurally identical to their Go-family counterparts, except that the required response is the direction opposite to the stimulus (i.e. the weaker direction in left and right). 

\subsection{Decision Making (DM) Task Family}

The DM family consists of five perceptual decision-making tasks: \textbf{DM~1}, \textbf{DM~2}, \textbf{Context-dependent DM~1 (Ctx DM~1)}, \textbf{Context-dependent DM~2 (Ctx DM~2)}, and \textbf{Multisensory DM (MultSen DM)}.

In each DM task, two stimuli are presented simultaneously and remain visible until the end of the trial. The direction of stimulus~1 is sampled uniformly from \([0^\circ, 360^\circ)\), while stimulus~2 is drawn such that its direction differs from stimulus~1 by an angle uniformly sampled between \(90^\circ\) and \(270^\circ\). The stimuli have different strengths.

In DM~1 and DM~2, both stimuli appear exclusively in modality~1 or modality~2, respectively, and the correct response corresponds to the direction of the stimulus with greater strength. In the context-dependent tasks, each stimulus appears in both modalities. In Ctx DM~1, information from modality~2 must be ignored, whereas in Ctx DM~2, information from modality~1 must be ignored. In the Multisensory DM task, the correct response is determined by the stimulus with the stronger combined strength across modalities~1 and~2.

\subsection{Delayed Decision Making (Dly DM) Task Family}

The Dly DM family includes \textbf{Dly DM~1}, \textbf{Dly DM~2}, \textbf{Ctx Dly DM~1}, and \textbf{Ctx Dly DM~2}. These tasks parallel the corresponding DM tasks but introduce temporal delay between the two stimuli.

In these tasks, both stimuli are presented briefly and separated by a delay period, followed by an additional short delay after the offset of the second stimulus. The Dly DM tasks are inspired by classical parametric working-memory experiments and require the network to retain stimulus information across delays to make a correct comparison.

\subsection{Matching Task Family}

The Matching family consists of \textbf{Delayed Match-to-Sample (DMS)}, \textbf{Delayed Non-Match-to-Sample (DNMS)}, \textbf{Delayed Match-to-Category (DMC)}, and \textbf{Delayed Non-Match-to-Category (DNMC)} tasks.

In these tasks, two stimuli are presented sequentially and separated by a delay period. Each stimulus may appear in either modality. The required response depends on whether the two stimuli are considered a match. In DMS and DNMS, stimuli are matched if they share the same direction, regardless of modality. In DMC and DNMC, stimuli are matched if their directions fall within the same category, with categories defined as \([0^\circ, 180^\circ)\) and \([180^\circ, 360^\circ)\).

In DMS and DMC tasks, the network should respond in the direction of the second stimulus if the stimuli match and maintain fixation otherwise. In DNMS and DNMC tasks, the network should respond only when the stimuli do not match and maintain fixation for matched trials.

During training, half of the trials are matching and half are non-matching. In DMS and DNMS, the first stimulus is drawn randomly; the second stimulus either matches the first or differs by an angle uniformly sampled between \(10^\circ\) and \(350^\circ\). In DMC and DNMC, both stimuli are drawn independently from a uniform directional distribution.

%% file: section/appendix/multilr.tex
\begin{figure}{}
    \centering
\includegraphics[width=0.6\linewidth]{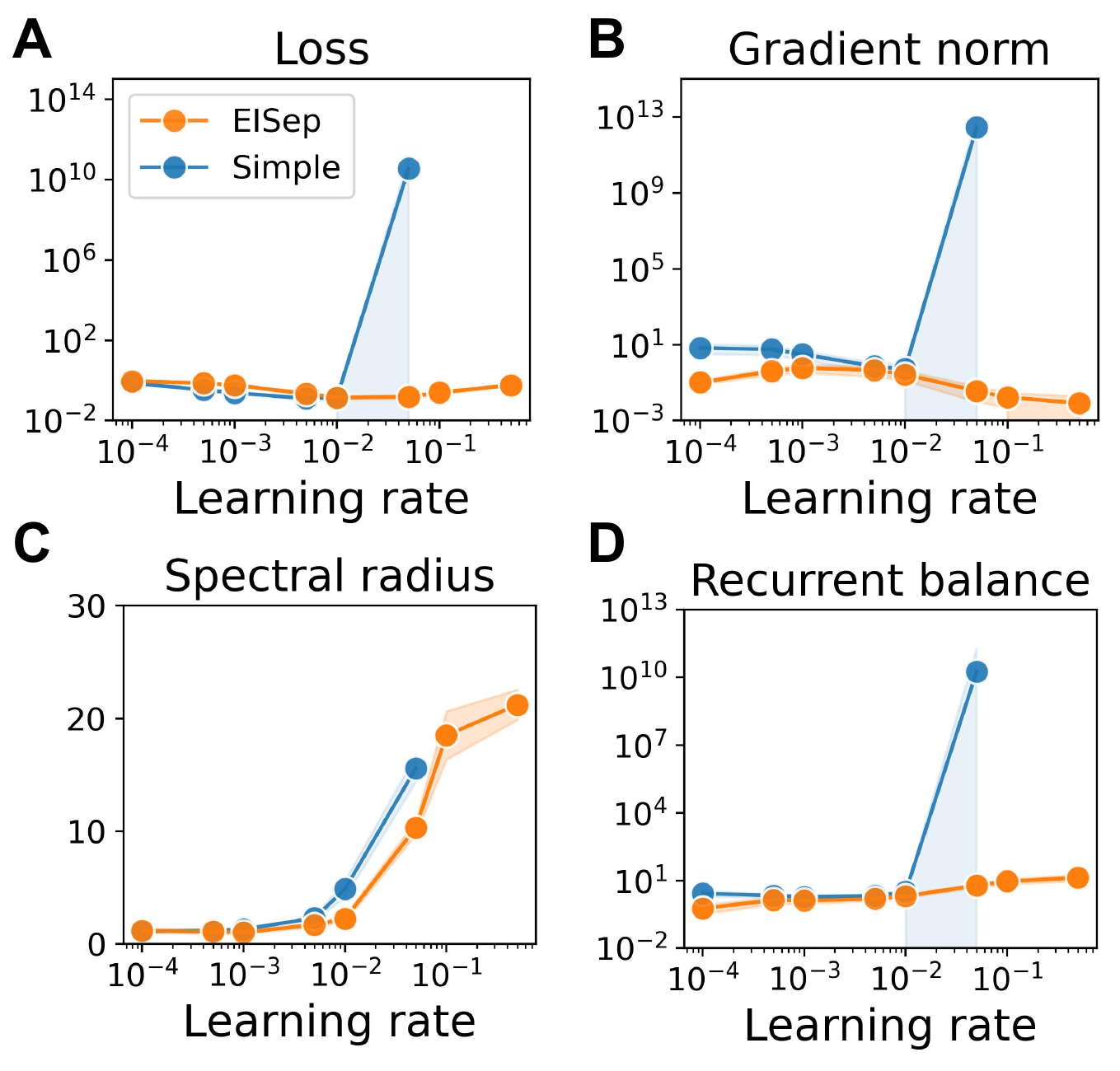}
  \caption{Training metrics under different learning rates in the cognitive task suite, including (\textbf{A.}) loss, (\textbf{B.}) gradient norm, (\textbf{C.}) spectral radius, and (\textbf{D.}) recurrent balance. }
    \vspace{-0.1in}
  \label{fig:appendix_multilr_eisep}
\end{figure}

Fig.~\ref{fig:cog_performance_stability}C shows that EISep RNNs maintain superior performance at higher learning rates and exhibit enhanced robustness during optimization. We hypothesize that this advantage arises from the intrinsic stabilized training dynamics induced by E-I separation. Here, we provide additional evidence to support this interpretation. Across a wide range of learning rates, EISep consistently exhibits lower training loss (Fig.~\ref{fig:appendix_multilr_eisep}A), smaller gradient norms (Fig.~\ref{fig:appendix_multilr_eisep}B), and a more tightly bounded spectral radius of the recurrent weight matrix (Fig.~\ref{fig:appendix_multilr_eisep}C). Notably, these stability indicators co-occur with a tighter balance between excitatory and inhibitory recurrent activity (Fig.~\ref{fig:appendix_multilr_eisep}D). The recurrent balance is defined as the averaged input of each neuron, calculated as $r_b=\text{mean}(Wh)$.

\begin{figure}{}
    \centering
\includegraphics[width=0.7\linewidth]{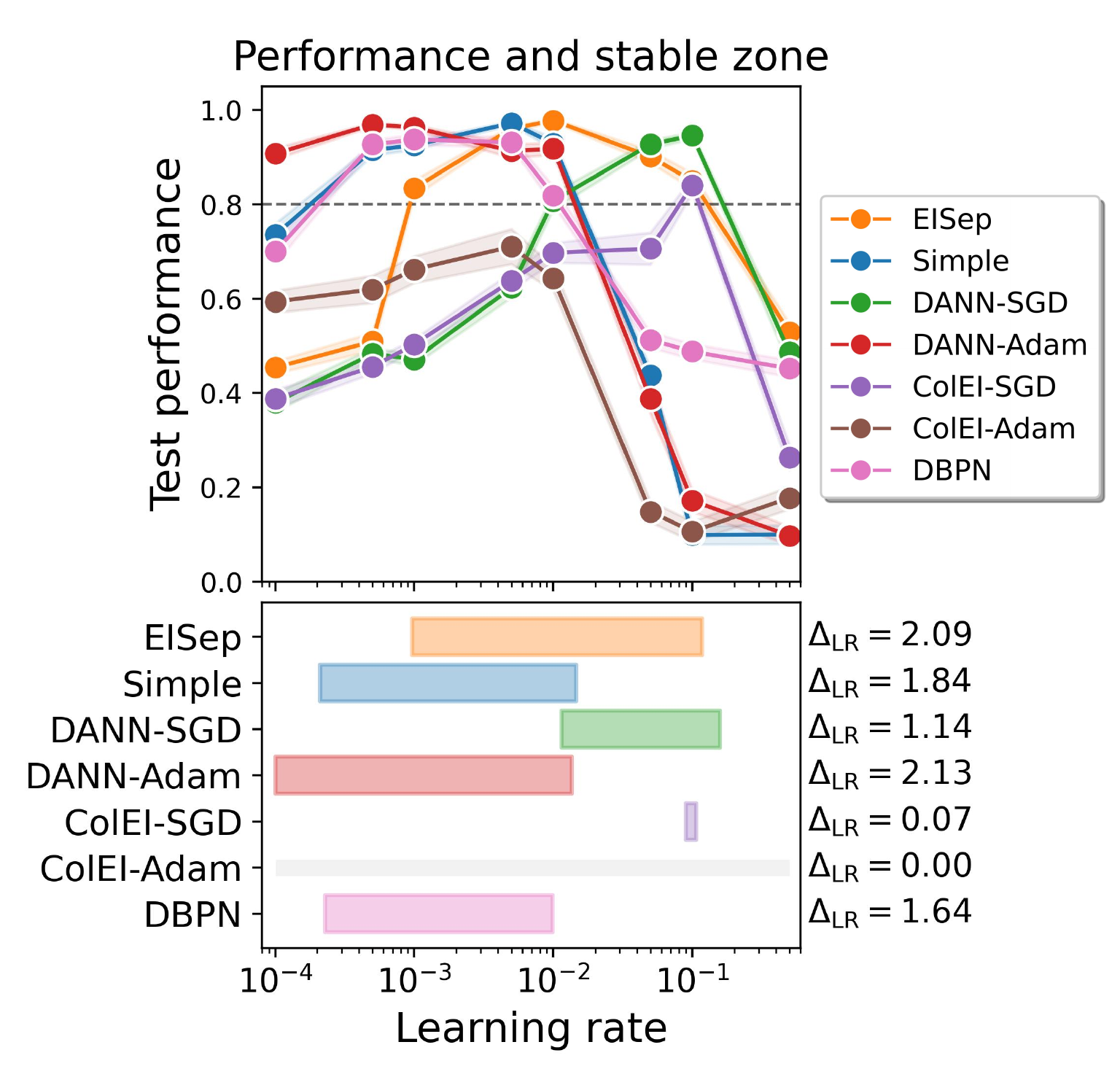}
  \caption{Performance and stable zones for different baselines vs. learning rates in the cognitive task suite.}
  \label{fig:appendix_multilr_baselines}
\end{figure}

We further evaluate all baseline models across the same range of learning rates (Fig.~\ref{fig:appendix_multilr_baselines}). To summarize training robustness, we define a \emph{stable zone} for each model as the range of log learning rates over which the lower confidence bound of performance (LCB, mean minus standard error) exceeds 0.8. To estimate its boundaries, we identify adjacent learning-rate values whose performance lies immediately above and below the threshold, and linearly interpolate between them to determine the intersection points with the line performance $=0.8$. We quantify the width of the stable zone as $\Delta_{\mathrm{LR}}$, defined as the difference between its right and left boundaries; if a model fails to exceed the threshold at any learning rate, $\Delta_{\mathrm{LR}}=0$. 

Under this criterion, EISep achieves a wider stable zone than most other baselines, with the exception of DANN-Adam. However, DANN-Adam's stable zone is concentrated at lower learning rates, indicating that it is better at escaping from the plateaus rather than stabilizing gradients. In contrast, EISep's stable zone resides at higher learning-rate area, consistent with the stabilization effects induced by balanced recurrent dynamics. Together, these results indicate that EISep achieves improved optimization stability through a biologically grounded mechanism (E-I balance).

%% file: section/appendix/modularity_q.tex
\subsection{Case study}\label{sec:appendix_case_modularity}

In the 2AFC task, the network chooses the stronger signal out of two noisy inputs (Sec.~\ref{sec:result_modularity}). To identify functional structure, we quantify neuron’s selectivity using signal-detection index $d'$. For each unit $i$, we compute 
\vspace{-0.5em}
\begin{align}
    d'_i = \frac{\mu_i^{(1)} - \mu_i^{(2)}}{\sqrt{((\sigma_i^{(1)})^2 + (\sigma_i^{(2)})^2)/2}}~,
\end{align}
where $\mu_i^{(k)}$ and $\sigma_i^{(k)}\quad (k=1,2)$ are the mean and standard deviation of its activity on those trials when signal $k$ is stronger. We label units with $d' > 0.1$ as selective for Signal 1 (Group 1, G1), units with $d' < -0.1$ as selective for Signal 2 (G2), and units with $|d'| \le 0.1$ as non-selective (NS).

\subsection{Population analysis}\label{sec:appendix_population_modularity}
Since the recurrent weight matrix is directed and signed, it does not satisfy the symmetric nonnegative assumptions of standard spectral clustering and graph-cut methods. We therefore use two adapted approaches for clustering directed signed graphs: \textit{Directed spectral clustering} and \textit{Lifted spectral clustering}. We then evaluate the resulting partitions using \textit{signed modularity}.

\textbf{Setup.}
Given a directed signed recurrent weight matrix $W\in\mathbb{R}^{n\times n}$, we first remove self-connections by setting the diagonal of $W$ to zero. We then split the matrix into nonnegative positive and negative parts,
\begin{align}
    A^+_{ij}=\max(W_{ij},0),\qquad
    A^-_{ij}=\max(-W_{ij},0).\nonumber
\end{align}

Rows correspond to outgoing connections. In practice, we transpose the recurrent weight matrix before graph construction so that graph rows align with outgoing synaptic projections.

Define the directed strengths and total masses
\begin{align}
    &k^{\pm,\mathrm{out}}_i=\sum_j A^\pm_{ij},\qquad
    k^{\pm,\mathrm{in}}_j=\sum_i A^\pm_{ij},\nonumber\\
    &m^\pm=\sum_{ij}A^\pm_{ij},\nonumber\\
    &D^{\pm,\mathrm{out}}=\mathrm{diag}(k^{\pm,\mathrm{out}}),\qquad
    D^{\pm,\mathrm{in}}=\mathrm{diag}(k^{\pm,\mathrm{in}}).\nonumber
\end{align}

To avoid numerical singularities from zero-degree nodes (e.g., neurons with exclusively inhibitory or exclusively excitatory outgoing weights under Dale's principle), all inverse square-root degree normalizations are stabilized with a small constant $\varepsilon$ before inversion.

\subsection*{Directed spectral clustering}

We first perform two-sided degree normalization to preserve source--target heterogeneity:
\begin{align}
    M^+
    &=
    (D^{+,\mathrm{out}})^{-1/2}
    A^+
    (D^{+,\mathrm{in}})^{-1/2},
    \nonumber\\
    M^-
    &=
    (D^{-,\mathrm{out}})^{-1/2}
    A^-
    (D^{-,\mathrm{in}})^{-1/2}.
    \nonumber
\end{align}

Each normalized matrix is then symmetrized,
\begin{align}
    S^+
    &=
    \tfrac12(M^+ + {M^+}^\top),
    \nonumber\\
    S^-
    &=
    \tfrac12(M^- + {M^-}^\top),
    \nonumber
\end{align}
and combined into a symmetric signed similarity matrix
\begin{align}
    S = S^+ - S^-.\nonumber
\end{align}

Define the signed degree
\begin{align}
    d_i=\sum_j |S_{ij}|,
    \qquad
    D=\mathrm{diag}(d),
    \nonumber
\end{align}
and the signed normalized Laplacian-like operator
\begin{align}
    L
    =
    I-D^{-1/2}SD^{-1/2}.
    \nonumber
\end{align}

We compute the $k$ eigenvectors associated with the smallest eigenvalues of $L$, row-normalize the embeddings, and apply $k$-means clustering to obtain cluster labels $c$.

\subsection*{Lifted spectral clustering}

We construct a lifted $2n\times2n$ adjacency matrix that encodes attraction and repulsion:
\begin{align}
    \widetilde A
    =
    \begin{bmatrix}
        A^+ & A^- \\
        A^- & A^+
    \end{bmatrix}.
    \nonumber
\end{align}

To obtain a symmetric spectral objective, we symmetrize the lifted adjacency,
\begin{align}
    \widetilde A_{\mathrm{sym}}
    =
    \tfrac12(\widetilde A+\widetilde A^\top),
    \qquad
    \widetilde D
    =
    \mathrm{diag}(\widetilde A_{\mathrm{sym}}\mathbf 1),
    \nonumber
\end{align}
and define the normalized Laplacian
\begin{align}
    \widetilde L
    =
    I-
    \widetilde D^{-1/2}
    \widetilde A_{\mathrm{sym}}
    \widetilde D^{-1/2}.
    \nonumber
\end{align}

We compute the $k$ smallest-eigenvalue eigenvectors of $\widetilde L$, row-normalize the embeddings, and apply $k$-means clustering on the $2n$ lifted nodes. The resulting labels $\widetilde c$ are folded back to the original graph by assigning each original node $i$ the label of its ``plus'' copy $i^+$.

\subsection*{Signed modularity}
Given a clustering assignment $c$, we compute the directed signed modularity
\begin{align}
    Q_{\mathrm{signed}}
    &=
    \frac{1}{m^+}
    \sum_{i,j}
    \left(
        A^+_{ij}
        -
        \frac{
            k_i^{+,\mathrm{out}}
            k_j^{+,\mathrm{in}}
        }{m^+}
    \right)
    \mathbf 1[c_i=c_j]
    \nonumber\\
    &\quad
    -
    \frac{1}{m^-}
    \sum_{i,j}
    \left(
        A^-_{ij}
        -
        \frac{
            k_i^{-,\mathrm{out}}
            k_j^{-,\mathrm{in}}
        }{m^-}
    \right)
    \mathbf 1[c_i=c_j].
    \nonumber
\end{align}

This quantity measures excess positive within-module connectivity minus excess negative within-module connectivity relative to directed degree-preserving null models. Larger $Q_{\mathrm{signed}}$ indicates stronger modular organization.

\subsection*{Evaluation protocol}

For each trained network and random seed, we extract the recurrent weight matrix and apply both directed signed spectral clustering and lifted spectral clustering. For each method, we scan the number of clusters from $k=2$ to $k=10$. Each clustering is repeated with five random restarts of $k$-means using different random seeds, and the partition with the largest signed modularity is retained. Fig.~\ref{fig:modularity} reports the mean modularity across random seeds together with standard error across seeds.

%% file: section/appendix/continual_learning.tex
\subsection{Training curve under different learning rates}

\begin{figure}[t]
    \centering
\includegraphics[width=\linewidth]{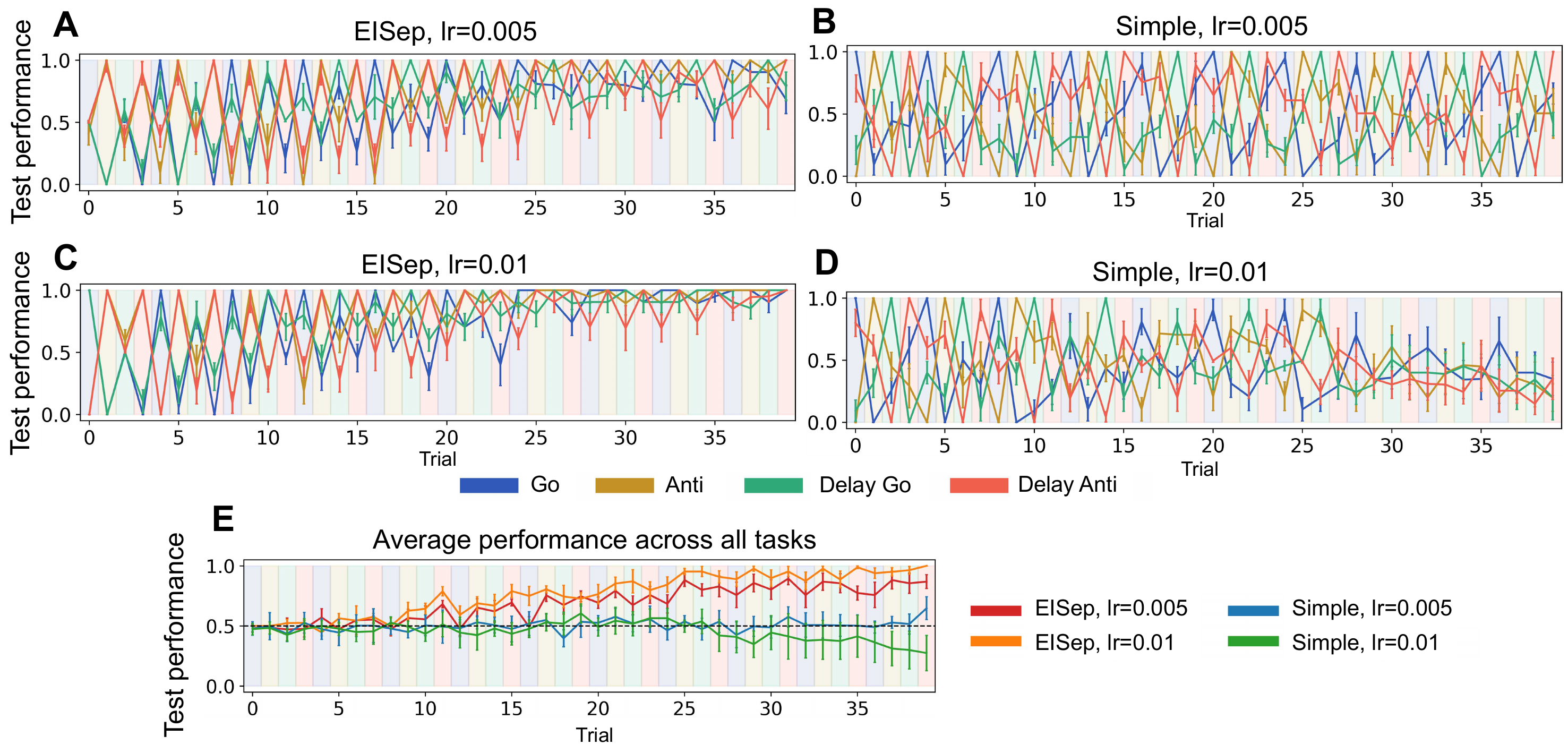}
    \vspace{-0.2in}
  \caption{In continual learning, a full task sequence is composed of four tasks, and the sequence is repeated for 10 times, resulting in 40 task trials in total. \textbf{A-D.} Training curve in continual learning for different models and learning rates. For a two-channel stimulus, the two ``go'' tasks require selecting the stronger channel, while the two ``anti'' tasks require the weaker one. The two ``delay'' variants insert a delay period between stimulus and response. \textbf{E.} Average performance across all tasks over training.}
    \vspace{-0.2in}
  \label{fig:appendix_continual_detailed}
\end{figure}

Here we show the performance of each task during the whole training process in Fig.~\ref{fig:biology_continual}A. EISep achieves higher performance with resistance to catastrophic forgetting (Fig.~\ref{fig:appendix_continual_detailed}).

\subsection{Decoding loss}\label{sec:appendix_continual_loss}

\begin{figure}{}
    \centering
\includegraphics[width=0.7\linewidth]{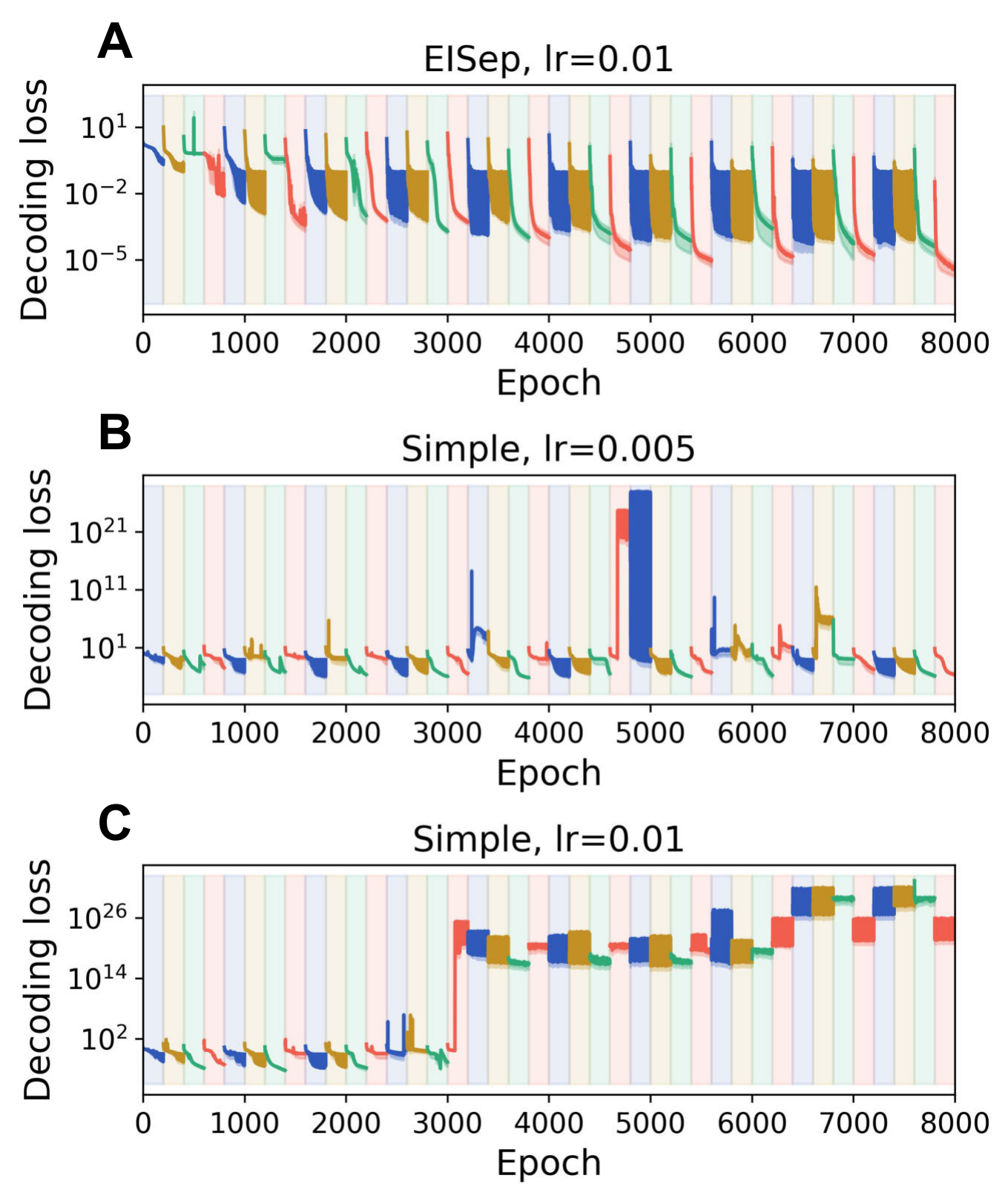}
  \caption{Loss in continual learning for tasks in Fig.~\ref{fig:biology_continual}. \textbf{A.} Loss curve of EISep with learning rate $1\times10^{-2}$. \textbf{B.} Loss curve of Simple with learning rate $5\times10^{-3}$. \textbf{C.} Loss curve of Simple with learning rate $1\times10^{-2}$.}
  \label{fig:appendix_continual_learning_loss}
\end{figure}

\begin{figure}{}
    \centering
\includegraphics[width=\linewidth]{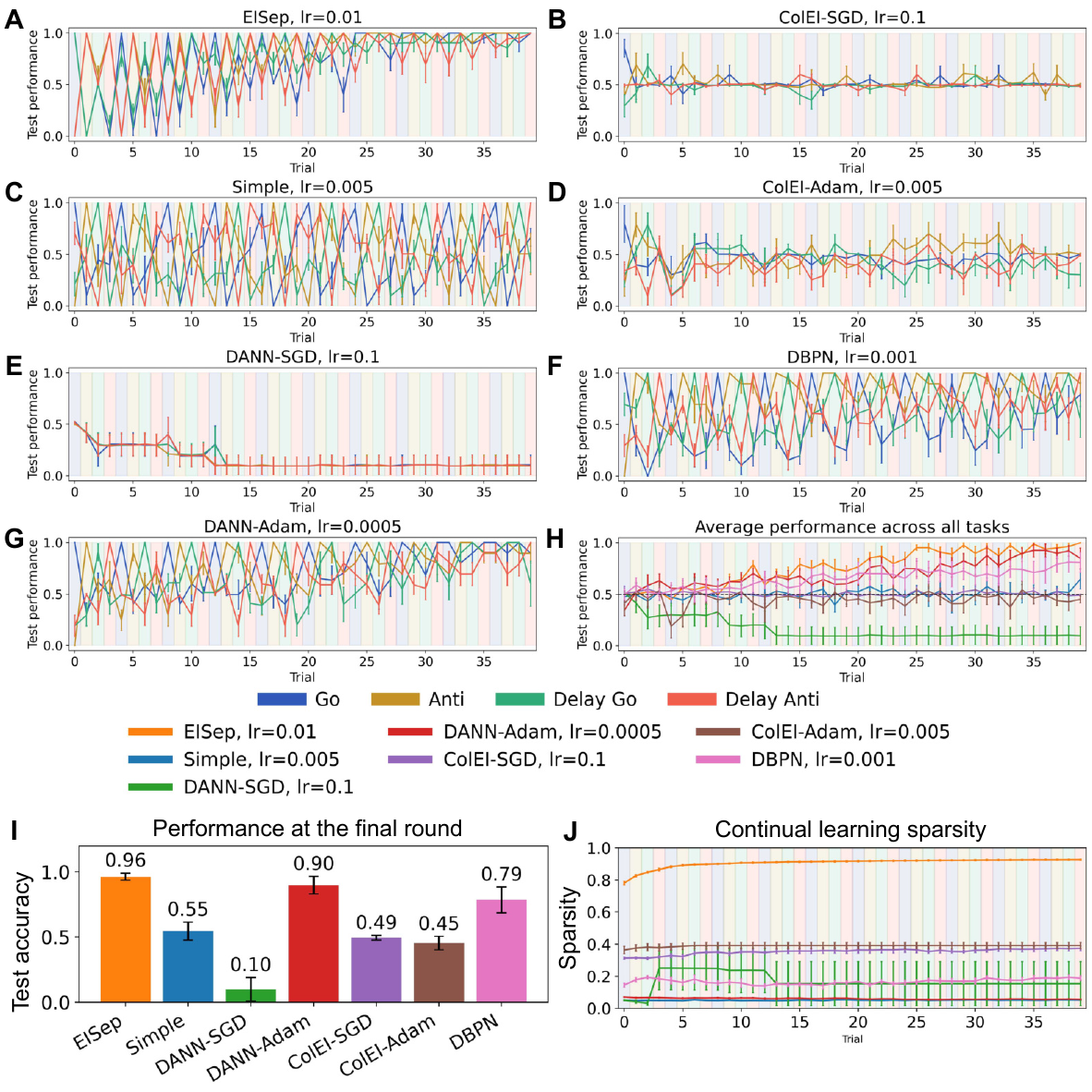}
    \vspace{-0.1in}
    \caption{
    \textbf{A--G.} Training curves of different models with their best learning rates.
    \textbf{H.} Average performance across tasks.
    \textbf{I.} Final-round performance averaged over four trials. \textbf{J.} The sparsity of all models across trials. 
    }
  \label{fig:appendix_continual_baselines}
\end{figure}

Here we report the training loss trajectories for the continual learning task described in Sec.~\ref{sec:continual_learning}. Compared to the Simple, EISep exhibits consistently lower loss and substantially fewer loss spikes throughout training (Fig.~\ref{fig:appendix_continual_learning_loss}), which translates into stronger resistance to catastrophic forgetting at convergence (Fig.~\ref{fig:biology_continual}). We further evaluate additional baseline models using their respective optimal learning rates in Fig.~\ref{fig:appendix_continual_baselines}. EISep achieves better overall performance than most other baseline models (Fig.~\ref{fig:appendix_continual_baselines}H). At the final round of the four task trials, we compute performance averaged across these four ending trials, and EISep still performs well, highlighting its superior optimization stability and robustness in continual learning settings.

%% file: section/appendix/supervised_learning.tex
To assess the generalizability of EISep RNN beyond cognitively motivated tasks, continual learning, and reinforcement learning, we further evaluate the model on a set of standard supervised benchmarks commonly used to test recurrent neural networks.

\begin{figure*}[t]
    \centering
\includegraphics[width=\textwidth]{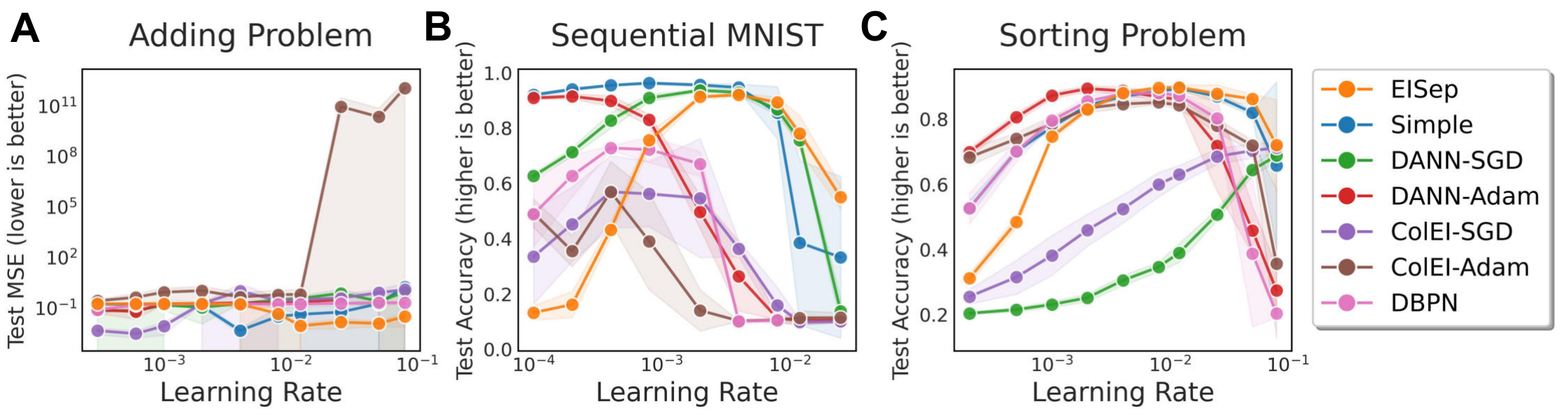}
  \caption{Performance in general tasks, including  the adding problem (\textbf{A}), sequential MNIST (\textbf{B}) and  sorting problem (\textbf{C}). }
    \vspace{-0.1in}
  \label{fig:supervised_learning}
\end{figure*}

We consider one vision task and two algorithmic benchmarks that probe complementary aspects of general-purpose sequence processing (Fig.~\ref{fig:supervised_learning}). Together, these tasks span long-term temporal integration, symbolic reasoning, and high-dimensional perceptual inference, each emphasizing distinct computational demands:

\textit{Adding problem.} The adding problem serves as a benchmark for long-term dependency learning. In this synthetic task, the network receives a length-20 sequence of two-dimensional inputs. The first dimension contains a random number in $[0,1]$ at each time step, while the second dimension is a binary indicator marking several positions in the sequence with probability $p=0.2$. The objective is to output the sum of the marked numbers at the final time step. Because all unmarked inputs act as distractors, the network must retain relevant information over long temporal gaps and perform multiple arithmetic operations. This task is widely used to evaluate an RNN’s ability to preserve information in the presence of vanishing gradients. To increase the difficulty, we change the number of marked locations from 2 (fixed) to a variable number $n_v$, sampled randomly for each sequence. 

\textit{Sorting problem.} The sorting task probes a model’s ability to perform algorithmic sequence transformations. The input consists of a fixed-length sequence $s$ of $n$ real-valued numbers ($n=5$ in our experiments). The network is required to output the rank of each element in the sequence, i.e., $\mathrm{argsort}(s)$. Because the correct output at each position depends on the global ordering of the entire sequence, successful performance requires learning relational and comparison-based representations across time.

\textit{Sequential MNIST.}  Sequential MNIST evaluates a network’s capacity to integrate high-dimensional perceptual information over extended temporal horizons. Each MNIST image is unrolled row by row and presented as a sequence of $28$ time steps, with each step corresponding to one row of the $28\times28$ image. The model must accumulate evidence across the entire sequence and produce the correct digit label only after processing the final row.

Across all settings, EISep exhibits the same performance pattern observed previously: stronger performance at higher learning rates and relatively weaker performance at lower learning rates. Nevertheless, EISep consistently outperforms most prior Dale-constrained architectures, indicating that enforcing excitatory–inhibitory separation does not substantially compromise general computational capability, contrary to concerns raised in earlier work~\citep{cornford2020learning, li2023learning}. However, the advantages here are not as apparent as in reinforcement learning, probably because in these benchmarks, networks do not suffer from gradient instability as in continual learning or RL tasks.

%% file: section/appendix/eilogit_convergence.tex
\begin{figure}{}
    \centering
\includegraphics[width=0.5\linewidth]{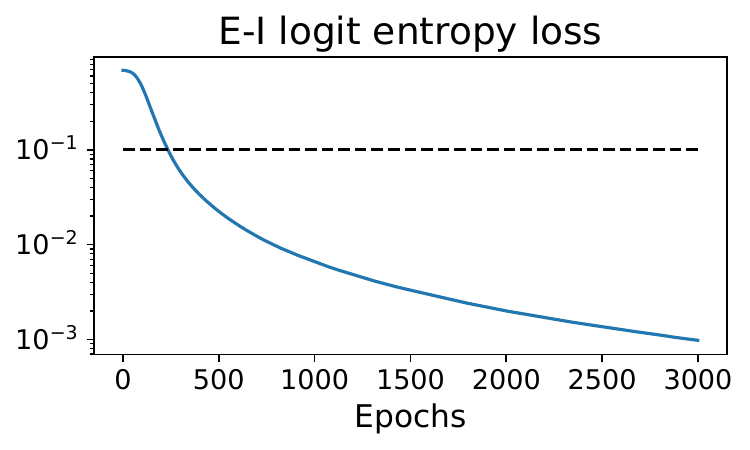}
  \caption{The EI logit entropy loss during the training stage for EISep. The fast convergence of the entropy objective indicates that the neuron signs also converge fast.}

  \label{fig:appendix_eilogit_entropy_loss}
\end{figure}

At first glance, allowing neuron signs to be learnable may appear to reduce the biological plausibility of the model. In biological neural circuits, neurons have well-established identities and do not switch between excitatory and inhibitory roles. However, the training process in general captures not only trial-by-trial synaptic plasticity but also longer-timescale developmental and even evolutionary changes~\citep{wang2021evolving}. From this perspective, transient sign changes during early learning actually correspond to neuronal differentiation, and are acceptable if this change is limited to early training. In other words, to avoid interfering with subsequent task-level learning, neuron signs should converge as training proceeds.

\begin{figure*}[t]
    \centering
\includegraphics[width=\linewidth]{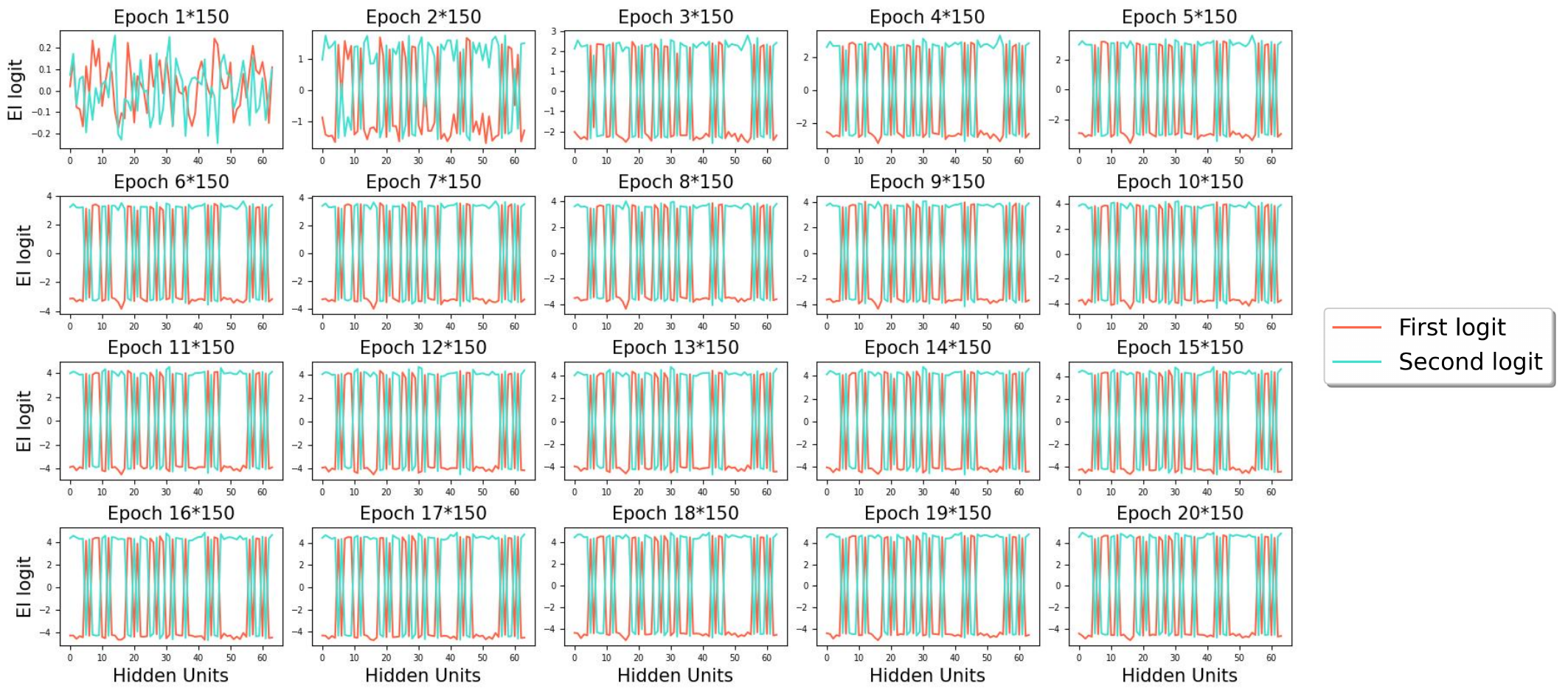}
  \caption{The EI logit matrix $M$ during the entire training stage, recorded at every 150 steps. }
  \label{fig:logit_convergence}
\end{figure*}

To see if the model has exhibited such behavior, we monitor the entropy loss defined in Sec.~\ref{sec:definition_eisep} and find that it rapidly drops below \(10^{-2}\) within the first quarter of training (Fig.~\ref{fig:appendix_eilogit_entropy_loss}), far less than the normal order of decoding loss ($10^{-1}$, Fig.~\ref{fig:cog_performance_stability}D left, and dash line in Fig.~\ref{fig:appendix_eilogit_entropy_loss}), which is supported by directly tracking the evolution of the logit matrix $M$ (Sec.~\ref{sec:definition_eisep}) throughout training (Fig.~\ref{fig:logit_convergence}). This result indicates that neuron signs converge much earlier than when the network finishes task optimization.

%% file: section/appendix/sparsity_validity.tex
\begin{figure*}[t]
    \centering
\includegraphics[width=\textwidth]{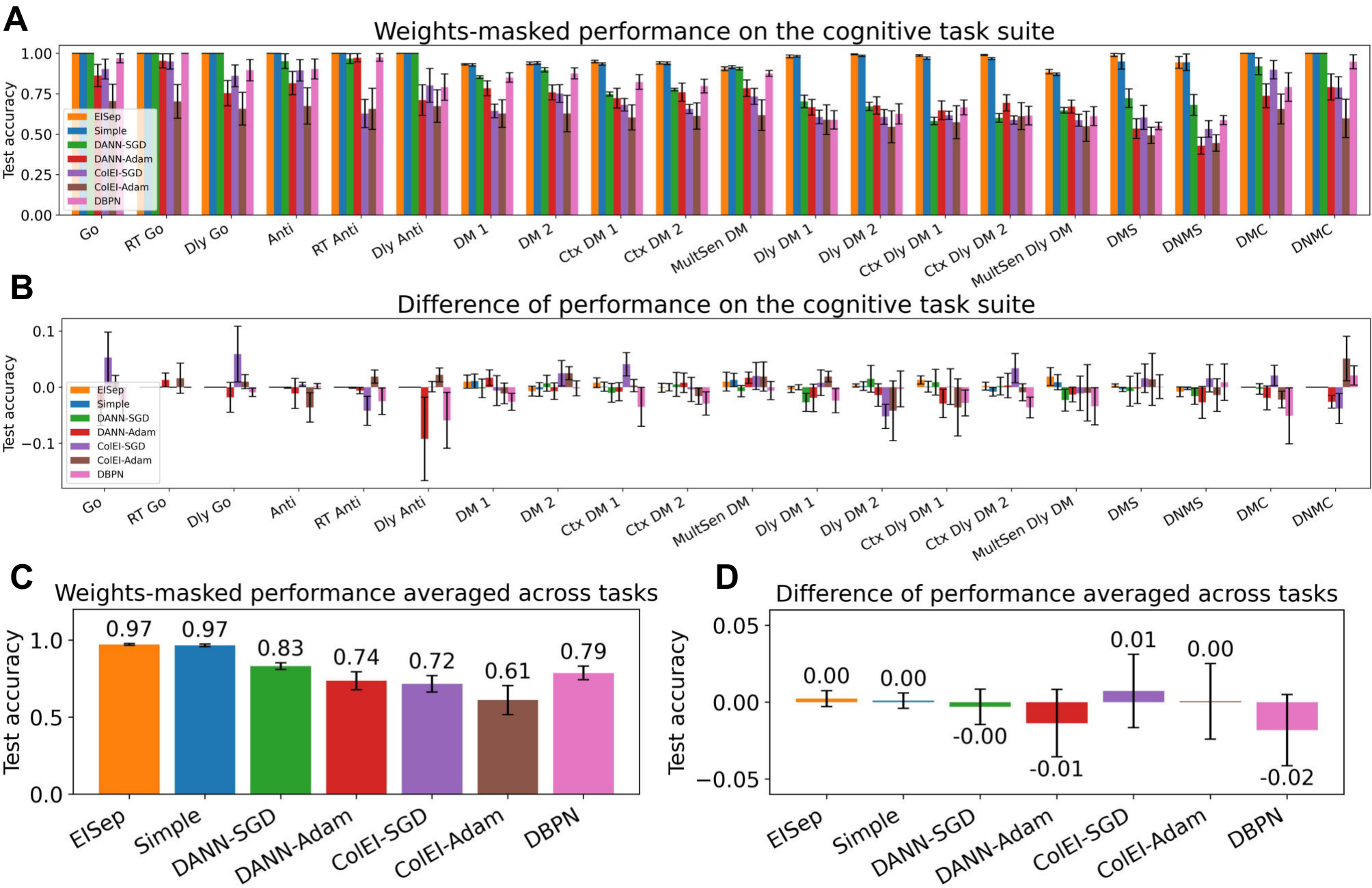}
  \caption{The performance of masked models and their relative errors with the original models. \textbf{A}. Performance of 7 masked models on 20 cognitive tasks. \textbf{B}. The performance differences of 7 masked models on 20 cognitive tasks. \textbf{C}. Performance of 7 models averaged across tasks. \textbf{D}. The performance differences of 7 models averaged across tasks.}
    \vspace{-0.2in}
  \label{fig:appendix_sparsity_validity}
\end{figure*}

In Sec.~\ref{sec:result_sparsity}, we show that the EISep RNN achieves substantially higher sparsity than Simple and other Dale-constrained networks. Formally, sparsity $r$ is defined as the proportion of small-magnitude weights,
\begin{align}
    r = \frac{1}{N^2}\sum_{i,j}\left[\mathds{1}\!\left(|W_{ij}| < 0.1\,\sigma(|W|)\right)\right],
\end{align}
where weights whose magnitudes fall below \(10\%\) of the standard deviation are considered negligible. To verify that these small weights do not play a critical role in model performance, we construct a sparsity-controlled mask that sets such weights to zero and re-evaluate the models. Performance remains comparable to the original, unmasked networks (Fig.~\ref{fig:appendix_sparsity_validity}A), with differences close to zero (Fig.~\ref{fig:appendix_sparsity_validity}B).

When averaged across tasks, all models are largely robust to the sparsity mask (Fig.~\ref{fig:appendix_sparsity_validity}C, D), proving the validity of the sparsity measure. At the level of individual tasks, several Dale-constrained networks exhibit mild performance degradation after small weights are removed, whereas the EISep and Simple consistently preserve their performance. This suggests that EISep relies on a more specific and compact set of large-magnitude connections, with effective computation primarily supported by a small subset of dominant weights.

%% file: section/appendix/rl_performance_details.tex
\subsection{Performance of other Dale's networks in the Lunar Lander task}\label{sec:appendix_lunarlander_baselines}

Here, we show the performance of other Dale-constrained baselines on the Lunar Lander task. As in EISep (Sec.~\ref{sec:reinforcement_learning}), both the actor and the critic can be implemented with Dale-constrained recurrent modules, so there are three variants for each baseline: EI(A) (only actor is Dale-constrained), EI(C) (only critic is Dale-constrained), and EI(A+C) (both actor and critic are Dale-constrained). While some baseline variants perform comparably in specific configurations (e.g., EI(A)), their best performance across variants remains substantially lower than EISep. This outcome is consistent with the result on the multi-task cognitive benchmark where they achieve lower training stability (Fig.~\ref{fig:cog_performance_stability}D).

\begin{figure*}[t]
    \centering
\includegraphics[width=\linewidth]{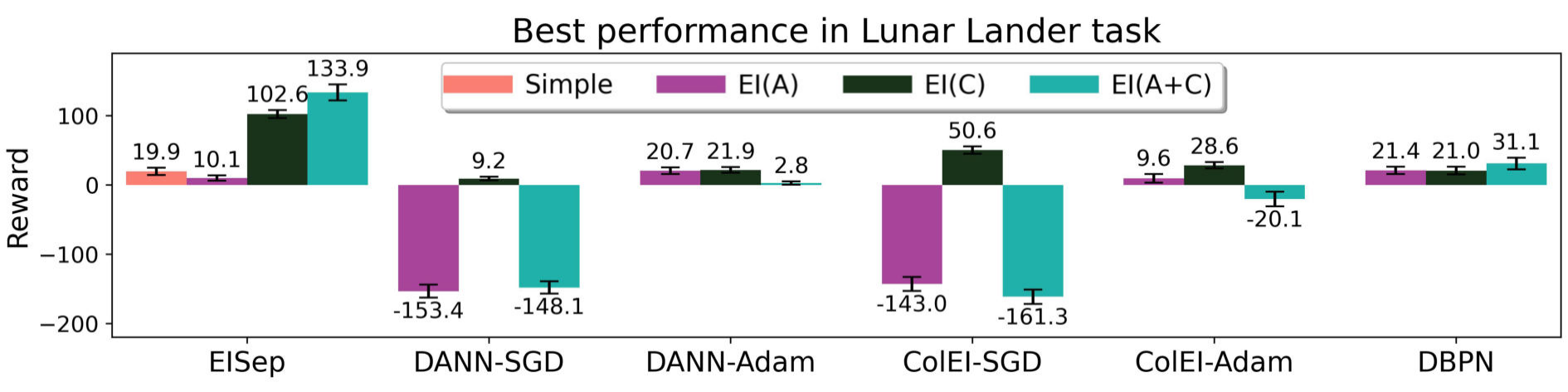}
  \caption{Performance of other Dale's networks in the Lunar Lander task. }
  \label{fig:appendix_lunarlander_baselines}
\end{figure*}

\subsection{Performance under different learning rates for EISep}\label{sec:appendix_lunarlander_eisep_lrall}

Here we show the learning-rate sweep results on Lunar Lander (Table~\ref{tab:lunarlander_lr_sweep}) and HalfCheetah (Table~\ref{tab:halfcheetah_lr_sweep}). 

Each configuration uses $n=5$ training seeds and each trained model is evaluated with $n=20$ test seeds. In the table, each cell reports test return (mean $\pm$ SE) for four model variants (top to bottom): Simple, EISep(A), EISep(C), EISep(A+C). Boldface marks the best learning-rate pair for each variant searched over the whole grid (9 pairs).

\input{figure/appendix/table_lunarlander_all_lr_eisep}
\input{figure/appendix/table_halfcheetah_all_lr_eisep}

\subsection{Performance of EISep FFN on Lunar Lander}\label{sec:appendix_eisep_ffn}
\begin{figure}[t]
    \centering
\includegraphics[width=0.5\linewidth]{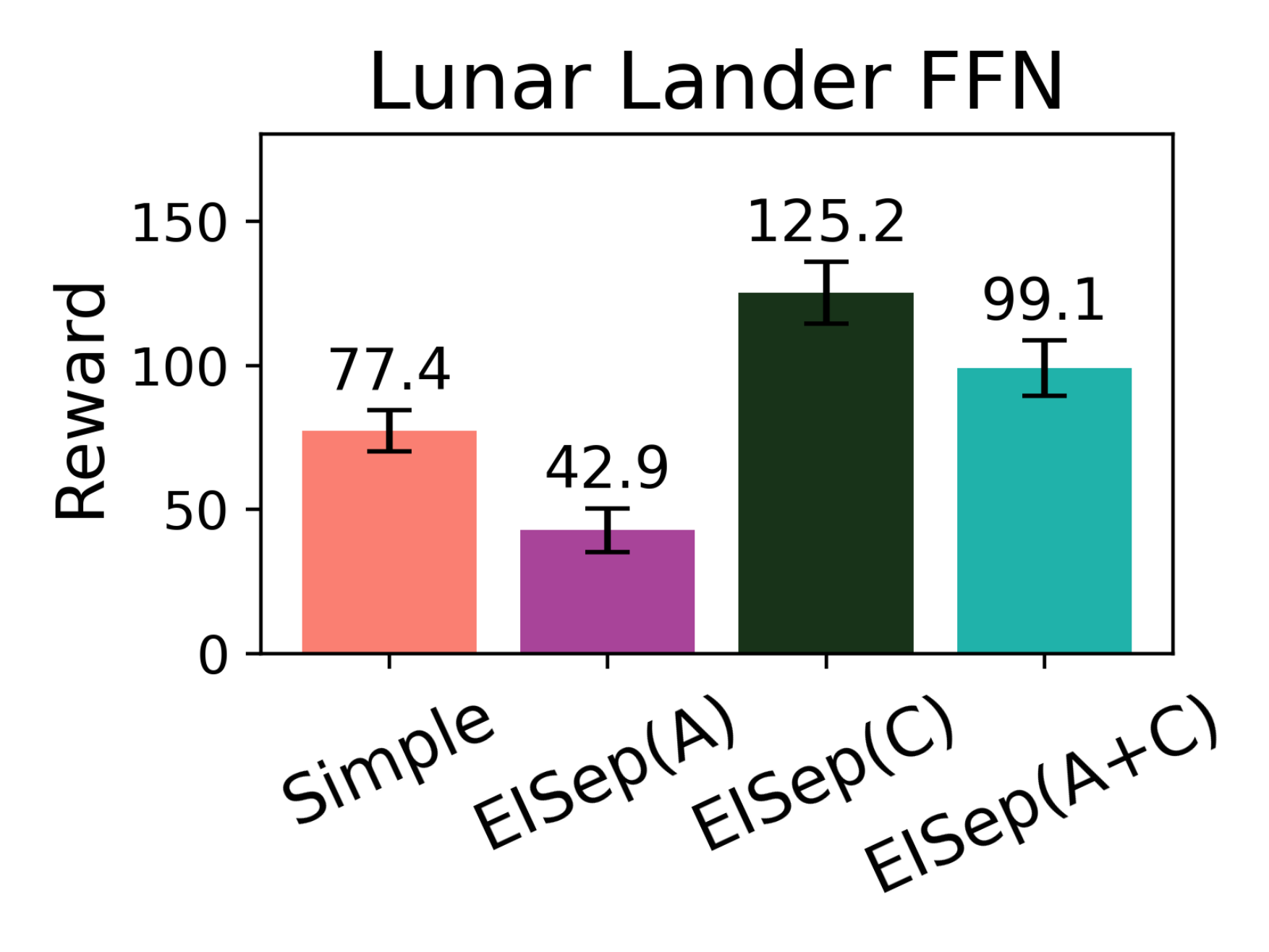}
  \caption{Performance of EISep FFN on the Lunar Lander task. }
    \vspace{-0.2in}
  \label{fig:appendix_lunarlander_ffn}
\end{figure}

Here we change the RNN module to the feedforward network (FFN) and test it on the Lunar Lander task. To control the parameter size of the network, we only replace the recurrent layer with a feedforward layer of the same hidden size, positioned between the input and output layers, and remove the hidden state entirely. Under this setting, the EISep variant continues to outperform the Simple baseline (Fig.~\ref{fig:appendix_lunarlander_ffn}), although the performance gap is smaller than in the recurrent case. This could be explained in two aspects. The higher performance of Simple FFN than Simple RNN suggests that feedforward architectures are less susceptible to training instability than RNNs, which are known to suffer from gradient explosion. The lower performance of EISep FFN than EISep RNN is consistent with the advantage of recurrent modules to utilize past experiences to make decisions. Taken together, these results indicate that EISep specifically mitigates the gradient instability inherent to recurrent architectures, thereby unlocking the representational and computational advantages of RNNs in complex control tasks.

%% file: figure/appendix/table_lunarlander_all_lr_eisep.tex
\begin{table}[t]
  \centering
  \caption{Learning-rate sweep on LunarLander-v3.}
  \label{tab:lunarlander_lr_sweep}
  \setlength{\tabcolsep}{3pt}
  \renewcommand{\arraystretch}{0.9}
    \footnotesize
  \begin{tabular}{lccc}
    \toprule
    \textbf{lr(C) $\backslash$ lr(A)} & $\mathbf{3\times10^{-4}}$ & $\mathbf{1\times10^{-3}}$ & $\mathbf{3\times10^{-3}}$ \\
    \midrule
    $\mathbf{1.5\times10^{-3}}$ & \makecell[tc]{$\mathbf{19.90 \pm 5.21}$\\[1pt]$-54.30 \pm 7.56$\\[1pt]$69.99 \pm 7.51$\\[1pt]$-75.01 \pm 11.42$} & \makecell[tc]{$14.60 \pm 2.97$\\[1pt]$7.04 \pm 3.08$\\[1pt]$66.74 \pm 7.52$\\[1pt]$11.57 \pm 2.24$} & \makecell[tc]{$1.19 \pm 5.62$\\[1pt]$-5.25 \pm 2.53$\\[1pt]$5.57 \pm 11.14$\\[1pt]$31.98 \pm 3.61$} \\
    \midrule
    $\mathbf{5\times10^{-3}}$ & \makecell[tc]{$10.63 \pm 3.28$\\[1pt]$-54.26 \pm 10.73$\\[1pt]$\mathbf{102.57 \pm 5.79}$\\[1pt]$-58.65 \pm 8.92$} & \makecell[tc]{$-6.09 \pm 2.68$\\[1pt]$\mathbf{10.13 \pm 3.94}$\\[1pt]$101.52 \pm 10.22$\\[1pt]$45.41 \pm 7.85$} & \makecell[tc]{$-17.44 \pm 6.36$\\[1pt]$-0.69 \pm 2.07$\\[1pt]$61.33 \pm 7.55$\\[1pt]$\mathbf{133.91 \pm 11.46}$} \\
    \midrule
    $\mathbf{1.5\times10^{-2}}$ & \makecell[tc]{$2.98 \pm 2.75$\\[1pt]$-77.64 \pm 12.96$\\[1pt]$70.89 \pm 7.52$\\[1pt]$-25.56 \pm 6.63$} & \makecell[tc]{$0.04 \pm 2.07$\\[1pt]$9.75 \pm 3.19$\\[1pt]$93.81 \pm 7.89$\\[1pt]$37.88 \pm 7.43$} & \makecell[tc]{$14.09 \pm 5.11$\\[1pt]$-29.33 \pm 3.74$\\[1pt]$14.11 \pm 9.87$\\[1pt]$52.23 \pm 9.68$} \\
    \bottomrule
  \end{tabular}

  \vspace{3pt}
    Chance level: $-200.36\pm4.16$.
    
  \textit{Column:} actor learning rates. \textit{Row:} critic learning rates.
  
  \textit{Order in each cell (top to bottom):} Simple, EISep(A), EISep(C), EISep(A+C).

\end{table}

%% file: figure/appendix/table_halfcheetah_all_lr_eisep.tex
\begin{table}[t]
  \centering
  \caption{Learning-rate sweep on HalfCheetah-v5.}
  \label{tab:halfcheetah_lr_sweep}
  \setlength{\tabcolsep}{3pt}
  \renewcommand{\arraystretch}{0.9}
    \footnotesize
  \begin{tabular}{lccc}
    \toprule
    \textbf{lr(C) $\backslash$ lr(A)} & $\mathbf{3\times10^{-4}}$ & $\mathbf{1\times10^{-3}}$ & $\mathbf{3\times10^{-3}}$ \\
    \midrule
    $\mathbf{1.5\times10^{-3}}$ & \makecell[tc]{$1387.94 \pm 7.32$\\[1pt]$1379.76 \pm 23.80$\\[1pt]$1510.74 \pm 17.22$\\[1pt]$1359.10 \pm 14.13$} & \makecell[tc]{$1503.11 \pm 7.33$\\[1pt]$1570.41 \pm 22.11$\\[1pt]$1550.80 \pm 25.58$\\[1pt]$1623.82 \pm 26.35$} & \makecell[tc]{$1581.90 \pm 8.66$\\[1pt]$1659.70 \pm 37.18$\\[1pt]$1781.78 \pm 46.34$\\[1pt]$1585.68 \pm 8.68$} \\
    \midrule
    $\mathbf{5\times10^{-3}}$ & \makecell[tc]{$1530.66 \pm 5.55$\\[1pt]$1426.88 \pm 24.39$\\[1pt]$3089.99 \pm 56.89$\\[1pt]$2865.65 \pm 92.40$} & \makecell[tc]{$1646.83 \pm 18.94$\\[1pt]$1676.68 \pm 36.92$\\[1pt]$2297.24 \pm 89.07$\\[1pt]$2439.39 \pm 102.15$} & \makecell[tc]{$1644.64 \pm 13.33$\\[1pt]$1552.41 \pm 20.62$\\[1pt]$2266.02 \pm 68.28$\\[1pt]$1971.96 \pm 73.05$} \\
    \midrule
    $\mathbf{1.5\times10^{-2}}$ & \makecell[tc]{$1592.10 \pm 6.12$\\[1pt]$1384.61 \pm 23.61$\\[1pt]$2870.01 \pm 128.35$\\[1pt]$2542.09 \pm 138.04$} & \makecell[tc]{$1670.19 \pm 5.71$\\[1pt]$\mathbf{2019.97 \pm 60.97}$\\[1pt]$3591.95 \pm 116.64$\\[1pt]$3258.02 \pm 133.81$} & \makecell[tc]{$\mathbf{1687.00 \pm 20.22}$\\[1pt]$1693.77 \pm 25.94$\\[1pt]$\mathbf{4332.40 \pm 64.58}$\\[1pt]$\mathbf{3870.66 \pm 132.16}$} \\
    \bottomrule
  \end{tabular}

  \vspace{3pt}  
  \textit{Column:} actor learning rates. \textit{Row:} critic learning rates.
  
  \textit{Order in each cell (top to bottom):} Simple, EISep(A), EISep(C), EISep(A+C).
\end{table}

%% file: section/appendix/phase_detection.tex
\subsection{Environment details and phase annotation for the LunarLander task}

In the Lunar Lander environment, the agent controls a rocket descending toward the lunar surface using left, right, and main engines. Each action fires one of the engines, consumes fuel and affects the lander's pose. The final stable landing requires the agent to balance energy efficiency and stability. In the following part, we introduce the definitions of state and action variables, before talking about how to manually annotate the agent based on these states and actions.

\paragraph{State variables.}
At each time step, the environment state is represented by an 8-dimensional vector
\[
(x, y, v_x, v_y, \theta, \dot{\theta}, \text{leg}_L, \text{leg}_R),
\]
where $(x,y)$ denote position, $(v_x,v_y)$ linear velocities, $\theta$ the lander’s angle, $\dot{\theta}$ its angular velocity, and $\text{leg}_L,\text{leg}_R \in \{0,1\}$ indicate ground contact of the left and right landing legs. The \emph{contact} is defined as whether either leg is in contact with the ground. 

The agent selects a discrete action $a \in \{0,1,2,3\}$ at each time step, corresponding to:
\begin{itemize}
    \item $0$: do nothing,
    \item $1$: fire the left orientation engine,
    \item $2$: fire the main engine,
    \item $3$: fire the right orientation engine.
\end{itemize}

Rewards increase when the lander achieves a smooth and upright touchdown with minimal oscillation, and decrease when tilting, excessive acceleration, or rolling happens.

To analyze the internal dynamics of recurrent agents and relate them to interpretable behavioral stages, we manually annotate each environment state in the LunarLander-v3 task with a discrete \emph{phase} label. These phases are defined using fixed, environment-based thresholds on observable state variables and actions, without reference to the agent’s internal representations. This procedure provides a deterministic and reproducible mapping from low-level states to high-level behavioral regimes, as well as an important reference to the internal state analyses (Sec.\ref{sec:rl_dynamics}).

\paragraph{Phase definitions.}
Each time step is assigned one of six mutually exclusive phases based on fixed thresholds introduced below.

\begin{itemize}
    \item \textbf{Phase 0: Free fall / early descent (Default).}
    The lander is at high altitude, descending rapidly, and the main engine is not activated, corresponding to an early, weakly controlled descent:
    \[
    y > 1.0\;\land\; v_y < -2.0\;\land\; a \neq 2.
    \]

    \item \textbf{Phase 1: Braking / vertical control.}  
    The main engine is actively fired while the lander remains low-speed in the air, indicating deliberate regulation of vertical velocity:
    \[
    y > 0 \;\land\; a = 2 \;\land\; v_y > -3.0 \;\land\; \text{no leg contact}.
    \]

    \item \textbf{Phase 2: Horizontal alignment / attitude control.}  
    The lander performs corrective maneuvers to regulate lateral motion or orientation in the air:
    \[
    \text{no leg contact} \;\land\;
    \bigl(|v_x| \ge 0.5 \;\lor\; |\theta| \ge 0.3 \;\lor\; |\dot{\theta}| \ge 0.15\bigr).
    \]

    \item \textbf{Phase 3: Touchdown preparation.}  
    The lander approaches the ground with reduced velocity and near-upright orientation, preparing for landing:
    \[
    y \le 0.2 \;\land\; |v_y| < 1.0 \;\land\; |\theta| < 0.3 \;\land\; \text{no leg contact}.
    \]

    \item \textbf{Phase 4: Landed / stable contact.}  
    One or both legs are in contact with the ground and the lander remains motionless, or with minimal motion to adjust landing positions:
    \[
    \begin{aligned}
        &\text{leg contact} \;\land\; y \le 0.3 \;\land\; |v_y| < 0.5\\
        &\;\land\; |v_x| < 0.5 \;\land\; |\theta| \le 0.7.
    \end{aligned}
    \]

    \item \textbf{Phase 5: Crash.}  
    A terminal episode ends without stable ground contact and exhibits at least one hazardous condition:
    \[
    \begin{aligned}
    &\text{done} \;\land\; \neg\text{leg contact} \;\land\;
    \bigl(
        y \le 0.1 \;\lor\; |\theta| > 0.7\\
        &\lor\; |v_y| > 3.0 \;\lor\; |v_x| > 3.0 \;\lor\; |x| > 1.0
    \bigr).
    \end{aligned}
    \]
\end{itemize}

\paragraph{Assignment logic.}
Phase labels are sequentially assigned using fixed threshold rules on altitude, velocities, orientation, angular velocity, leg contact, action choices, and episode termination flags. When multiple phase conditions are satisfied, later-defined phases overwrite earlier ones, ensuring that terminal events such as landing or crash take precedence. If no specific condition is met, the state defaults to Phase~0 (free fall). In practice, the proposed criteria cover the majority of visited states, with a non-default coverage rate of $60.4\%\pm 0.4\%$ for the Simple agent and $71.5\% \pm 3.2\%$ for EISep(A+C). This shows that it is acceptable to use Phase~0 as a default label for unassigned residual states. Notably, the Simple agent tends to hover near the ground and rarely enters well-defined touchdown preparation or landing phases, resulting in a lower non-default coverage compared to EISep.

These manually annotated phases serve as an external, interpretable reference for evaluating learned representations and temporal segmentation in recurrent networks. Importantly, they are not used during training and rely solely on environment-level information, ensuring that all analyses remain strictly post hoc.

\subsection{Environment introduction for the HalfCheetah task}

In the HalfCheetah environment, a robot agent with multiple body parts connected by joints is required to run forward as efficiently and stably as possible. The agent applies continuous torques to the joints to maximize forward velocity while minimizing energy consumption. This task tests a model’s ability to perform coordinated continuous control over long time horizons and presents a substantially more challenging optimization problem than Lunar Lander. See the environment documents~\citep{towers2024gymnasium} for more details.

%% file: section/appendix/eisep_lstm.tex
In Sec.~\ref{sec:reinforcement_learning}, EISep LSTM and Simple LSTM (vanilla LSTM) are used to deal with the complex HalfCheetah environment. Here is a brief introduction of these two models.

\textbf{Simple LSTM.} 
A Long Short-Term Memory (LSTM) network augments a standard recurrent network with gating mechanisms that regulate information flow. The four gates are four recurrent transformations, corresponding to the input, forget, candidate, and output gates. 
At each time step $t$, the LSTM maintains a hidden state $h_t$ and a cell state $c_t$, updated as:
\begin{align}
i_t &= \sigma(W_{ii}x_t + W_{hi}h_{t-1} + b_i), \\
f_t &= \sigma(W_{if}x_t + W_{hf}h_{t-1} + b_f), \\
g_t &= \tanh(W_{ig}x_t + W_{hg}h_{t-1} + b_g), \\
o_t &= \sigma(W_{io}x_t + W_{ho}h_{t-1} + b_o), \\
c_t &= f_t \odot c_{t-1} + i_t \odot g_t, \\
h_t &= o_t \odot \tanh(c_t).
\end{align}
These gating mechanisms allow the network to selectively store, forget, and expose information across time, mitigating vanishing gradients and enabling long-term dependency modeling.

\vspace{0.5em}
\textbf{EISep adaptation.}
We modify the recurrent weight matrices 
$W_{hi}, W_{hf}, W_{hg}, W_{ho}$ 
to satisfy Dale’s principle, enforcing excitatory and inhibitory separation. Similar to EISep RNN (Sec.~\ref{sec:definition_eisep}), each recurrent matrix is reparameterized as:
\begin{equation}
W = \mathrm{ReLU}(W_0) D,
\end{equation}
where $W_0$ is a free learnable matrix and $D$ is a soft diagonal sign matrix with diagonal entries computed using the Gumbel-Softmax sampling.

\begin{figure}[t]
    \centering
\includegraphics[width=0.7\linewidth]{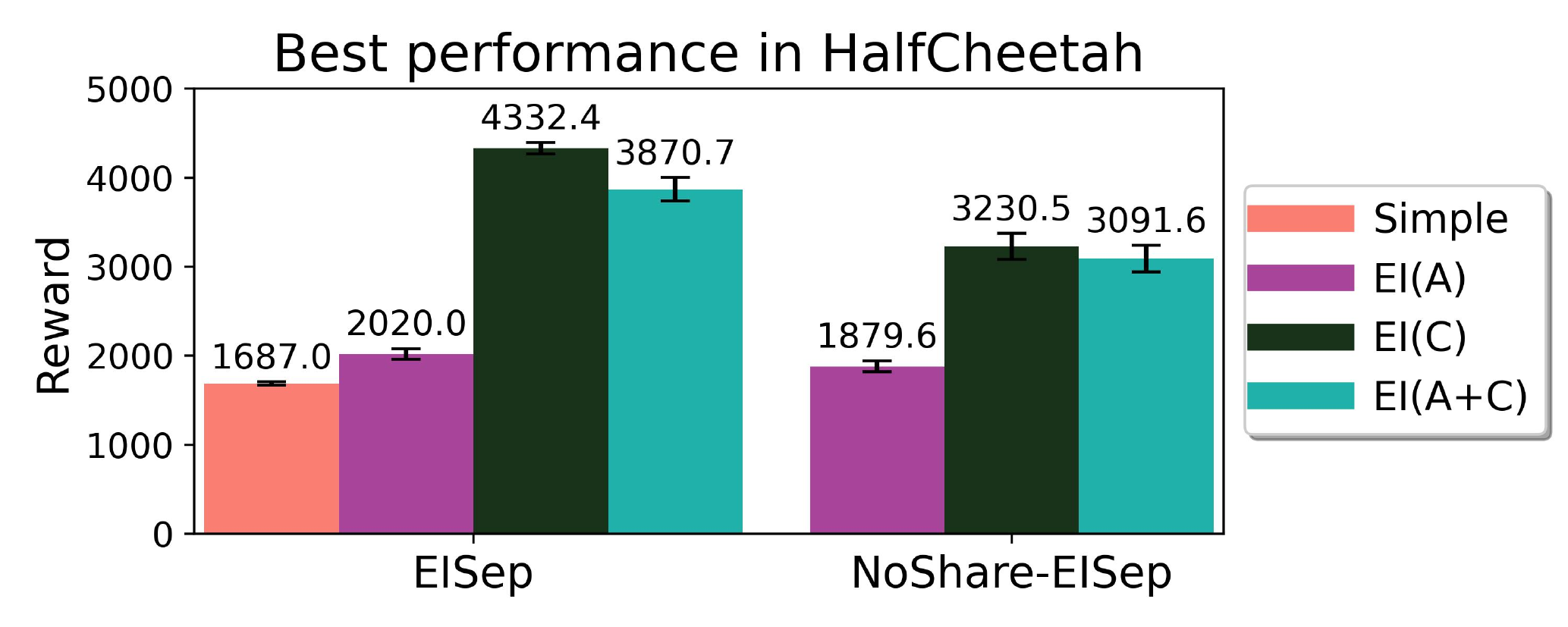}
  \caption{Performance for EISep LSTM with shared E-I identities across gates (left) or with independent identities across gates (right).}
  \label{fig:eisep_lstm_noshare}
\end{figure}

Because an LSTM updates its hidden state through four gates, a natural design question is whether these gate weight matrices should share a common sign matrix $D$, or use independent matrices $D_1, D_2, ...$. From the perspective of biological realism, if we see the hidden state as neuronal firing rates, we would expect that each neuron maintains the same excitatory or inhibitory identity across all transformations. Therefore, we enforce a shared sign mask $D$ across all recurrent matrices in the LSTM. This design is also proved better in practice (Fig.~\ref{fig:eisep_lstm_noshare}). 

Consequently, \textbf{EISep LSTM} preserves the advantages of LSTM—robust memory gating and long-term credit assignment—while inheriting the stability advantages of EISep. 

\begin{figure}[t]
    \centering
\includegraphics[width=0.7\linewidth]{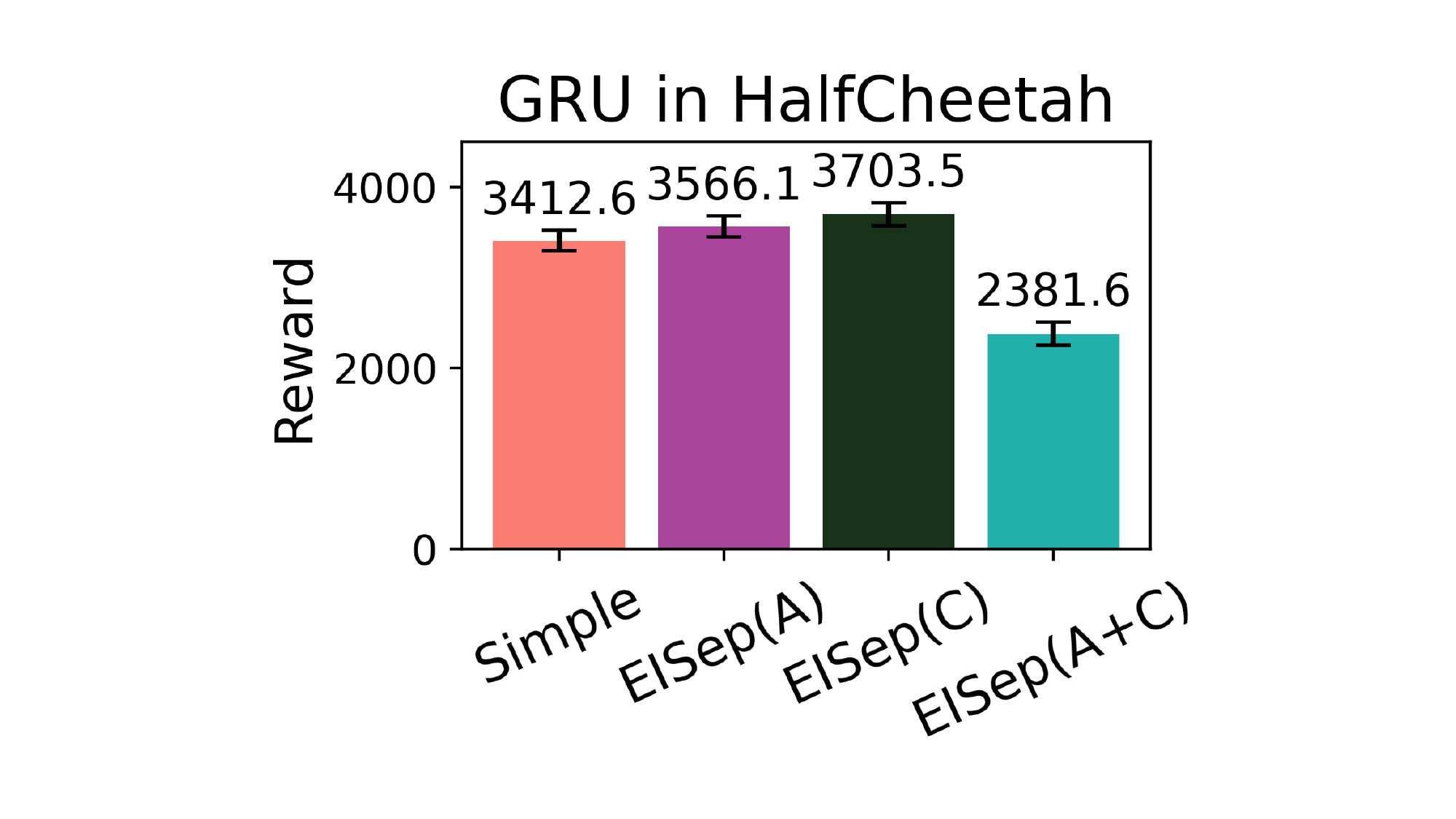}
  \caption{Performance for EISep GRU on HalfCheetah.}
  \label{fig:eisep_gru}
\end{figure}

To show that EISep can generalize to different backbones, we also apply similar adaptation to GRU. EISep still performs better than Simple (Fig.~\ref{fig:eisep_gru}). 

%% file: section/appendix/rl_sparsity.tex
\begin{figure}{}
    \centering
\includegraphics[width=0.5\linewidth]{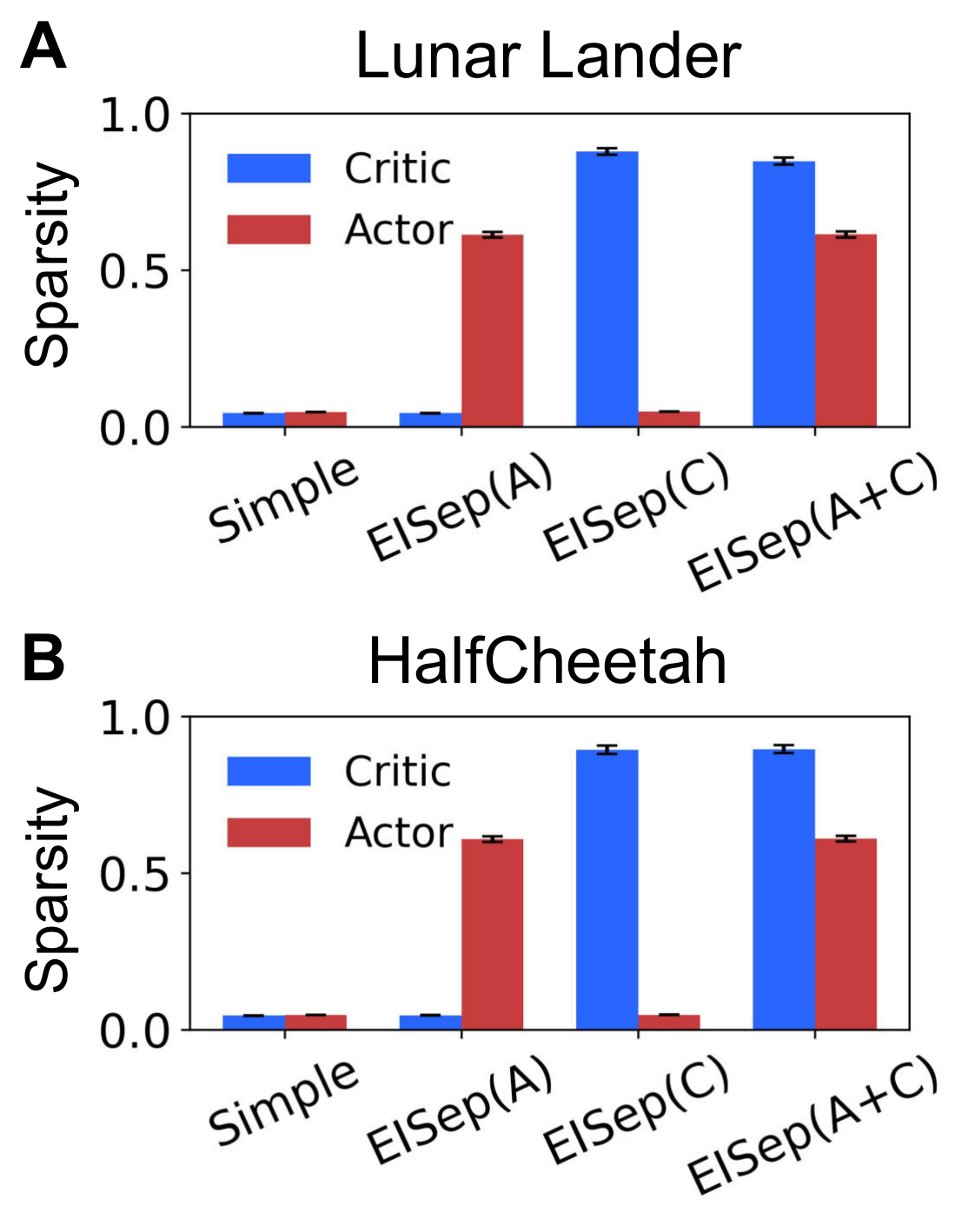}
  \caption{Sparsity averaged across all learning rates and seeds in the \textbf{A.} Lunar Lander task, and \textbf{B.} HalfCheetah task.}
    \vspace{-0.2in}
  \label{fig:rl_appendix_sparsity}
\end{figure}

Here, we examine the sparsity of the actor and critic modules in the reinforcement learning tasks. As expected, the EISep-parameterized modules exhibit higher sparsity, whereas modules that remain as vanilla RNNs (or LSTMs) show sparsity close to zero (Fig.~\ref{fig:rl_appendix_sparsity}A, B). This pattern is consistent with the sparsity observed for EISep in the multi-task cognitive benchmark (Sec.~\ref{sec:result_sparsity}), further indicating that EISep induces compact connectivity across different learning settings.

%% file: section/appendix/hmm_details.tex
The number of possible hidden states is $5$, close to the maximal possible number of phases ($6$, but \textit{Crash} state is extremely rare. See Appendix~\ref{sec:appendix_phase_detection}). 

\subsection{Long clips of hidden activities}

\begin{figure}
    \centering
\includegraphics[width=\linewidth]{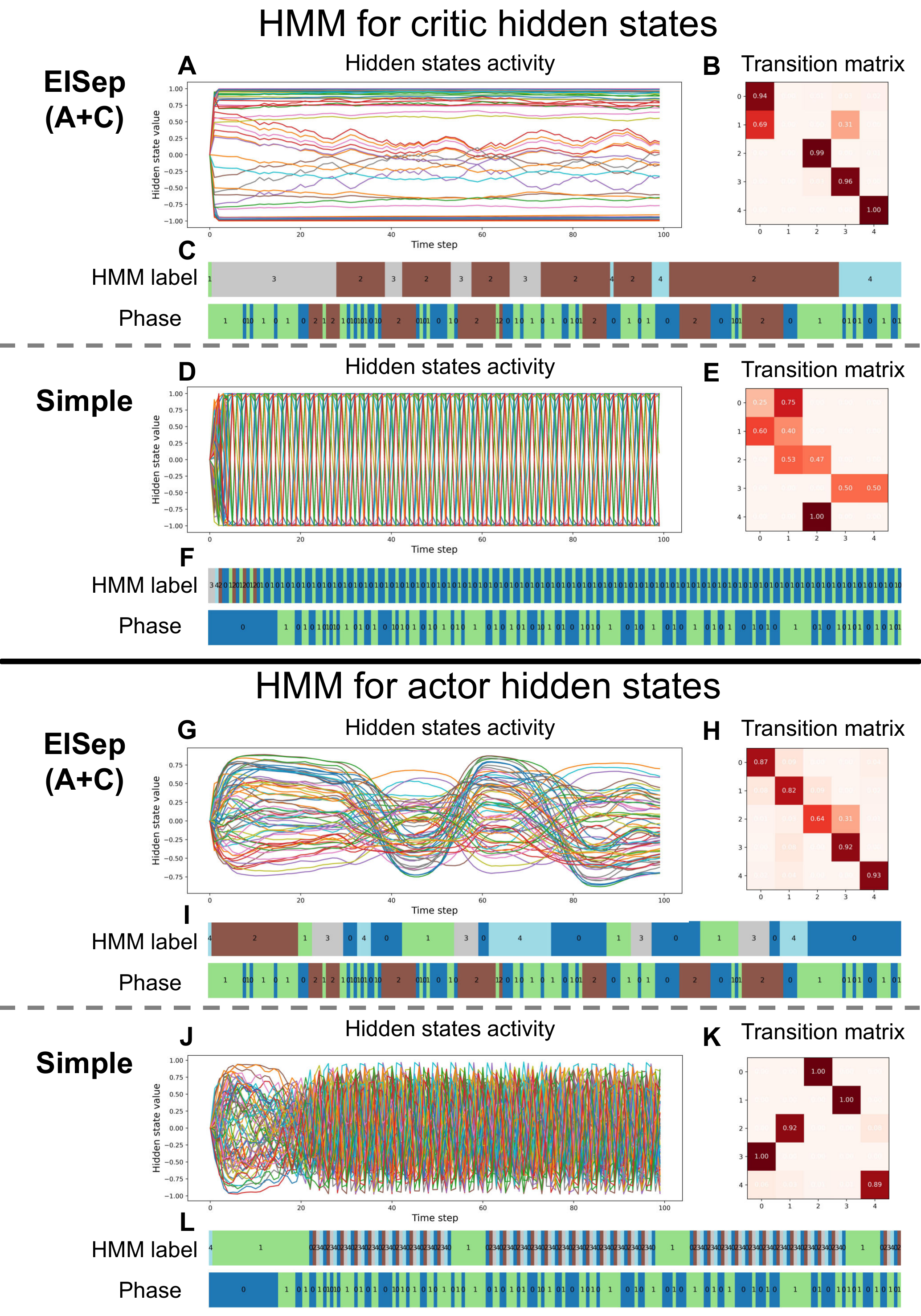}
  \caption{Extended visualizations of the HMM fitting results for both EISep(A+C) and Simple agents. \textbf{A.} A raw clip (length=100, beginning of the first episode) of hidden state activities for all 64 units in the EISep critic RNN. \textbf{B.} HMM transition matrix across all hidden labels. \textbf{C.} HMM labels (up) and manually annotated phases (bottom) for a clip (length=200, beginning of the first episode). The first 100 timesteps are aligned with the timesteps in \textbf{A}. \textbf{D-F.} Same as \textbf{A-C} for Simple critic RNN. \textbf{G.-I.} Same as \textbf{A-C} for EISep(A+C) actor RNN. \textbf{J.-L.} Same as \textbf{A-C} for Simple actor RNN. }
    \vspace{-0.2in}
  \label{fig:hmm_details}
\end{figure}

In Sec.~\ref{sec:rl_dynamics}, we only presented short clips of actor hidden activities. Here, to provide a more comprehensive visualization, we extend the temporal window to $100$ timesteps for hidden activities and provide fitted HMM labels for $200$ timesteps. The learning rate settings are the same as in Sec.~\ref{sec:rl_dynamics}:
\begin{align}
    \mathrm{lr(critic)}=0.005, \quad \mathrm{lr(actor)}=0.001.
\end{align}

As shown in Fig.~\ref{fig:hmm_details}A, D, G, and J, the hidden-state activity of the Simple exhibits highly regular, high-frequency oscillations, while EISep displays richer, more heterogeneous, and less stereotyped activity patterns. Notably, the oscillation period of Simple's hidden states is extremely short (approximately 4 time steps) compared to the full episode length (approximately 500 time steps), indicating that the recurrent dynamics have effectively collapsed onto a low-dimensional limit cycle. As a result, Simple hidden states fail to encode much meaningful information about the task structure or environment context.

This dynamical collapse is directly reflected in the HMM labels. In Simple, the inferred HMM labels switch rapidly and over-regularly (Fig.~\ref{fig:hmm_details}C, F, I, and L), tracking the fast oscillatory dynamics of the hidden states rather than revealing interpretable task phases. Correspondingly, the estimated transition matrices exhibit an ill-posed structure (Fig.~\ref{fig:hmm_details}B, E, H, and K): the self-transition probabilities are abnormally low (low stickiness), while transition probabilities to other states are disproportionately high (high entropy). This pattern indicates that HMM states are unstable and short-lived, consistent with the absence of persistent latent modes in the underlying RNN dynamics. In contrast, EISep produces transition matrices with high stickiness and low entropy, reflecting stable latent states and coherent temporal organization.

Taken together, these visualizations reinforce our earlier findings: EISep’s advantage lies in its ability to sustain rich, non-collapsed recurrent dynamics that support stable and informative latent states, and effective control.

\subsection{Quantitative metrics}

\begin{figure}
    \centering
\includegraphics[width=0.7\linewidth]{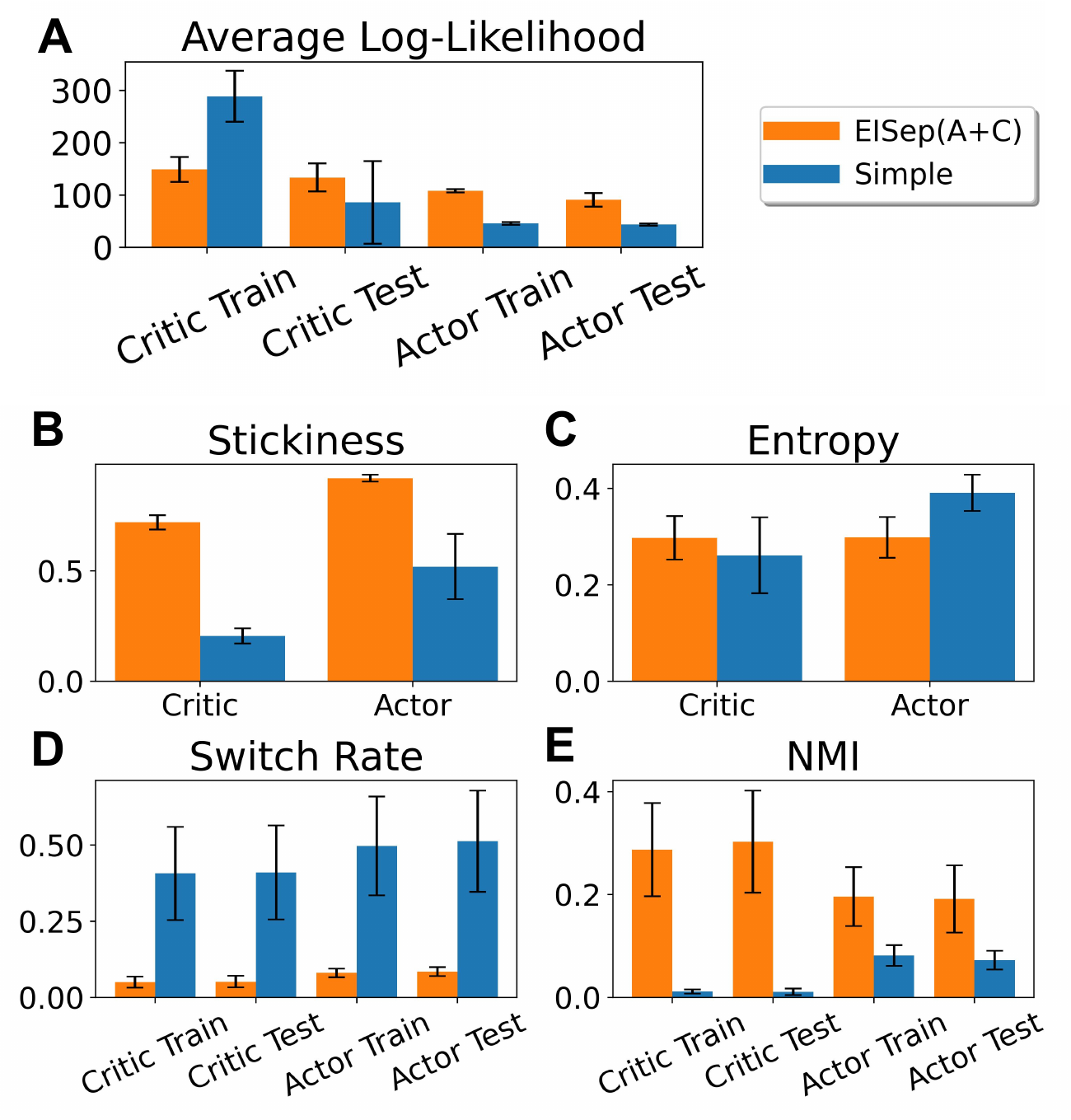}
  \caption{Different metrics of the HMM results of EISep. \textbf{A.} The HMM log-likelihood on the train/test dataset for the critic/actor of the two agents(EISep/Simple). \textbf{B}. Stickiness of the transition matrix. \textbf{C}. Entropy of the transition matrix. \textbf{D}. Switch rate of HMM hidden states. \textbf{E}. Normalized mutual information between HMM hidden states and handmade labels. }
    \vspace{-0.2in}
  \label{fig:hmm_metrics}
\end{figure}

Here, we use HMM to fit hidden state trajectories in EISep and Simple agents. EISep achieves comparable or higher test log-likelihood than Simple (Fig.~\ref{fig:hmm_metrics}A), indicating that the latent trajectories learned under E-I separation are more predictable over time. It seems that Simple critic achieves higher training log-likelihood, but as shown in Fig.~\ref{fig:hmm_details}, the hidden states of the Simple critic have collapsed and the higher training log-likelihood is actually overfitting. As quantified in Sec.~\ref{sec:rl_dynamics}, these latent representations contain negligible mutual information with the environmental states (Fig.~\ref{fig:hmm_metrics}E). Consequently, they provide little utility for downstream decision-making and lead to inferior control performance (Fig.~\ref{fig:rl_best_model_performance_eisep}B).

Next, we focus on the temporal organization, stability, and behavioral relevance of the latent dynamics. To quantify the temporal coherence of the inferred latent dynamics, we propose two metrics to probe the transition structure: (1) Stickiness $s$ measures the self-transition probabilities, and is defined as $s=\mathrm{trace}(P)$ where $P$ is the estimated transition matrix. (2) Entropy $h$ measures uncertainty in state transitions and is defined as $h=-\frac{1}{L}\sum_{i,j} P_{i,j}\log P_{i,j}$ where $L$ is the number of hidden states. EISep networks exhibits higher stickiness than Simple (Fig.~\ref{fig:hmm_metrics}B), consistent with the fact that their latent states persist over longer temporal windows (Fig.~\ref{fig:hmm_details}). In the meantime, EISep's actor shows lower transition entropy than Simple's actor (Fig.~\ref{fig:hmm_metrics}C), consistent with less jittery state transitions. EISep agents also exhibits substantially lower HMM state-switch rates in both actor and critic networks (Fig.~\ref{fig:hmm_metrics}D). Together, these measures corroborate quantitatively that E-I separation stabilizes recurrent dynamics and develops smooth evolution of internal representations over time. The higher NMI of EISep, as introduced in Sec.~\ref{sec:rl_dynamics}, shows stronger alignment between latent dynamics and interpretable task phases, indicating that EISep's latent assignments are environmentally informative.

%% file: section/appendix/ablation.tex
\begin{figure}[t]
\centering
\begin{tikzpicture}[
    box/.style={draw, rounded corners, align=center, minimum width=3.2cm, minimum height=1.1cm, font=\small},
    smallbox/.style={draw, rounded corners, align=center, minimum width=2.7cm, minimum height=0.9cm, font=\footnotesize},
    arrow/.style={-{Latex}, thick},
    node distance=0.9cm and 0.9cm
]

\node[box, fill=gray!10] (eisep) {
\textbf{EISep}\\
$W=\mathrm{ReLU}(W_0)D$\\
learnable neuron signs
};

\node[smallbox, below left=of eisep, xshift=-1.1cm] (abs) {
\textbf{Abs-EISep}\\
$W=|W_0|D$\\
replace ReLU filter
};

\node[smallbox, below=of eisep] (fixed) {
\textbf{Fixed-EISep}\\
$W=\mathrm{ReLU}(W_0)D$\\
fixed neuron signs
};

\node[smallbox, below right=of eisep, xshift=1.1cm] (entry) {
\textbf{EntryEI}\\
$W=\mathrm{ReLU}(W_0)\odot K$\\
entry-wise signs
};

\node[smallbox, right=of eisep, xshift=1.2cm] (init) {
\textbf{EISep(N1/N2)}\\
same form as EISep\\
different initialization
};

\node[smallbox, left=of eisep, xshift=-1.2cm] (cstr) {
\textbf{Cstr-Simple}\\
Simple RNN\\
EISep-style initialization
};

\draw[arrow] (eisep) -- (abs)
    node[midway, above left, font=\scriptsize, align=center] {filter};
\draw[arrow] (eisep) -- (fixed)
    node[midway, right, font=\scriptsize, align=center] {sign\\learnability};
\draw[arrow] (eisep) -- (entry)
    node[midway, above right, font=\scriptsize, align=center] {sign\\granularity};
\draw[arrow] (eisep) -- (init)
    node[midway, above, font=\scriptsize] {initialization};
\draw[arrow] (eisep) -- (cstr)
    node[midway, above, font=\scriptsize] {architecture};


\end{tikzpicture}

\vspace{-0.05in}
\caption{
Schematic comparison of EISep and its ablation variants. 
Abs-EISep replaces the ReLU recurrent-weight filter with an absolute-value filter; 
Fixed-EISep removes learnable neuron signs; 
EntryEI relaxes neuron-level E--I separation into entry-wise sign masks; 
EISep(N1/N2) preserves the EISep parameterization but changes the initialization scheme; 
and Cstr-Simple tests whether EISep-style initialization alone explains the observed gains.
}
\label{fig:ablation_schematic}
\end{figure}
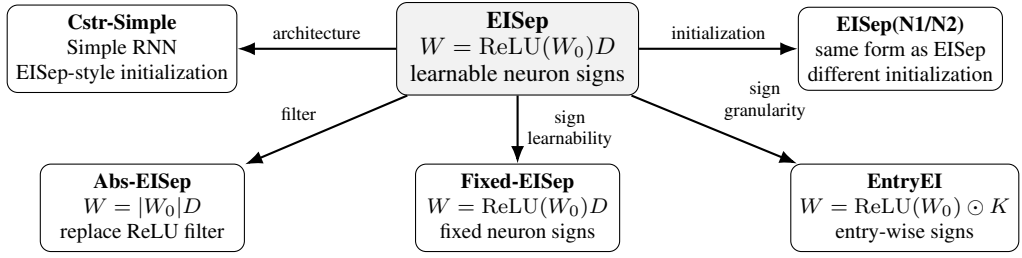

To disentangle the contributions of individual design choices in EISep, such as recurrent weight filter, learnable signs, and neuron-level sign constraints, we further conduct a systematic ablation study here, as a more detailed complementary to Sec.~\ref{sec:result_ablation}. We summarize different ablation models in Fig.~\ref{fig:ablation_schematic}.

\subsection{Absolute as the recurrent weight filter of EISep}\label{sec:appendix_abs_eisep}

\begin{figure}[]
    \centering
\includegraphics[width=\linewidth]{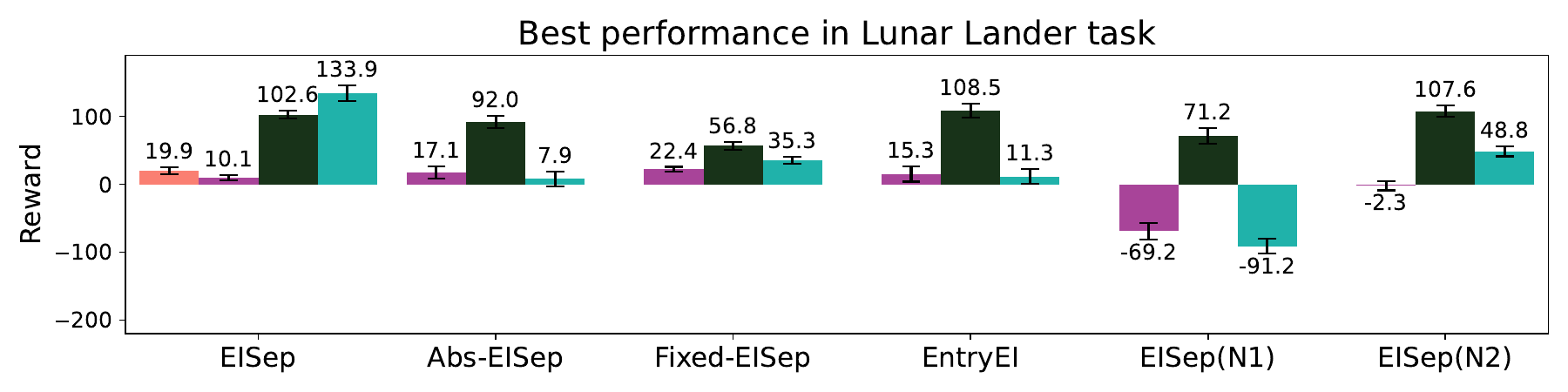}
  \caption{Performance in the Lunar Lander task for different ablation models. The performance of Cstr-Simple has been shown in Sec.~\ref{sec:result_ablation}. }
  \label{fig:appendix_rl_lunarlander_ablation}
\end{figure}

\begin{figure}
    \centering
\includegraphics[width=\linewidth]{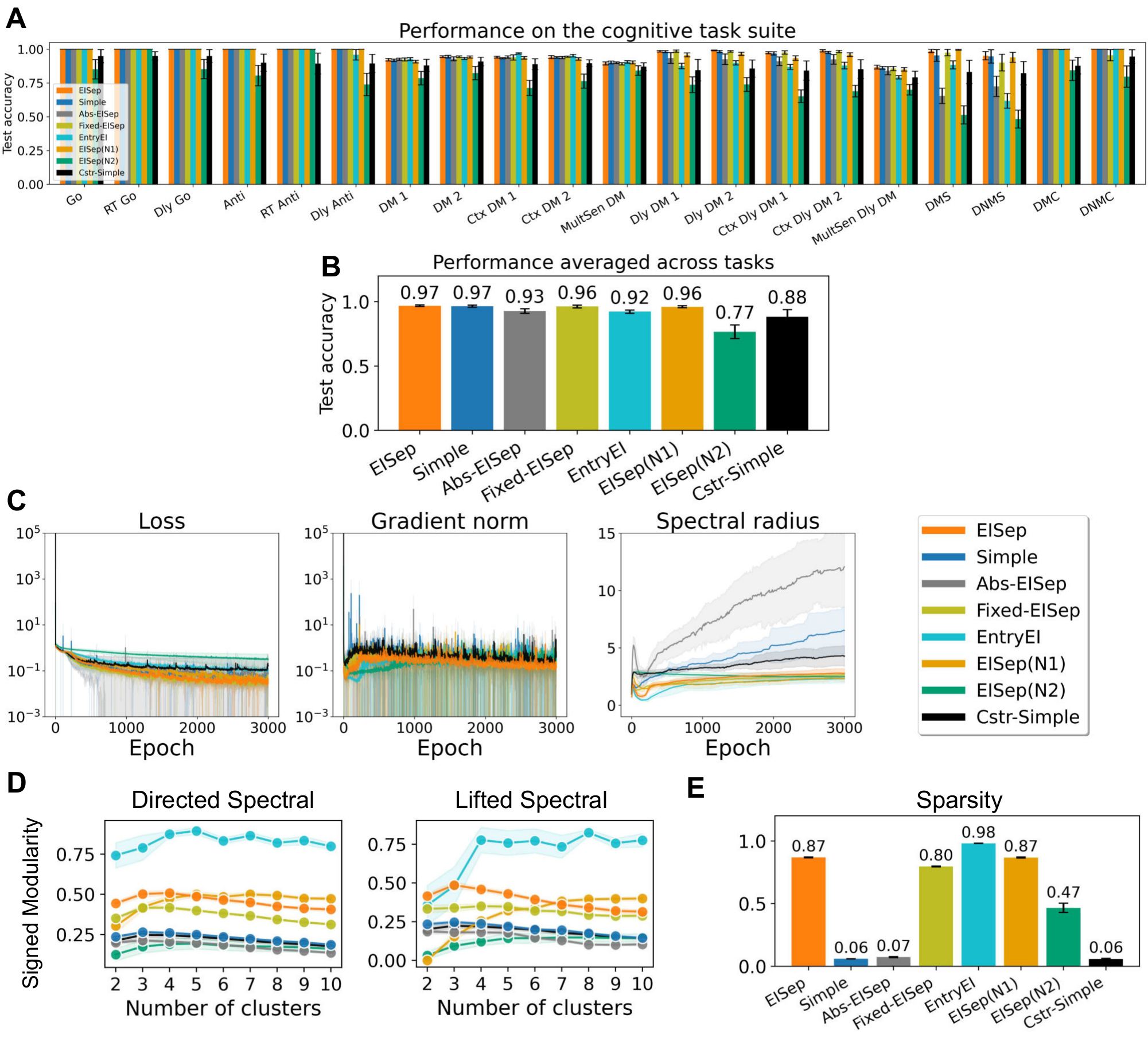}
    \vspace{-0.1in}
  \caption{Metrics on the cognitive task suite for different ablation models. We compare Abs-EISep, Fixed-EISep, and EntryEI against the EISep and Simple baselines. \textbf{A}. Task-wise performance of the 5 models across 20 cognitive tasks. \textbf{B}. Performance averaged across all tasks for each model. \textbf{C}. Recurrent weight sparsity averaged across tasks for different models. \textbf{D}. Training dynamics over the full training process, including decoding loss (left), gradient norm (middle), and spectral radius of the recurrent weight matrix (right). }
  \label{fig:appendix_ablation}
\end{figure}

There remains a question as to whether the sparsity completely arises from the ReLU operator (Eq.~\ref{eq:w=w0d}) and is independent from the performance. To disentangle these effects, we replace the ReLU nonlinearity with an absolute value operator: $W=|W_0| D$ and refer to the new variant as \textbf{Abs-EISep}. Under this modification, the sparsity advantage indeed vanishes (Fig.~\ref{fig:appendix_ablation}E). However, the performance is also affected. While most tasks show little or no degradation, performance drops significantly on two relatively more challenging matching tasks—DMS and DNMS (Fig.~\ref{fig:appendix_ablation}A), which require comparing two angles randomly sampled from $[0^\circ, 360^\circ)$. This effect is also reflected in the averaged performance across tasks (Fig.~\ref{fig:appendix_ablation}B), which decreases from 0.97 for EISep to 0.93 for Abs-EISep (paired t-test, $p=1.2\times10^{-5}$). Moreover, Abs-EISep exhibits an unbounded growth in spectral radius (Fig.~\ref{fig:appendix_ablation}C right), as well as reduced performance in continual learning (Fig.~\ref{fig:appendix_continual_ablation}J) and reinforcement learning (Fig.~\ref{fig:appendix_rl_lunarlander_ablation}), indicating unstable recurrent dynamics during training. Together, these results demonstrate that the ReLU operator plays a critical role not only in inducing sparsity, but also in stabilizing recurrent dynamics and supporting performance on challenging tasks. This suggests that sparsity of EISep is not an incidental side effect of the parameterization, but is associated with performance through the ReLU-based constraint on recurrent weights.

\begin{figure}[]
    \centering
\includegraphics[width=0.7\linewidth]{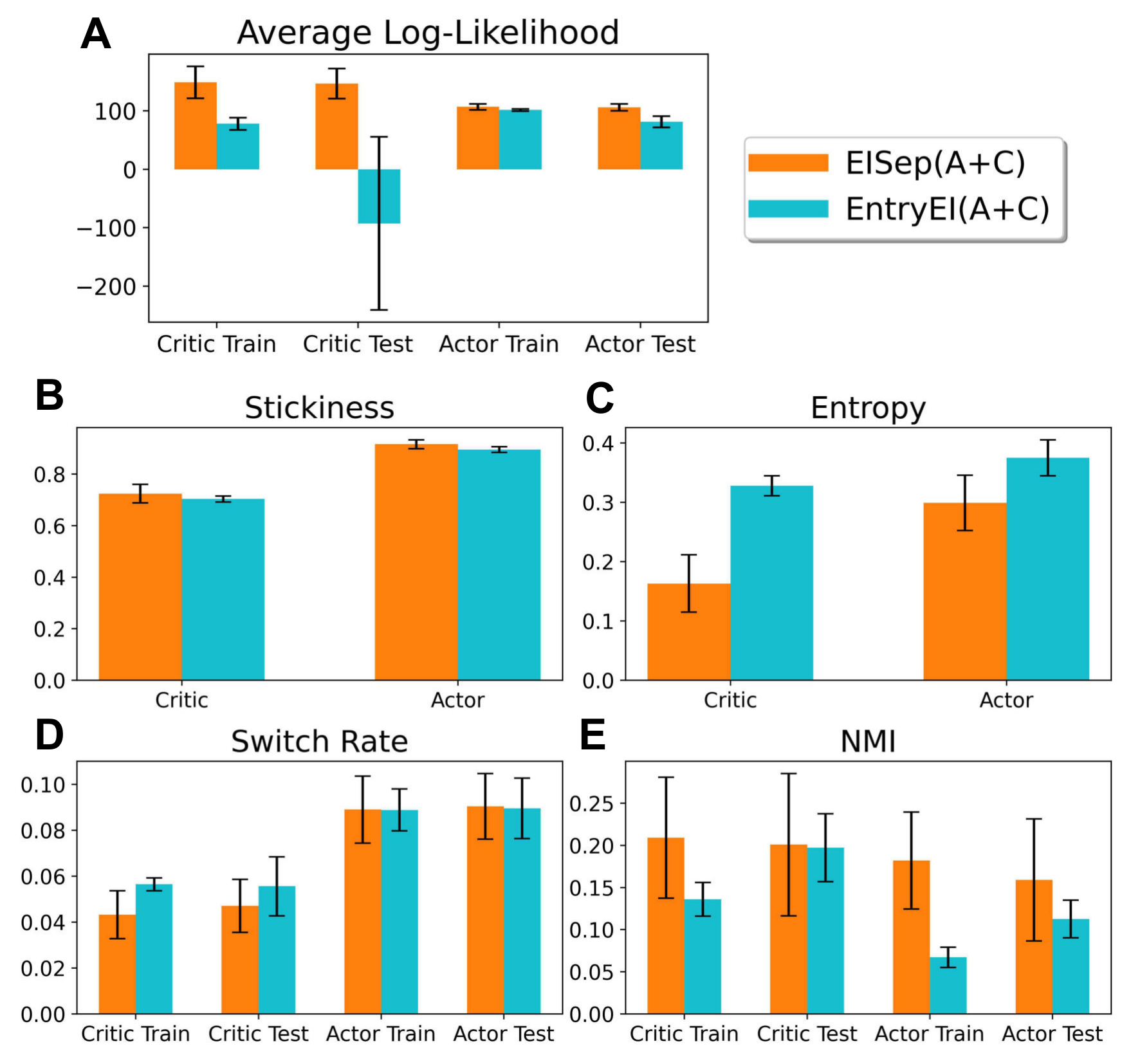}
    \vspace{-0.1in}
  \caption{HMM metrics for EntryEI(A+C), including (\textbf{A}) log-likelihood, (\textbf{B}) stickiness of transition matrices, (\textbf{C}) entropy of transition matrices, (\textbf{D}) switch rate of HMM labels, and (\textbf{E}) normalized mutual information between HMM labels and handmade labels. Same as Fig.~\ref{fig:hmm_metrics}}.
    \vspace{-0.2in}
  \label{fig:hmm_ablation}
\end{figure}

\subsection{EISep with fixed neuron signs}\label{sec:appendix_fixed_eisep}
\begin{figure}{}
    \centering
\includegraphics[width=\linewidth]{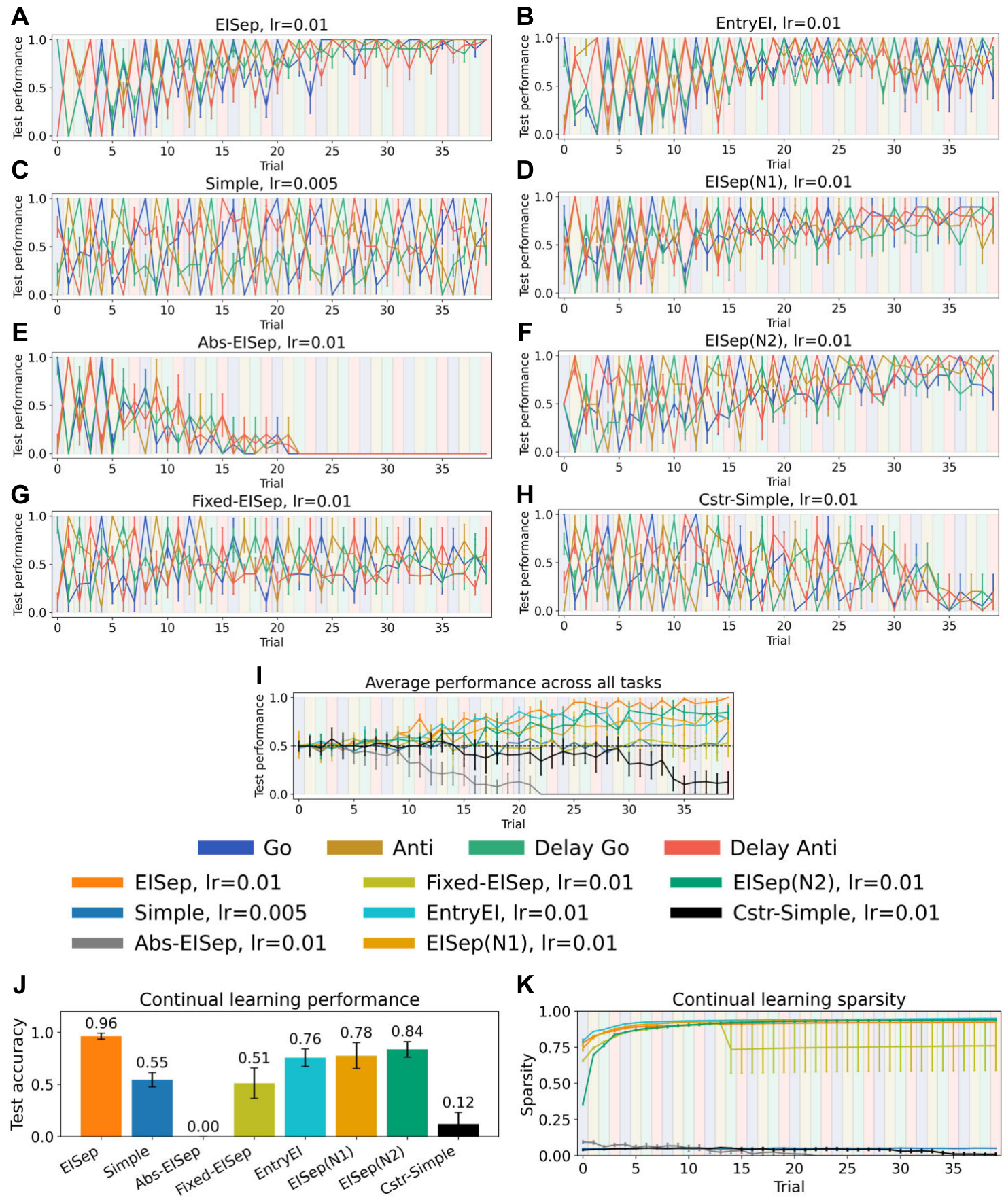}
  \caption{
    \textbf{A--H.} Training curves of different ablation models with their best learning rates in continual learning.
    \textbf{I.} Average performance across tasks.
    \textbf{J.} Final-round performance averaged over four trials. \textbf{K.} The sparsity of all models across trials.}
  \label{fig:appendix_continual_ablation}
\end{figure}

Although we have shown in Sec.~\ref{sec:appendix_logit_convergence} that the neuron signs have converged at the early stage of training, it is still a question whether allowing the signs to be learnable is itself beneficial, or the early convergence is the only side effect of the learnable sign. To address this question, we randomly sample the sign matrix $D$ (Eq.~\ref{eq:w=w0d}) and fix it throughout the training. This variant, referred to as \textbf{Fixed-EISep} achieves performance, sparsity, and training dynamics comparable to those of the learnable EISep model on the multi-task cognitive benchmark (Fig.~\ref{fig:appendix_ablation}B), indicating that E/I separation alone is sufficient for solving these relatively simple tasks. However, Fixed-EISep performs worse than EISep in continual learning (Fig.~\ref{fig:appendix_continual_ablation}J) and the reinforcement learning task (Fig.~\ref{fig:appendix_rl_lunarlander_ablation}), suggesting that a fixed sign structure lacks the flexibility required to capture more intricate dynamics and to support complex control behaviors.

\begin{figure}
    \centering
\includegraphics[width=0.6\linewidth]{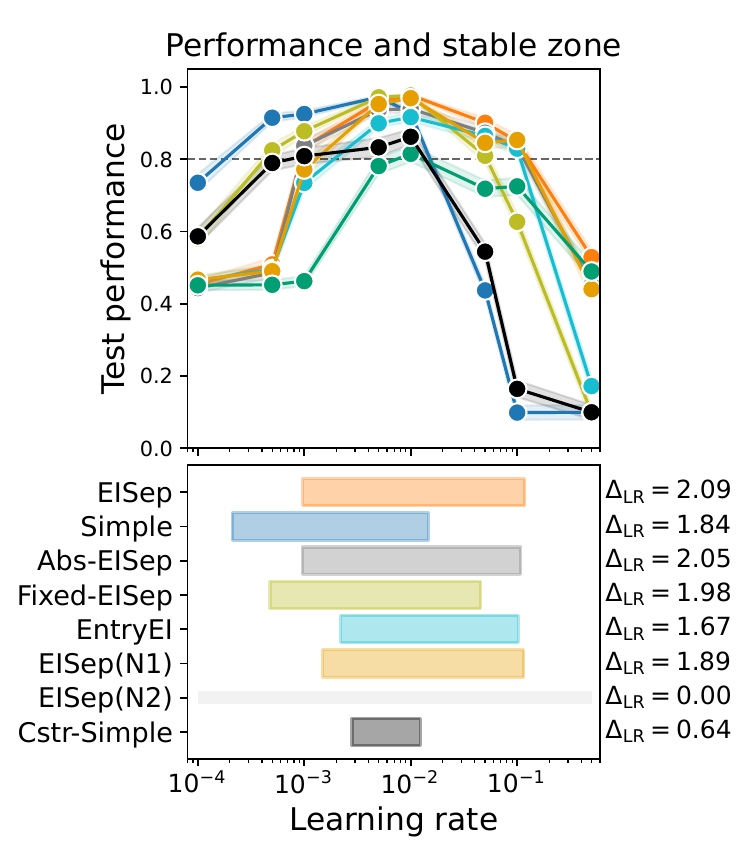}
    \vspace{-0.1in}
  \caption{Performance and stable zones for different ablation models vs. learning rates in the cognitive task suite. }
  \label{fig:appendix_multilr_ablation}
\end{figure}

\begin{figure}
    \centering
\includegraphics[width=0.5\linewidth]{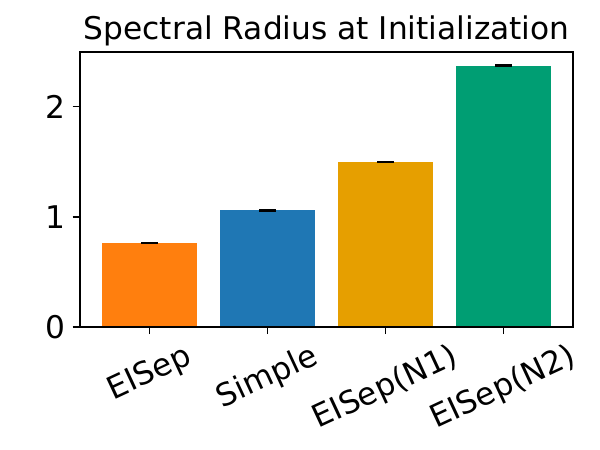}
    \vspace{-0.1in}
  \caption{The spectral radius of different models at initialization ($n=1000$, Error bar indicates standard error).}
  \label{fig:appendix_spectral_radius_initialization}
\end{figure}

\subsection{Entry-wise masks for the EISep recurrent weight}\label{sec:appendix_entryei}

Another question is whether EISep's advantages stem from the specific \emph{neuron-level} E-I separation induced by Dale’s principle, or more generally from imposing value-sign separation on weights. Therefore, we introduce \textbf{EntryEI}, which relaxes the column-wise E-I constraint of EISep to an entry-wise one.

Here is the comparison of the designs. The original EISep enforces a consistent outgoing sign for each presynaptic neuron through a diagonal sign matrix \(D\) (Eq.~\ref{eq:w=w0d})
\begin{equation}
    W = \mathrm{ReLU}(W_0) D , \qquad D_{ii} \in \{-1, 1\}.
\end{equation}
In contrast, EntryEI applies an independent sign constraint to each individual weight,
\begin{equation}
    W = \mathrm{ReLU}(W_0) \odot K , \qquad K_{ij} \in \{-1, 1\},
\end{equation}
where \(K\) is a dense matrix and \(\odot\) denotes element-wise multiplication. 

This modification isolates the role of neuron-level sign consistency while increasing the number of free parameters, allowing each recurrent connection to independently select its excitatory or inhibitory sign. From the structural viewpoint, EntryEI therefore represents a similar but more powerful and less constrained model class than EISep. Empirically, it also achieves a similar structural advantage with EISep: high sparsity (Fig.~\ref{fig:appendix_ablation}E), indicating that sparsity alone can be recovered without neuron-level E-I separation. However, EntryEI consistently underperforms EISep on several cognitively demanding tasks, including DMS, DNMS, Dly DM, and Ctx Dly DM (Fig.~\ref{fig:appendix_ablation}A) in the multi-task cognitive benchmark. It also exhibits narrower stable learning-rate regimes (Fig.~\ref{fig:appendix_multilr_ablation}), inferior RL performance (Fig.~\ref{fig:appendix_rl_lunarlander_ablation}) and reduced dynamic coherence (Fig.~\ref{fig:hmm_ablation}) in reinforcement learning. 

\subsection{EISep with different initialization}\label{sec:appendix_init_ablation}

Here we show how the initialization method affects the task performance of EISep. We adapt the method from ColEI~\citep{li2023learning} and DBPN~\citep{balwani2025constructing} to initialize the value matrix $Q$ and the E-I logit matrix $M$. The initialization of $Q$ strictly follows the original implementations~\ref{sec:appendix_baseline_init}. However, we have learnable E-I identities while they determine signs at the beginning. To address this gap, we assign $M_i=[1.0,0.0]$ to each excitatory neuron $i$ and $M_j=[0.0,1.0]$ to each inhibitory neuron $j$, and calculate sign matrix $D$ as before (Sec.~\ref{sec:definition_eisep}). EISep with ColEI-style initialization is denoted "\textbf{EISep(N1)}", and EISep with DBPN-style initialization is denoted "\textbf{EISep(N2)}". Since DANN implements Dale's principle using a fundamentally different structure with three coupled matrices (Appendix~\ref{sec:appendix_baseline_init}), we do not include it in this comparison.

On the multi-task cognitive benchmark, EISep(N1) achieves similar average performance and sparsity (Fig.~\ref{fig:appendix_ablation}B, E), but exhibits a narrower stable learning-rate zone (Fig.~\ref{fig:appendix_multilr_ablation}) and lower continual learning (Fig.~\ref{fig:appendix_ablation}J) and RL performance (Fig.~\ref{fig:appendix_rl_lunarlander_ablation}). EISep(N2) cannot match EISep on most evaluation metrics. We further find these initialization methods yield larger spectral radii (Fig.~\ref{fig:appendix_spectral_radius_initialization}), consistent with their degraded performance. These results suggest that the proper initialization method of EISep constrains the spectral radius and improves training stability and overall performance. 

\subsection{Simple with EISep-style initialization}
To test whether EISep’s advantage stems from initialization alone, we initialize Simple using the same scheme and denote the resulting variant as \textbf{Cstr-Simple}. This model still fails to match EISep across all three tasks, and even underperforms Simple on multi-task (Fig.~\ref{fig:appendix_multilr_ablation}) and continual learning (Fig.~\ref{fig:appendix_continual_ablation}J). See Fig.~\ref{fig:hmm_dynamics_ablation}F for RL performance. These results suggest that EISep-style initialization alone does not account for the observed gains, and that initialization strategies may not transfer directly across different model architectures.

Together, these results demonstrate that sparsity and modularity alone are insufficient to account for the advantages observed in EISep, and that increased parameterization does not necessarily translate into improved functional performance. More broadly, this comparison highlights that biologically motivated constraints can outperform less realistic but more flexible alternatives, even when the latter possess a larger hypothesis space. The failure of EntryEI suggests that less constrained synaptic sign organization—while more expressive—may hinder the emergence of stable, interpretable, and task-relevant dynamics. In contrast, EISep’s neuron-level E-I separation provides a favorable inductive bias that supports robust learning and generalization across both cognitive tasks and reinforcement learning.

%% file: section/appendix/critic_actor_asymmetry.tex
An interesting emergent phenomenon in experiments is worth noting. In RL tasks, EISep parameterization exhibits different effects on the critic and actor. EISep(C) outperforms EISep(A) across nearly all scenarios, and EISep(A) often performs worse than Simple (Fig.~\ref{fig:rl_best_model_performance_eisep}B, C, Fig.~\ref{fig:appendix_lunarlander_baselines},  Fig.~\ref{fig:appendix_lunarlander_ffn}, Fig.~\ref{fig:appendix_continual_ablation}). This advantage is unlikely to be explained solely by the higher learning rate of critic, since this advantage still exists even when the critic has a lower learning rate than the actor (Sec.~\ref{sec:appendix_lunarlander_eisep_lrall}). Notably, a similar asymmetry has also been discovered for other inductive biases like modularity~\cite{zhang2024inductive} where modularizing the critic produces more benefits. This result suggests that Dale's principle, as an inductive bias to improve stability and modularity, aligns more closely with the critic's computation, like value estimation or belief update, than with the actor's computation, like flexible and context-dependent control. From the machine learning perspective, the critic’s bootstrapped value updates introduce strong correlations between predicted and target values~\citep{mnih2015human}, forming a self-referential feedback loop that is a well-known source of instability~\citep{van2016deep, kumar2020conservative}. We assume that EISep acts as a dynamical regularizer that mitigates this bootstrapping-induced instability. In contrast, the actor optimizes a supervised-like policy objective without bootstrapping and is therefore less susceptible to such instabilities. Enforcing EISep constraints may thus unnecessarily restrict the actor. EISep(A+C) retains the benefit of the EISep critic but additionally constrains the actor, explaining why it can underperform EISep(C). This aligns with observations in supervised learning tasks (Appendix~\ref{sec:appendix_supervised_learning}). However, strict control is needed to confirm this conjecture.